\RequirePackage{fix-cm}
\documentclass[twocolumn]{svjour3}       %
\smartqed  %

\usepackage{amsmath}
\usepackage{times}
\usepackage{epsfig}
\usepackage{graphicx}
\usepackage{amssymb}
\usepackage{balance}
\usepackage[english]{babel}
\usepackage{hyphenat}
\usepackage{times}
\usepackage{epsfig}
\usepackage{graphicx}
\usepackage{amsmath}
\usepackage{amssymb}
\usepackage{amsfonts}
\usepackage{tabularx}
\usepackage{bbm}
\usepackage{mathtools}
\usepackage{multirow}
\usepackage[dvipsnames,table]{xcolor}
\usepackage[pagebackref=true,breaklinks=true,colorlinks,bookmarks=false]{hyperref}
\usepackage{flushend}
\usepackage{multirow}
\usepackage{booktabs}
\usepackage{bbding}
\usepackage{pifont}
\usepackage{wasysym}
\hypersetup{colorlinks=true,linkbordercolor=red,linkcolor=Green,citecolor=Green,pdfborderstyle={/S/U/W 1}}
\usepackage[inline]{enumitem}
\usepackage{adjustbox}
\usepackage{stmaryrd}
\usepackage{trimclip}

\newcommand{\comment}[1]{}

\newcommand{\bI}{\mathbf{I}}

\newcommand{\bF}{\mathbf{F}}

\newcommand{\bK}{\mathbf{K}}

\newcommand{\fig}[1]{Fig.~\ref{fig:#1}}

\newcommand{\tbl}[1]{Table~\ref{tbl:#1}}
\newcommand{\secref}[1]{Section~\ref{sec:#1}}

\definecolor{orange}{rgb}{1,0.5,0}
\definecolor{blue}{rgb}{0,0,0.6}

\definecolor{color1}{RGB}{0,199,1}
\definecolor{color2}{RGB}{224,43,28}

\newcommand{\ky}[1]{{\color{orange}{#1}}}
\newcommand{\KY}[1]{{\color{orange}{\bf [Kwang: #1]}}}
\newcommand{\et}[1]{{\color{magenta}{#1}}}
\newcommand{\ET}[1]{{\color{magenta}{\bf [Eduard: #1]}}}

\newcommand{\PF}[1]{{\color{blue}{\bf [Pascal: #1]}}}

\newcommand{\AM}[1]{{\color{OliveGreen}{\bf [Anastasiia: #1]}}}

\newcommand{\DM}[1]{{\color{Purple}{\bf [Dmytro: #1]}}}

\newcommand{\JM}[1]{{\color{Aquamarine}{\bf [Jiri: #1]}}}

\newcommand{\YJ}[1]{{\color{Bittersweet}{\bf [Yuhe: #1]}}}
\newcommand{\edit}[1]{{\color{Cyan}{#1}}}
\newcommand{\revision}[1]{{\color{Cyan}{#1}}}

\renewcommand{\ky}[1]{#1}
\renewcommand{\KY}[1]{}
\renewcommand{\et}[1]{#1}
\renewcommand{\ET}[1]{}

\renewcommand{\PF}[1]{}

\renewcommand{\AM}[1]{}

\renewcommand{\DM}[1]{}

\renewcommand{\JM}[1]{}

\renewcommand{\YJ}[1]{}
\renewcommand{\edit}[1]{#1}
\renewcommand{\revision}[1]{#1}

\newcommand{\bE}{\mathbf{E}}

\newcommand{\imsize}{1cm}
\newcommand{\hspacer}{0.3mm}
\newcommand{\vspacer}{0.1mm}

\newcommand{\numim}{N}
\newcommand{\numfeat}{K}

\newcommand{\covisibility}{v}

\newcommand{\imscl}{0.01}

\newcommand{\ransacConf}{\tau}

\newcommand{\ransacReprojTh}{\eta}
\newcommand{\ransacMaxIters}{\Gamma}

\newcommand{\ratiotest}{r}

\makeatletter
\DeclareRobustCommand{\shortto}{%
  \mathrel{\mathpalette\short@to\relax}%
}

\newcommand{\short@to}[2]{%
  \mkern2mu
  \clipbox{{.5\width} 0 0 0}{$\m@th#1\vphantom{+}{\shortrightarrow}$}%
  }
\makeatother

\renewcommand{\paragraph}[1]{\vspace{0.5em}\noindent{\bf#1}}

\newcommand{\ie}{\textit{i.e.}}
\newcommand{\eg}{\textit{e.g.}}
\newcommand{\etal}{\textit{et al.}}

\makeatletter
\newcommand\footnoteref[1]{\protected@xdef\@thefnmark{\ref{#1}}\@footnotemark}
\makeatother

\begin{document}

\sloppy

\title{Image Matching Across Wide Baselines: From Paper to Practice%
\thanks{%
This work was partially supported by the Natural Sciences and Engineering
Research Council of Canada (NSERC) Discovery Grant ``Deep Visual Geometry
Machines'' (RGPIN-2018-03788), by systems supplied by Compute Canada, and by Google's Visual Positioning Service.
DM and JM were supported by OP VVV funded project CZ.02.1.01/0.0/0.0/16 019/0000765 ``Research Center for Informatics''. DM was also supported by CTU student grant SGS17/185/OHK3/3T/13 and by the Austrian Ministry for Transport, Innovation and Technology, the Federal Ministry for Digital and Economic Affairs, and the Province of Upper Austria in the frame of the COMET center SCCH. AM was supported by the Swiss National Science Foundation.
}
}
\subtitle{}

\author{%
Yuhe Jin \and
Dmytro Mishkin \and
Anastasiia Mishchuk \and
Jiri Matas \and
Pascal Fua \and
Kwang Moo Yi \and
Eduard Trulls%
}

\institute{%
Y. Jin, K.M. Yi \at
Visual Computing Group, University of Victoria\\
\email{\{yuhejin, kyi\}@uvic.ca}%
\and
A. Mishchuk, P. Fua \at
Computer Vision Lab,
\'{E}cole Polytechnique F\'{e}d\'{e}rale de Lausanne\\
\email{\{anastasiia.mishchuk, pascal.fua\}@epfl.ch}
\and
D. Mishkin, J. Matas \at
Center for Machine Perception, Czech Technical University in Prague\\
\email{\{mishkdmy, matas\}@cmp.felk.cvut.cz}
\and
E. Trulls \at
Google Research\\
\email{trulls@google.com}
}

\date{Received: date / Accepted: date}

\maketitle

\begin{abstract}
We introduce a comprehensive benchmark for local features and robust estimation algorithms, focusing on the downstream task -- the accuracy of the reconstructed camera pose -- as our primary metric. Our pipeline's modular structure allows easy integration, configuration, and combination of different methods and heuristics. This is demonstrated by embedding dozens of popular algorithms and evaluating them, from seminal works to the cutting edge of machine learning research. We show that with proper settings, classical solutions may still outperform the {\em perceived state of the art}.

Besides establishing the {\em actual state of the art}, the conducted experiments reveal unexpected properties of Structure from Motion (SfM) pipelines that can help improve their performance, for both algorithmic and learned methods.
Data and code are online\footnote{\label{code}{\scriptsize \url{https://github.com/vcg-uvic/image-matching-benchmark}}}, providing an easy-to-use and flexible framework for the benchmarking of local features and robust estimation methods, both {\em alongside} and {\em against} top-performing methods. This work provides a basis for \revision{the Image Matching Challenge}\footnote{\label{website}{\scriptsize \url{https://vision.uvic.ca/image-matching-challenge}}}.
\end{abstract}
\begin{figure}
\centering
\renewcommand{\imsize}{6.5cm}
\renewcommand{\hspacer}{0.3mm}
\renewcommand{\vspacer}{0.1mm}
\renewcommand{\imscl}{0.01}
\includegraphics[scale=\imscl,width=0.495\linewidth, trim = 0 0 0 80, clip]{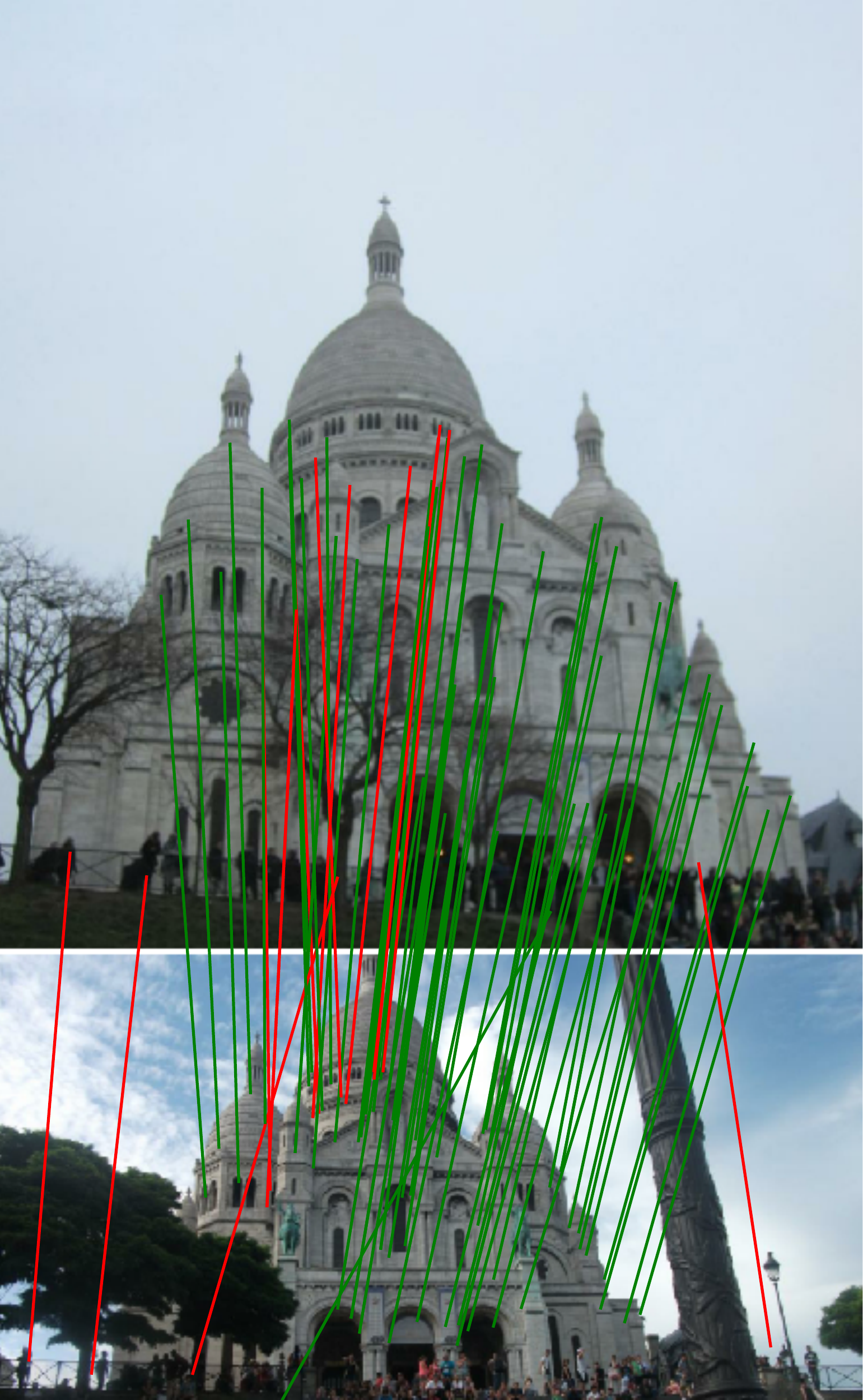}\hspace{\hspacer}%
\includegraphics[scale=\imscl,width=0.495\linewidth, trim = 0 0 0 80, clip]{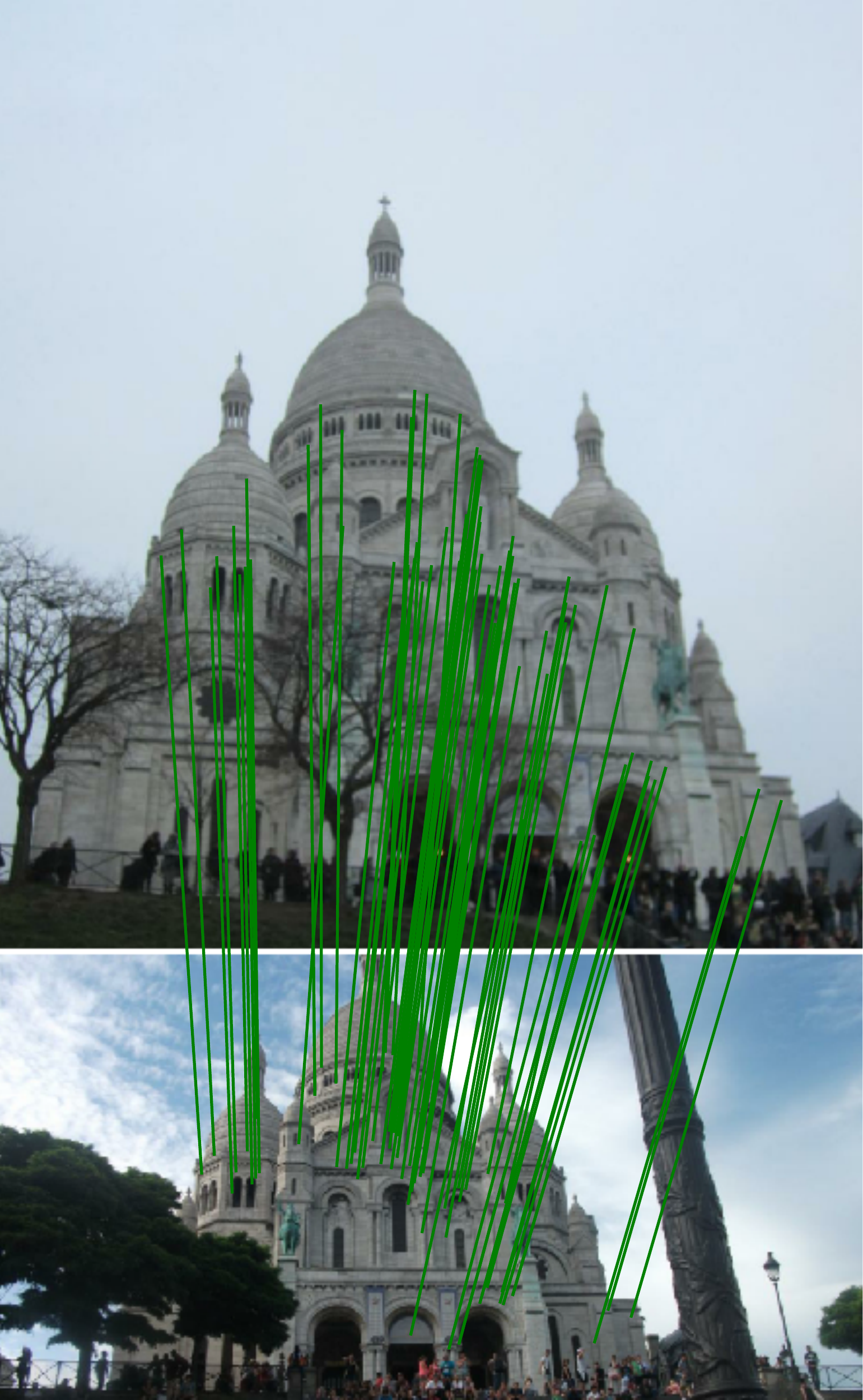}\hspace{\hspacer}%
\caption{
Every paper claims to outperform the state of the art. Is this possible, or an artifact of insufficient validation? On the left, we show stereo matches obtained with {\bf D2-Net} (2019)~\cite{Dusmanu19}, a state-of-the-art local feature, using OpenCV RANSAC with its default settings. We color the inliers in green if they are correct and in red otherwise. On the right, we show
{\bf SIFT} (1999)~\cite{Lowe04} with a carefully tuned MAGSAC~\cite{magsac2019} -- notice how the latter performs much better. This illustrates our take-home message: to correctly evaluate a method's performance, it needs to be embedded within the pipeline used to solve a given problem, and the different components in said pipeline need to be tuned carefully and jointly, which requires engineering and domain expertise.
We fill this need with a new, modular benchmark for sparse image matching,
incorporating dozens of built-in methods.}

\label{fig:teaser}
\vspace{-4mm}
\end{figure}

\vspace{-3mm}
\section{Introduction}
\label{sec:intro}

Matching two or more views of a scene is at the core of fundamental computer vision problems, including image retrieval~\cite{Lowe04,Arandjelovic16,Radenovic16,Tolias2016,Noh17}, 3D reconstruction~\cite{Agarwal09,Heinly15,Schoenberger16a,Zhu_2018_CVPR}, re-localization~\cite{Sattler12,Sattler18,Lynen19}, and SLAM~\cite{Mur15,DeTone17a,DeTone17b}.
\edit{Despite decades of research, image matching remains unsolved in the general, wide-baseline scenario.
Image matching is a challenging problem with many factors that need to be taken into account, e.g., viewpoint, illumination, occlusions, and camera properties.
It has therefore been traditionally approached with sparse methods -- that is, with local features.}

Recent effort have moved towards \edit{\emph{holistic}}, end-to-end solutions \cite{Kendall15a,Balntas_2018_ECCV,Bui_2019_ICCV}. 
\edit{Despite their promise},
they \edit{are yet to} outperform the \edit{\emph{separatists}}~\cite{Sattler19,zhou2019learn} \edit{that are based on the classical paradigm of step-by-step solutions.}
For example, in \edit{a classical wide baseline stereo pipeline~\cite{Pritchett98a}, one} 
(1)~extracts local features, such as SIFT \cite{Lowe04},
(2)~creates a list of tentative matches by nearest-neighbor search in descriptor space, and 
(3)~retrieves the pose with a minimal solver inside a robust estimator, such as the 7-point algorithm \cite{Hartley94_7pt} in a RANSAC loop \cite{Fischler81}.
To build a 3D reconstruction out of a set of images, same matches are fed to a bundle adjustment pipeline~\cite{Hartley00,Triggs00} to jointly optimize the camera intrinsics, extrinsics, and 3D point locations.
This modular structure simplifies engineering a solution to the problem and allows for incremental improvements, of which there have been hundreds, if not thousands.

New methods for each of the sub-problems, such as feature extraction and pose estimation, are typically studied in isolation, using intermediate metrics, which simplifies their evaluation.
However, there is no guarantee that gains in one part of the pipeline will translate to the final application, as these components interact in complex ways.
For example, patch descriptors, including very recent work~\cite{He18b,Wei18,Tian19,Mukundan19}, are often evaluated on Brown's seminal patch retrieval database~\cite{Brown10},
\edit{introduced in 2007}.
They 
show dramatic improvements -- \edit{up to 39x relative~\cite{Wei18}} -- over handcrafted methods such as SIFT, but it is unclear whether this 
\edit{remains} true on real-world applications. In fact, we later demonstrate that the gap
narrows dramatically when 
\edit{decades-old}
baselines are properly tuned.

We posit that it is \edit{critical}
to look beyond intermediate metrics and focus on downstream performance.
This is particularly important {\em now}
\revision{as, while
deep networks are reported to outperform algorithmic solutions on classical, \edit{sparse} problems such as outlier filtering \cite{Yi18a,Ranftl18,Zhang19,Sun19,brachmann2019ngransac}, bundle adjustment \cite{Tang19,shi2019selfsupervised}, SfM \cite{Vijayna17,Deepsfm2019} and SLAM \cite{Tateno_2017_CVPR,gradslam},
our findings in this paper suggest that this may not always be the case.
}
To this end, we introduce a benchmark for wide-baseline image matching, including:
\begin{enumerate}[label=(\alph*)]
  \item \edit{A dataset with thousands of phototourism images of 25 landmarks, taken from diverse viewpoints, with different cameras, in varying illumination and weather conditions -- all of which are necessary for a comprehensive evaluation.
  We reconstruct the scenes 
  with 
  SfM, without the need for human intervention,
  providing depth maps and ground truth poses for 26k images, and reserve another 4k for a private test set.}
  \item A modular pipeline incorporating dozens of methods for feature extraction, matching, and pose estimation,
  both classical and state-of-the-art, as well as multiple heuristics, \edit{all of} which can be swapped out and tuned separately.
  \item Two downstream tasks -- stereo and multi-view reconstruction -- evaluated with both downstream and intermediate metrics, for comparison.
  \item A thorough study of dozens of methods and techniques, both hand-crafted and learned, and their combination, along with a \edit{recommended} procedure for \edit{effective} hyper-parameter selection.
\end{enumerate}

The framework
\edit{enables}
researchers to evaluate how a new approach performs in a standardized pipeline, both {\em against} its competitors, and {\em alongside} state-of-the-art solutions for other components, from which it
\edit{simply cannot be detached.}
This is crucial, as true performance can be easily hidden by sub-optimal hyperparameters.

\begin{figure*}
\centering
\renewcommand{\arraystretch}{0.2}
\renewcommand{\tabcolsep}{0mm}
\footnotesize
\centering
\renewcommand{\imsize}{1.28cm}
\begin{tabular}{@{}cccccccccc@{}}
\includegraphics[scale=\imscl,height=\imsize]{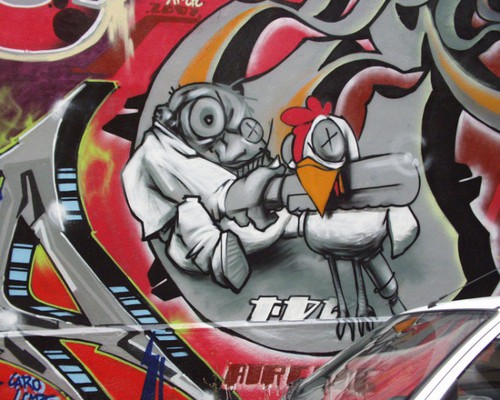}\hspace{\hspacer} &
\includegraphics[scale=\imscl,height=\imsize]{fig/other-benchmarks/new/hpatches0-fit.jpg}\hspace{\hspacer} &
\includegraphics[scale=\imscl,height=\imsize]{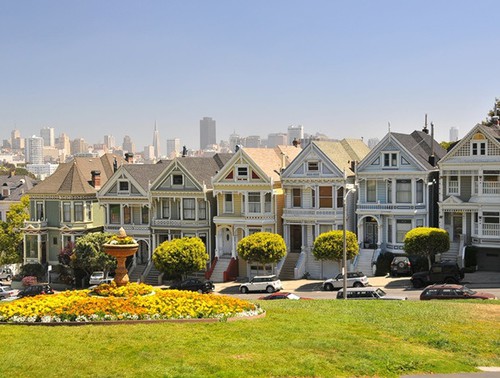}\hspace{\hspacer} &
\includegraphics[scale=\imscl,height=\imsize]{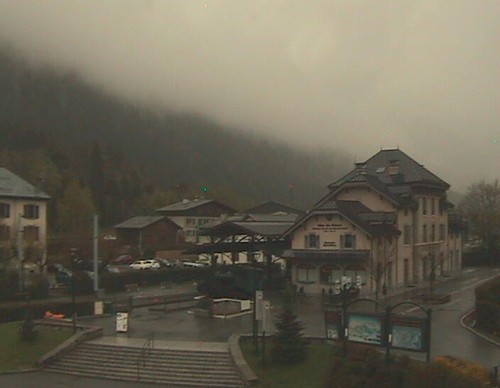}\hspace{\hspacer} &
\includegraphics[scale=\imscl,height=\imsize]{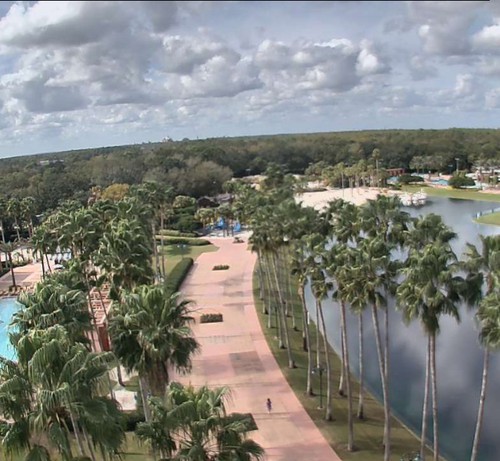}\hspace{\hspacer} &
\includegraphics[scale=\imscl,height=\imsize]{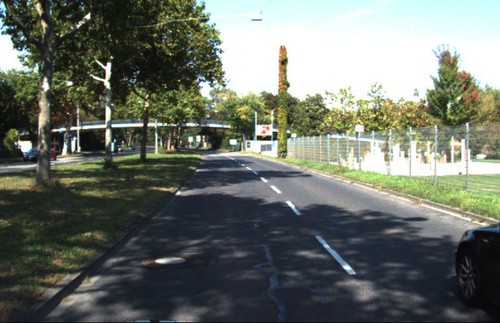}\hspace{\hspacer} &
\includegraphics[scale=\imscl,height=\imsize]{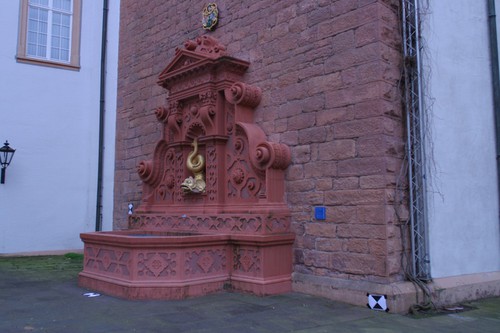}\hspace{\hspacer} &
\includegraphics[scale=\imscl,height=\imsize]{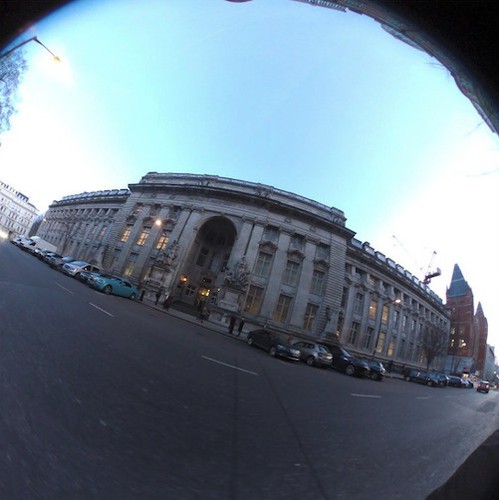}\hspace{\hspacer} &
\includegraphics[scale=\imscl,height=\imsize]{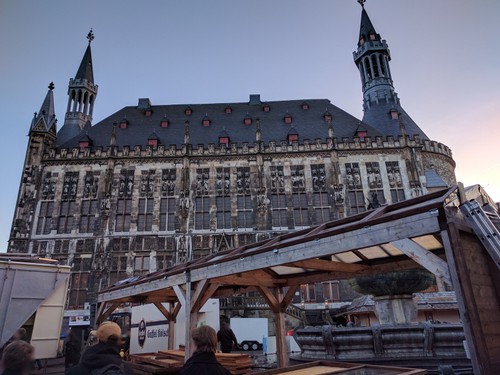}\hspace{\hspacer} &
\includegraphics[scale=\imscl,height=\imsize]{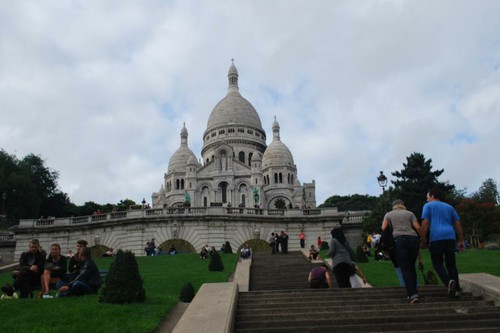}\hspace{\hspacer} \\
\includegraphics[scale=\imscl,height=\imsize]{fig/other-benchmarks/new/oxford1-fit.jpg}\hspace{\hspacer} &
\includegraphics[scale=\imscl,height=\imsize]{fig/other-benchmarks/new/hpatches1-fit.jpg}\hspace{\hspacer} &
\includegraphics[scale=\imscl,height=\imsize]{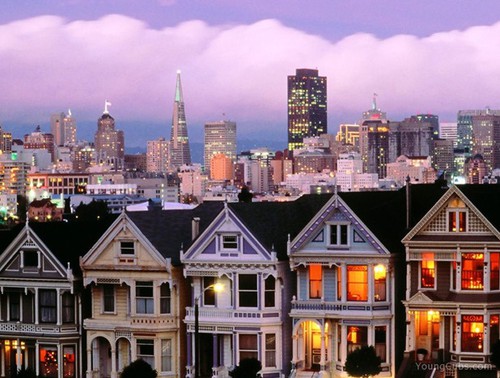}\hspace{\hspacer} &
\includegraphics[scale=\imscl,height=\imsize]{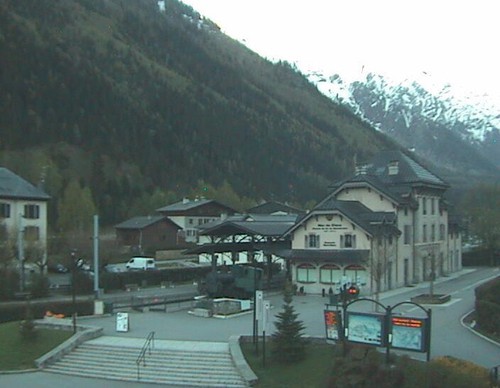}\hspace{\hspacer} &
\includegraphics[scale=\imscl,height=\imsize]{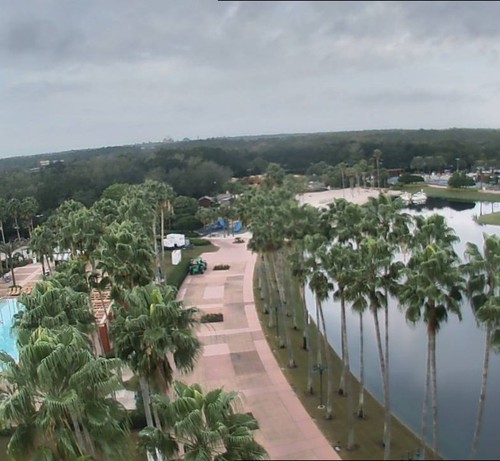}\hspace{\hspacer} &
\includegraphics[scale=\imscl,height=\imsize]{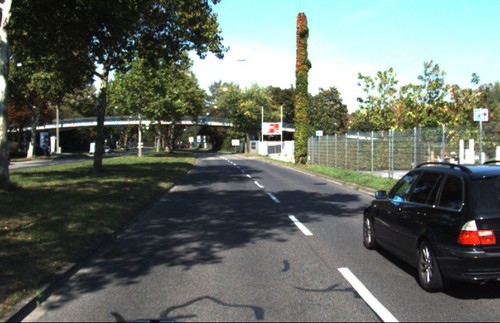}\hspace{\hspacer} &
\includegraphics[scale=\imscl,height=\imsize]{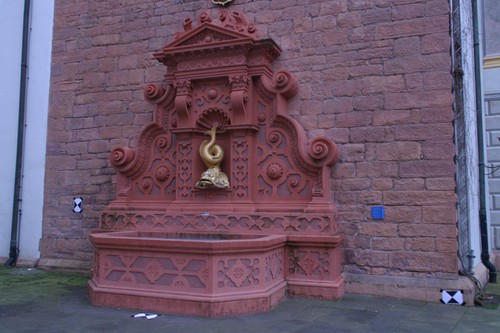}\hspace{\hspacer} &
\includegraphics[scale=\imscl,height=\imsize]{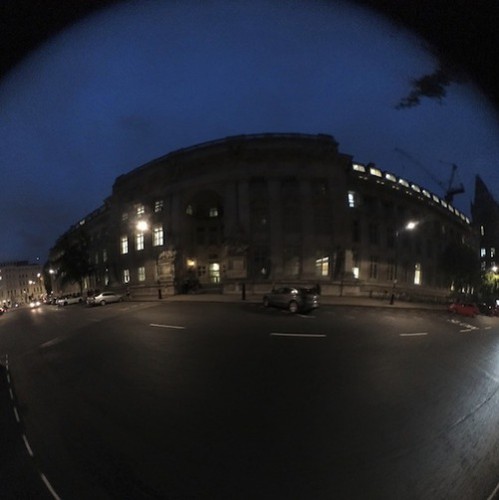}\hspace{\hspacer} &
\includegraphics[scale=\imscl,height=\imsize]{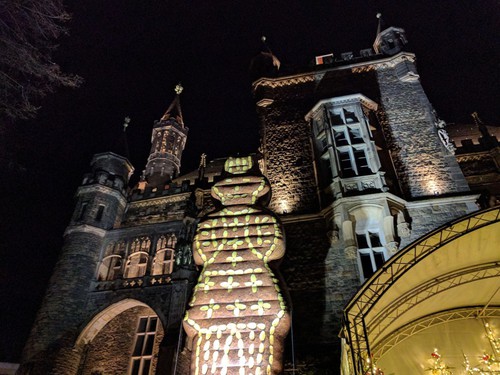}\hspace{\hspacer} &
\includegraphics[scale=\imscl,height=\imsize]{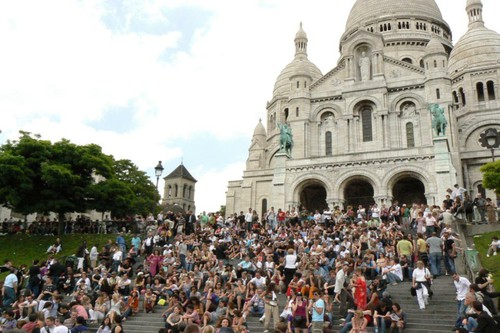}\hspace{\hspacer} \\
\includegraphics[scale=\imscl,height=\imsize]{fig/other-benchmarks/new/oxford2-fit.jpg}\hspace{\hspacer} &
\includegraphics[scale=\imscl,height=\imsize]{fig/other-benchmarks/new/hpatches2-fit.jpg}\hspace{\hspacer} &
\includegraphics[scale=\imscl,height=\imsize]{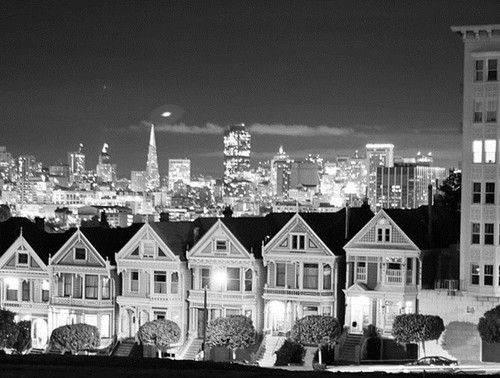}\hspace{\hspacer} &
\includegraphics[scale=\imscl,height=\imsize]{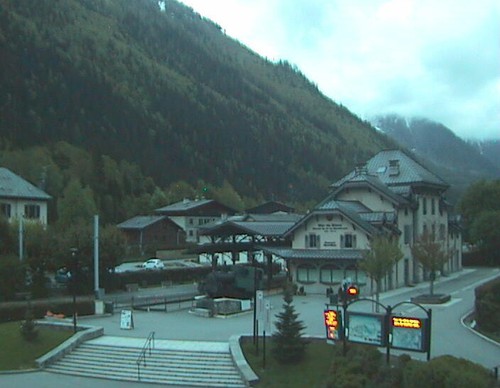}\hspace{\hspacer} &
\includegraphics[scale=\imscl,height=\imsize]{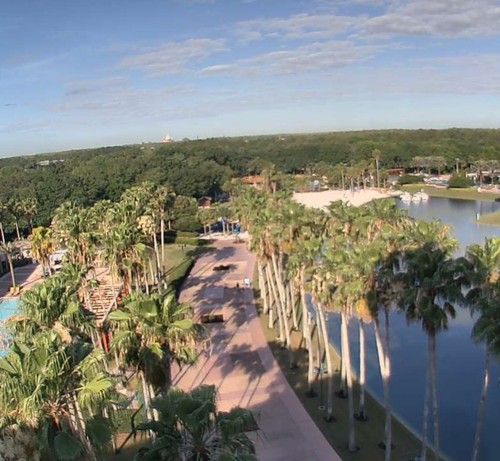}\hspace{\hspacer} &
\includegraphics[scale=\imscl,height=\imsize]{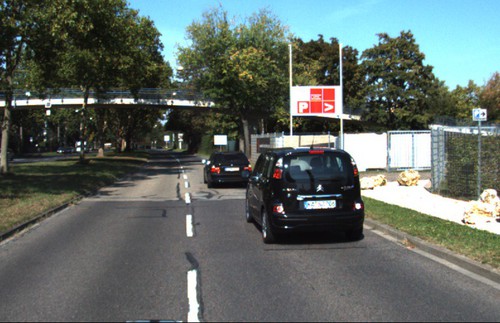}\hspace{\hspacer} &
\includegraphics[scale=\imscl,height=\imsize]{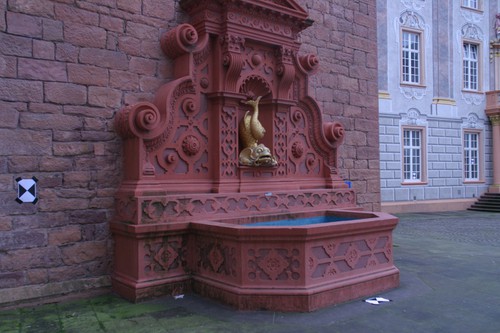}\hspace{\hspacer} &
\includegraphics[scale=\imscl,height=\imsize]{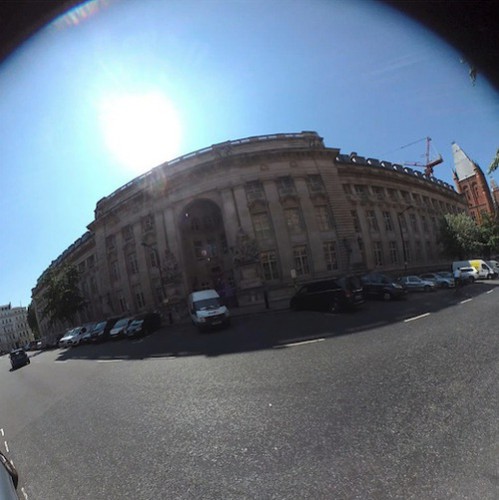}\hspace{\hspacer} &
\includegraphics[scale=\imscl,height=\imsize]{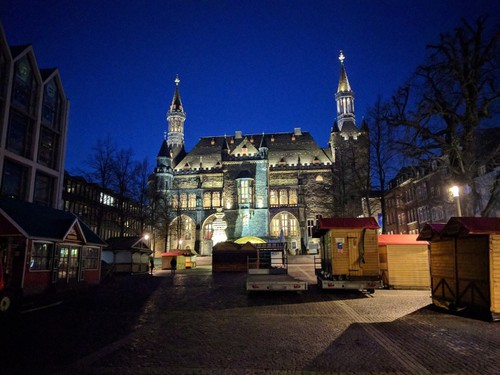}\hspace{\hspacer} &
\includegraphics[scale=\imscl,height=\imsize]{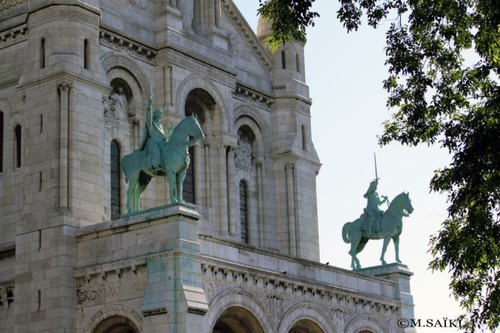}\hspace{\hspacer} \\
\vspace{1mm} \\
{\small (a)} & {\small (b)} & {\small (c)}  & {\small (d)} & {\small (e)} & {\small (f)} & {\small (g)} & {\small (h)} & {\small (i)} & {\small (j)} \\
\end{tabular}
\caption{\edit{
{\bf On the limitations of previous datasets.}
\edit{To highlight the need for a new benchmark,} we show examples from datasets and benchmarks featuring posed images that have been previously used to evaluate local features and robust matchers (some images are cropped for the sake of presentation).
From left to right: 
(a) VGG Oxford~\cite{Miko04b}, 
(b) HPatches~\cite{Balntas17}, 
(c) Edge Foci~\cite{Zitnick11}, 
(d) Webcam~\cite{Verdie15}, 
(e) AMOS~\cite{Jacobs07}, 
(f) Kitti~\cite{Geiger12}, 
(g) Strecha~\cite{Strecha08b}, 
(h) SILDa~\cite{SILDa2018}, 
(i) Aachen~\cite{Sattler12}, 
(j) Ours. 
Notice that many (a-e) contain only planar structures or illumination changes, which makes it easy -- or trivial -- to obtain ground truth poses, encoded as homographies. 
Other datasets are small -- (g) contains very accurate depth maps, but only two scenes and 19 images total -- or do not contain a wide enough range of photometric and viewpoint transformations.} \edit{Aachen (i) is closer to ours (j), but relatively small, limited to one scene, and focused on re-localization on the day vs. night use-case.}}
\label{fig:other_benchmarks}
\end{figure*}
\section{Related Work}
\label{sec:related}
The literature on image matching is too vast for a thorough overview. We cover relevant methods for feature extraction and matching, pose estimation, 3D reconstruction, applicable datasets, and evaluation frameworks.

\subsection{Local features}
Local features became a staple in computer vision with the introduction of SIFT~\cite{Lowe04}. They typically involve three distinct steps: keypoint detection, orientation estimation, and descriptor extraction. Other popular \edit{classical} solutions are SURF~\cite{Bay06}, ORB~\cite{Rublee11}, and AKAZE~\cite{Alcantarilla13}.

Modern descriptors often train deep networks on pre-cropped patches, typically from SIFT keypoints (\ie Difference of Gaussians or DoG). They include Deepdesc~\cite{Simo-Serra15}, TFeat~\cite{Balntas16b}, L2-Net~\cite{Tian17}, HardNet~\cite{Mishchuk17}, SOSNet~\cite{Tian19}, and LogPolarDesc~\cite{Ebel19} -- 
most of them are trained on the same dataset~\cite{Brown10}. Recent works leverage additional cues, such as geometry or global context, including GeoDesc~\cite{Luo18} and ContextDesc~\cite{Luo19}.
There have been multiple attempts to learn keypoint detectors separately from the descriptor, including TILDE~\cite{Verdie15}, TCDet~\cite{Zhang18bTCDet}, QuadNet~\cite{Savinov17}, and Key.Net~\cite{BarrosoLaguna2019ICCVKeyNet}.
An alternative is to treat this as an end-to-end learning problem, a trend that started with the introduction of LIFT~\cite{Yi16b} and also includes DELF~\cite{Noh17}, SuperPoint~\cite{DeTone17b}, LF-Net~\cite{Ono18}, D2-Net~\cite{Dusmanu19} and R2D2~\cite{R2D22019}.

\subsection{Robust matching}
Inlier ratios in wide-baseline stereo can be below 10\% -- and sometimes \edit{even} lower.
This is typically approached with iterative sampling schemes based on RANSAC~\cite{Fischler81}, relying on closed-form solutions for pose solving such as the 5-~\cite{Nister03}, 7-~\cite{Hartley94_7pt} or 8-point algorithm~\cite{Hartley97}.
\edit{Improvements to this classical framework include local optimization~\cite{Chum03}, using likelihood instead of reprojection (MLESAC)~\cite{Torr00}, speed-ups using probabilistic sampling of hypotheses (PROSAC)~\cite{Chum05a}, degeneracy check using homographies (DEGENSAC)~\cite{Degensac2005},
Graph Cut as a local optimization (GC-RANSAC)~\cite{gcransac2018}, and auto-tuning of thresholds using confidence margins (MAGSAC)~\cite{magsac2019}.

\edit{As an alternative} 
direction, recent works, starting with CNe (Context Networks) in~\cite{Yi18a}, train deep networks for outlier rejection taking correspondences as input, often followed by a RANSAC loop. 
Follow-ups include~\cite{Ranftl18,Zhao19,Sun19,Zhang19}.}
\edit{Differently from RANSAC solutions, they typically process all correspondences in a single forward pass, without the need to \edit{iteratively} sample hypotheses.}
Despite their promise, it remains unclear how well they perform in real\edit{-world} settings, \edit{compared to a well-tuned RANSAC}.
\subsection{Structure from Motion (SfM)}
In \edit{SfM,} one jointly optimizes the location of the 3D points and the camera intrinsics and extrinsics.
Many improvements have been proposed over the years~\cite{Agarwal09,Heinly15,Cui_2017_CVPR_HSFM,PSfMO2017,Zhu_2018_CVPR}.
The most popular frameworks are VisualSFM~\cite{Wu13} and COLMAP~\cite{Schoenberger16a} --
we rely on the latter, to \edit{both} generate the ground truth and as the backbone of our multi-view task.
\begin{figure*}
\centering
\renewcommand{\imsize}{1.82cm}
\renewcommand{\hspacer}{0.3mm}
\renewcommand{\vspacer}{0.1mm}
\renewcommand{\imscl}{0.01}
\includegraphics[scale=\imscl,height=\imsize]{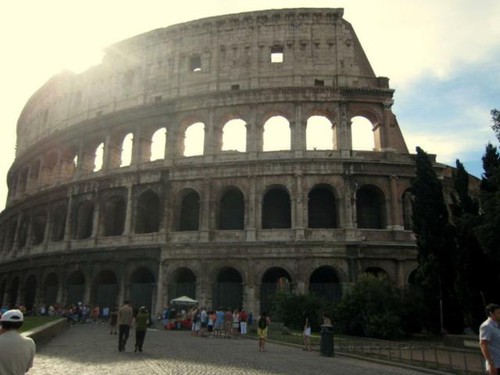}\hspace{\hspacer}%
\includegraphics[scale=\imscl,height=\imsize]{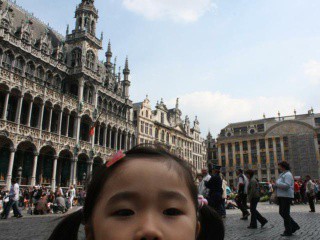}\hspace{\hspacer}%
\includegraphics[scale=\imscl,height=\imsize]{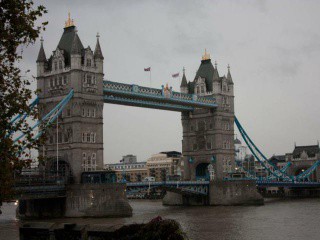}\hspace{\hspacer}%
\includegraphics[scale=\imscl,height=\imsize]{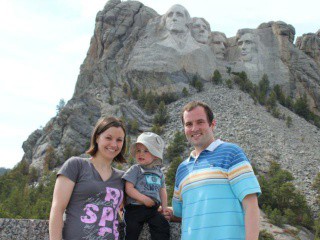}\hspace{\hspacer}%
\includegraphics[scale=\imscl,height=\imsize]{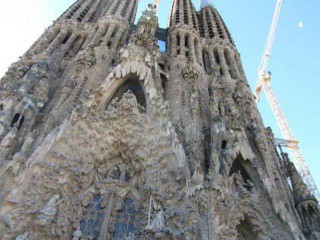}\hspace{\hspacer}%
\includegraphics[scale=\imscl,height=\imsize]{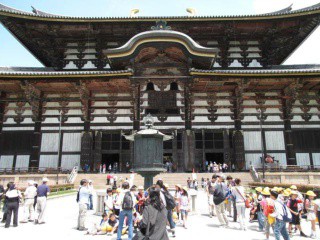}\hspace{\hspacer}%
\includegraphics[scale=\imscl,height=\imsize]{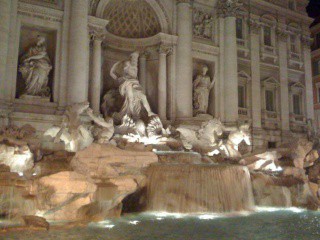}\hspace{\hspacer}%
\includegraphics[scale=\imscl,height=\imsize]{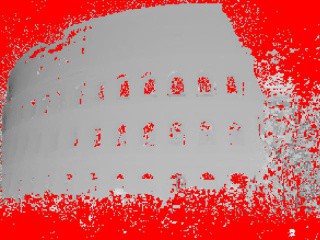}\hspace{\hspacer}%
\includegraphics[scale=\imscl,height=\imsize]{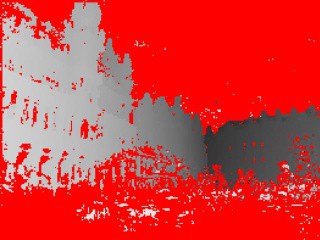}\hspace{\hspacer}%
\includegraphics[scale=\imscl,height=\imsize]{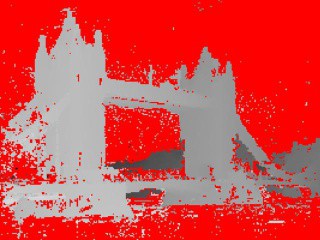}\hspace{\hspacer}%
\includegraphics[scale=\imscl,height=\imsize]{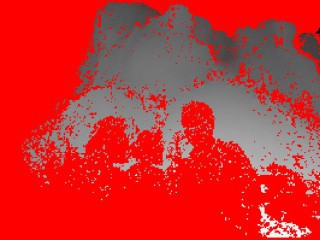}\hspace{\hspacer}%
\includegraphics[scale=\imscl,height=\imsize]{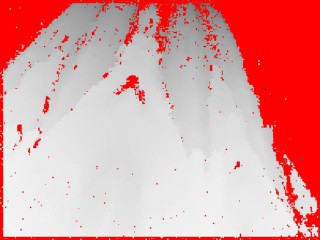}\hspace{\hspacer}%
\includegraphics[scale=\imscl,height=\imsize]{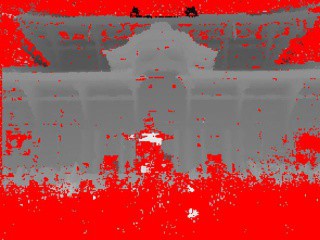}\hspace{\hspacer}%
\includegraphics[scale=\imscl,height=\imsize]{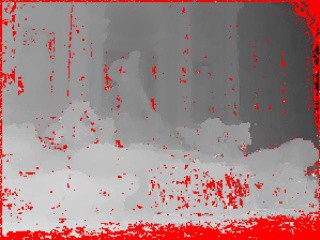}\hspace{\hspacer}%
\caption{
{\bf The Image Matching Challenge PhotoTourism (IMC-PT) dataset.}
We show a few selected images from our dataset and their corresponding depth maps, with occluded pixels marked in red. 
}
\label{fig:dataset_samples}
\vspace{-2mm}
\end{figure*}
\subsection{Datasets and benchmarks}
Early works \edit{on local features and robust matchers typically} relied on the Oxford dataset~\cite{Miko04b}, \edit{which contains} 48 images and ground truth homographies. 
It helped establish two common metrics \edit{\edit{for evaluating} local feature performance}: repeatability and matching score.
Repeatability evaluates the keypoint detector: given \edit{two} sets of keypoints over two images, projected into each other, it is defined as the ratio of keypoints whose support regions' \edit{overlap} is above a threshold.
The matching score (MS) is similarly defined, but also requires their descriptors to be nearest neighbours.
Both require pixel-to-pixel correspondences -- \ie, features outside valid areas are ignored.

A modern alternative to Oxford is HPatches~\cite{Balntas17}, which contains 696 images with differences in illumination {\em or} viewpoint -- \edit{but not both}. However, the scenes are planar, without occlusions, \edit{limiting its applicability}.

Other datasets 
\edit{that have been used to evaluate local features} 
include DTU~\cite{Aanaes12}, Edge Foci~\cite{Zitnick11}, Webcam~\cite{Verdie15}, AMOS~\cite{Pultar19}, and Strecha's~\cite{Strecha08b}. 
They all have limitations -- \edit{be it} narrow baselines, noisy ground truth, or a small number of images.
\edit{In fact, most learned descriptors have been trained and evaluated on~\cite{Brown07}, which provides a database of pre-cropped patches with correspondence labels, and measures performance in terms of patch retrieval. 
While this seminal dataset and evaluation methodology helped developed many new methods, it is not clear how results translate to different scenarios -- particularly since new methods outperform classical ones such as SIFT by orders of magnitude, which suggests overfitting.}

Datasets used for navigation, re-localization, or SLAM, in outdoor environments are also \edit{relevant to our problem.
These include} KITTI~\cite{Geiger12}, Aachen~\cite{Sattler12b}, Robotcar~\cite{Maddern17}, and CMU seasons~\cite{Sattler18,Badino11}. 
However, they do not feature the wide range of transformations present in phototourism data.
\edit{Megadepth~\cite{Li18f} is a more representative, phototourism-based dataset,
which using COLMAP, as in our case}
-- \edit{it could, in fact, easily be folded into ours.}

Benchmarks, by contrast, are few and far between -- they include VLBenchmark~\cite{Lenc11}, HPatches~\cite{Balntas17}, and SILDa~\cite{SILDa2018} -- all limited in scope.
A large-scale benchmark for SfM was proposed in~\cite{Schonberger17},
\edit{which built 3D reconstructions with different local features. However, only a few scenes contain ground truth, so most of their metrics are qualitative -- \eg number of observations or average reprojection error.}
Yi \etal~\cite{Yi18a} and Bian \etal~\cite{Bian19}
evaluate different methods for pose estimation on several datasets -- however, few methods are considered and they are not carefully tuned. 

We highlight some of these datasets/benchmarks, and their limitations, in \fig{other_benchmarks}.
We are, to the best of our knowledge, the first to introduce a public, modular benchmark for 3D reconstruction with sparse methods
using downstream metrics, and featuring a comprehensive dataset with a large range of image transformations.

\section{\revision{The~Image~Matching~Challenge~PhotoTourism~Dataset}}
\label{sec:data}
While it is possible to obtain very accurate poses and depth maps under controlled scenarios with devices like LIDAR, this is costly and requires a specific set-up that
does not scale well.
For example, Strecha's dataset~\cite{Strecha08b}
follows that approach but contains only 19 images.
We argue that a truly representative dataset must contain a wider range of transformations -- including different imaging devices, time of day, weather, partial occlusions, etc.
Phototourism images satisfy this condition
and are readily available.

We thus build on \edit{25} collections of popular landmarks originally selected in \cite{Heinly15,Thomee16}, each with hundreds to thousands of images.
\revision{Images are downsampled with bilinear interpolation to a maximum size of 1024 pixels along the long-side and their poses were obtained with COLMAP \cite{Schoenberger16a}, which provides the (pseudo) ground truth}.
We do exhaustive image matching before Bundle Adjustment -- unlike~\cite{Schoenberger16b}, which uses only 100 pairs for each image -- and thus provide enough matching images for any conventional SfM to return near-perfect results \revision{in standard conditions}.

\edit{%
\revision{Our approach is to obtain a ground truth signal using reliable, off-the-shelf technologies, while making the problem as easy as possible -- and then evaluate new technologies on a much harder problem, using only a subset of that data.}
\edit{For example, we reconstruct a scene with hundreds or thousands of images with vanilla COLMAP and then evaluate ``modern'' features and matchers against its poses using only two images (``stereo'') or up to 25 at a time (``multiview'' with SfM).}
For a discussion regarding the accuracy of our ground truth data, please refer to \secref{eval_gt}.}

In addition to point clouds, COLMAP provides dense depth estimates. 
\revision{These are noisy, and have no notion of occlusions -- a depth value is provided for every pixel.}
We remove occluded pixels \revision{from depth maps} using the reconstructed model \edit{from COLMAP}; see~\fig{dataset_samples} for examples.
We rely on these ``cleaned'' depth maps to compute \edit{classical,} pixel-wise metrics -- repeatability and matching score.
We find that some images are flipped 90$^\circ$, and use the \edit{reconstructed pose} to rotate them -- \edit{along with their poses} -- so they are roughly `upright', \edit{which is a reasonable assumption for this type of data.}

\subsection{Dataset details}
\begin{table}%
\renewcommand{\arraystretch}{1}
\renewcommand{\tabcolsep}{7pt}
\small
\begin{center}
\begin{tabular}{@{}lcrr@{}}
\toprule
Name & Group & Images & 3D points \\
\midrule
``Brandenburg Gate'' & T & 1363 & 100040 \\
``Buckingham Palace'' & T & 1676 & 234052 \\
``Colosseum Exterior'' & T & 2063 & 259807 \\
``Grand Place Brussels''& T & 1083 & 229788 \\
``Hagia Sophia Interior''& T & 888 & 235541 \\
``Notre Dame Front Facade'' &T & 3765 & 488895 \\
``Palace of Westminster''  &T & 983 & 115868 \\
``Pantheon Exterior'' & T& 1401 & 166923 \\
``Prague Old Town Square'' & T& 2316 & 558600 \\
``Reichstag'' & T& 75 & 17823 \\
``Taj Mahal'' &T & 1312 & 94121 \\
``Temple Nara Japan'' &T & 904 & 92131 \\
``Trevi Fountain'' & T & 3191 & 580673 \\
``Westminster Abbey'' & T & 1061 & 198222 \\
``Sacre Coeur'' (SC) & V& 1179 & 140659 \\
``St. Peter's Square'' (SPS) & V & 2504 & 232329 \\
\midrule
Total T + V & & 25.7k & 3.6M \\
\midrule
``British Museum'' (BM) & E & 660 & 73569 \\
``Florence Cathedral Side'' (FCS) & E & 108 & 44143 \\
``Lincoln Memorial Statue'' (LMS) &  E &850 & 58661 \\
``London (Tower) Bridge'' (LB) &  E &629 & 72235 \\
``Milan Cathedral'' (MC) &  E &124 & 33905 \\
``Mount Rushmore'' (MR) &  E &138 & 45350 \\
``Piazza San Marco'' (PSM) &  E &249 & 95895 \\
``Sagrada Familia'' (SF) &  E &401 & 120723 \\
``St. Paul's Cathedral'' (SPC) &  E &615 & 98872 \\
\midrule
Total E & & 3774 & 643k \\
\bottomrule
\end{tabular}
\end{center}
\caption{
\revision{{\bf Scenes in the IMC-PT dataset.} 
We provide some statistics for the training (T), validation (V), and test (E) scenes.}}
\label{tbl:sequences_trainval}
\end{table}
\edit{%
Out of the 25 scenes, containing almost 30k registered images in total, we select 2 for validation and 9 for testing. \revision{The remaining scenes can be used for training, if so desired, and are not used in this paper}.
We provide images, 3D reconstructions, camera poses, and depth maps, for every training and validation scene. 
For the test scenes we release only a subset of 100 images and keep the ground truth private, \revision{which is an integral part of the Image Matching Challenge. Results on the private test set can be obtained by sending submissions to the organizers, who process them.}

\revision{The scenes used for training, validation and test are listed in \tbl{sequences_trainval}},
along with the acronyms used in several figures.
For the validation experiments
-- Sections \ref{sec:ablation} and \ref{sec:appendix} -- 
we choose
two of the larger scenes, ``Sacre Coeur'' and ``St. Peter's Square'', which in our experience 
\edit{provide results that are quite representative of what should be expected on \revision{the Image Matching Challenge Dataset}.}
These two subsets have been released, so that the validation results are reproducible and comparable.
They can all be downloaded from the challenge website\footnoteref{website}.
}
\begin{figure}%
\centering
\renewcommand{\imscl}{0.01}
\includegraphics[scale=\imscl,width=0.8\linewidth]{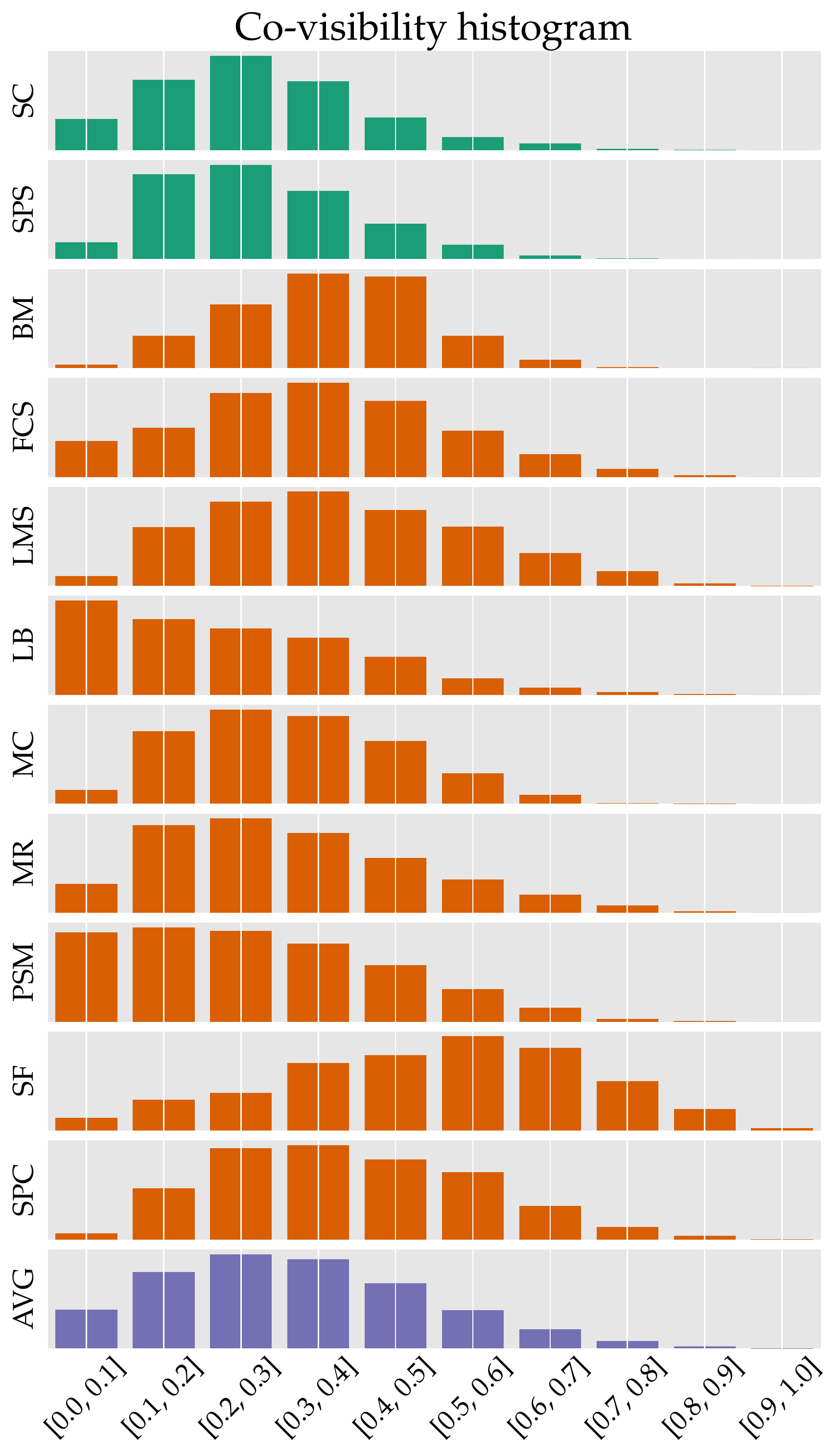}
\caption{
\edit{{\bf Co-visibility histogram.} 
We break down the co-visibility measure for each scene in the validation (green) and test (red) sets, as well as the average (purple). 
Notice how the statistics may vary significantly from scene to scene.}
}
\label{fig:visibility_histograms}
\end{figure}

\begin{figure*}
\centering
\renewcommand{\imsize}{7.3cm}
\renewcommand{\hspacer}{0.3mm}
\renewcommand{\vspacer}{0.1mm}
\renewcommand{\imscl}{0.01}
\includegraphics[scale=\imscl,height=\imsize]{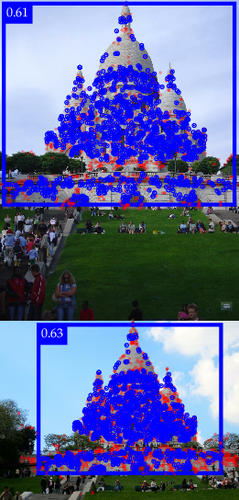}\hspace{\hspacer}%
\includegraphics[scale=\imscl,height=\imsize]{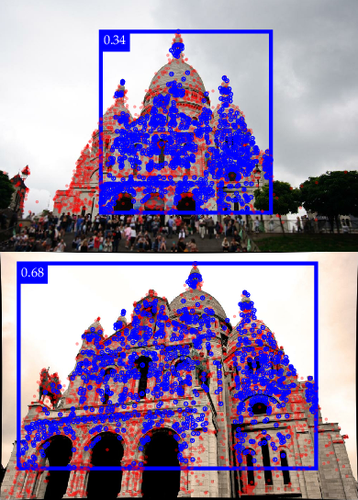}\hspace{\hspacer}%
\includegraphics[scale=\imscl,height=\imsize]{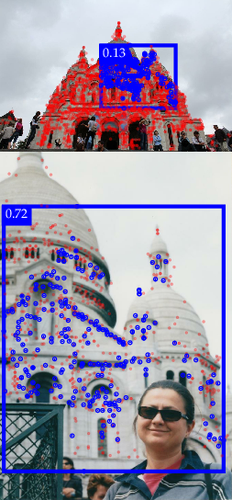}\hspace{\hspacer}%
\includegraphics[scale=\imscl,height=\imsize]{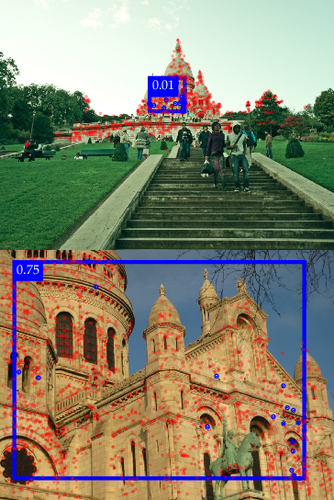}\hspace{\hspacer}\\
\vspace{\vspacer}%
\vspace{1ex}%
\renewcommand{\imsize}{5.95cm}%
\includegraphics[scale=\imscl,height=\imsize]{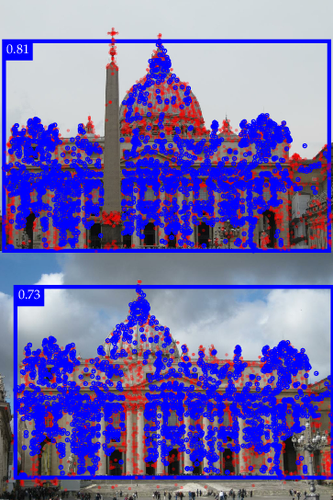}\hspace{\hspacer}%
\includegraphics[scale=\imscl,height=\imsize]{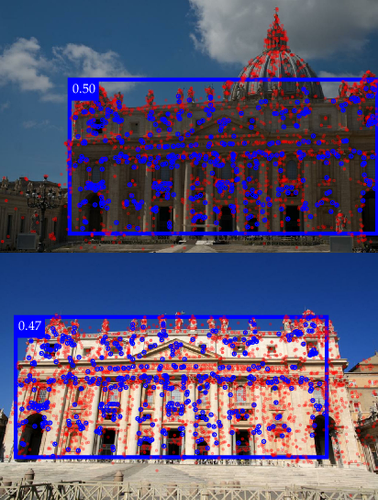}\hspace{\hspacer}%
\includegraphics[scale=\imscl,height=\imsize]{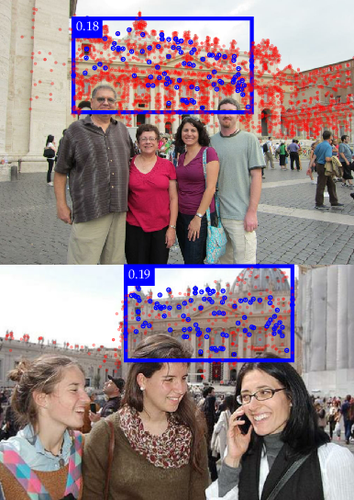}\hspace{\hspacer}%
\includegraphics[scale=\imscl,height=\imsize]{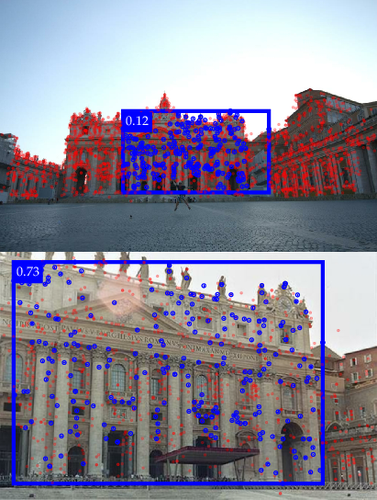}\hspace{\hspacer}%
\caption{{\bf Co-visibility examples }-- image pairs from validation scenes ``Sacre Coeur'' (top) and ``St. Peter's Square'' (bottom). Image keypoints that are part of the 3D reconstruction are blue if they are co-visible in both images, are red otherwise. The bounding box of the co-visible points, used to compute a per-image co-visibility ratio, is shown in blue. The co-visibility value for the image pair is the lower of these two values.
Examples include different `difficulty' levels. All of the pairs are used in the evaluation except the top-right one, as we set a cut-off at 0.1.
}
\label{fig:visibility_examples}
\vspace{-2mm}
\end{figure*}
\subsection{\edit{Estimating the co-visibility between two images}}%
\label{sec:covis}
\edit{%
\edit{When evaluating image pairs, one has to be sure that two given images share a common part of the scene} --
they may be registered by the SfM reconstruction without having any pixels in common as long as other images act as a `bridge' between them.}
Co-visible pairs of images are determined with a simple heuristic.
3D model points, detected in both the considered images, are localised in 2D in each image. The bounding box around the keypoints is estimated. The ratio of the bounding box area and the whole image is the ``visibility'' of image $i$ in $j$ and $j$ in $i$ respectively;  \edit{see~\fig{visibility_examples} for examples}.
The minimum of the two ``visibilities'' is the ``co-visibility'' ratio  $\covisibility_{i,j} \in [0,1]$, which characterizes how challenging matching of the particular image pair is.
The co-visibility varies significantly from scene to scene. The histogram of co-visibilities is shown in \fig{visibility_histograms}, providing insights into
how ``hard'' each scene is -- without accounting for some occlusions.

For stereo, the minimum co-visibility threshold is set to 0.1. 
For the multi-view task, subsets where at least 100 3D points are visible in each image, are selected, as in~\cite{Yi18a,Zhang19}.
We find that both criteria work well in practice.

\subsection{On the quality of the ``ground-truth''}
\label{sec:eval_gt}

Our core assumption is that accurate poses can be obtained from large sets of images without human intervention. Such poses are used as the ``ground truth'' for evaluation of image matching performance on pairs or small subsets of images -- a harder, proxy task.
Should this assumption hold, the (relative) poses retrieved with a large enough number of images would not change as more images are added, and these poses would be the same regardless of which local feature is used.
\begin{table}
\begin{center}
\scriptsize
\setlength\tabcolsep{4pt} %
\begin{tabular}{l ccccc}
\toprule
\multirow{2}{*}{Local featured type} & \multicolumn{4}{c}{Number of Images}\\
\cmidrule{2-6}
& 100 & 200 & 400 & 800 & all \\
\midrule
SIFT~\cite{Lowe04}&0.06$^\circ$&0.09$^\circ$&0.06$^\circ$&0.07$^\circ$&0.09$^\circ$\\
SIFT (Upright)~\cite{Lowe04}& 0.07$^\circ$& 0.07$^\circ$& 0.04$^\circ$& 0.06$^\circ$& 0.09$^\circ$\\
HardNet (Upright)~\cite{Mishchuk17}& 0.06$^\circ$& 0.06$^\circ$& 0.06$^\circ$& 0.04$^\circ$& 0.05$^\circ$\\
SuperPoint~\cite{DeTone17b}& 0.31$^\circ$& 0.25$^\circ$& 0.33$^\circ$& 0.19$^\circ$& 0.32$^\circ$\\
R2D2~\cite{Revaud19}& 0.12$^\circ$& 0.08$^\circ$& 0.07$^\circ$& 0.08$^\circ$& 0.05$^\circ$\\
\bottomrule
\end{tabular}
\end{center}
\caption{
\revision{{\bf Standard deviation of the pose difference of three COLMAP runs with different number of images}. Most of them are below 0.1$^\circ$, except for SuperPoint.}
}
\label{tbl:colmap_std}
\end{table}
\begin{table}
\begin{center}
\scriptsize
\setlength\tabcolsep{2pt} %
\begin{tabular}{lcccc}
\toprule
\multirow{2}{*}{Local feature type} & \multicolumn{4}{c}{Number of images}\\
\cmidrule{2-5}
& 100 vs. all & 200 vs. all & 400 vs. all & 800 vs. all \\
\midrule
SIFT~\cite{Lowe04} & 0.58$^\circ$ / 0.22$^\circ$ & 0.31$^\circ$ / 0.08$^\circ$ & 0.23$^\circ$ / 0.05$^\circ$ & 0.18$^\circ$ / 0.04$^\circ$\\
SIFT (Upright)~\cite{Lowe04} & 0.52$^\circ$ / 0.16$^\circ$ & 0.29$^\circ$ / 0.08$^\circ$ & 0.22$^\circ$ / 0.05$^\circ$ & 0.16$^\circ$ / 0.03$^\circ$\\
HardNet (Upright)~\cite{Mishchuk17} & 0.35$^\circ$ / 0.10$^\circ$ & 0.33$^\circ$ / 0.08$^\circ$ & 0.23$^\circ$ / 0.06$^\circ$ & 0.14$^\circ$ / 0.04$^\circ$\\
SuperPoint~\cite{DeTone17b} & 1.22$^\circ$ / 0.71$^\circ$ & 1.11$^\circ$ / 0.67$^\circ$ & 1.08$^\circ$ / 0.48$^\circ$ & 0.74$^\circ$ / 0.38$^\circ$\\
R2D2~\cite{Revaud19} & 0.49$^\circ$ / 0.14$^\circ$ & 0.32$^\circ$ / 0.10$^\circ$ & 0.25$^\circ$ / 0.08$^\circ$ & 0.18$^\circ$ / 0.05$^\circ$\\
\bottomrule
\end{tabular}
\end{center}
\caption{
\edit{{\bf \edit{Pose convergence in SfM.}}
\edit{We report the mean/median of the difference (in degrees) between the poses extracted with the full set of 1179 images for ``Sacre Coeur'', and different subsets of it, for \revision{four} local feature methods -- to keep the results comparable we only look at the 100 images in common across all subsets. 
We report the maximum among the angular difference between rotation matrices and translation vectors. 
The estimated poses are stable, with as little as 100 images.}
}}
\label{tbl:conv_gt}
\end{table}
\begin{table}
\begin{center}
\scriptsize
\setlength\tabcolsep{3pt} %
\begin{tabular}{c cccc}
\toprule
\multirow{2}{*}{Reference} & \multicolumn{4}{c}{Compared to}\\
\cmidrule{2-5}
& SIFT (Upright) & HardNet (Upright) & SuperPoint & R2D2 \\
\midrule
SIFT~\cite{Lowe04} & 0.20$^\circ$ / 0.05$^\circ$ & 0.26$^\circ$ / 0.05$^\circ$ & 1.01$^\circ$ / 0.62$^\circ$ & 0.26$^\circ$ / 0.09$^\circ$ \\
\bottomrule
\end{tabular}
\end{center}
\caption{
{\bf Difference between poses obtained with different local features.}
We report the mean/median of the difference (in degrees) between the poses extracted with SIFT (Upright), HardNet (Upright), SuperPoint, or R2D2, and those extracted with SIFT.
We use the maximum of the angular difference between rotation matrices and translation vectors.
SIFT (Upright), HardNet (Upright), and R2D2 give near-identical results to SIFT.}
\label{tbl:val_gt}
\end{table}
To validate this, we pick the scene ``Sacre Coeur'' and compute SfM reconstructions with a varying number of images: 100, 200, 400, 800, and 1179 images (the entire ``Sacre Coeur'' dataset), where each set contains the previous one; new images are being added and no images are removed. 
\revision{We run each reconstruction three times, and report the average result of the three runs, to account for the variance inside COLMAP. The standard deviation among different runs is reported in \tbl{colmap_std}.}
Note that these reconstructions are only defined up to a scale factor -- we do not know the absolute scale that could be used to compare the reconstructions against each other. That is why we use a simple, pairwise metric instead. We 
\edit{pick all the pairs out of the 100 images present in the smallest subset, and compare how much their relative pose change with respect to their counterparts reconstructed using the entire set -- we do this for every subset, \ie, 100, 200, etc.}
\edit{Ideally, we would like the differences between the relative poses to approach zero as more images are added.}
\edit{We list the results in \tbl{conv_gt}, for different local feature methods.}
\edit{Notice how the poses converge, especially in terms of the median, as more images are used, for all methods -- and that the reconstructions using only 100 images are already very stable.
For SuperPoint we use a smaller number of features (2k per image), which is not enough to achieve pose convergence, but the error is still reduced as more images are used.}

\begin{figure}
\centering
\renewcommand{\arraystretch}{0.2}
\renewcommand{\tabcolsep}{1mm}
\footnotesize
\centering
\renewcommand{\imsize}{1.28cm}
\begin{tabular}{@{}ccc@{}}
\includegraphics[scale=\imscl,width=0.33\linewidth,height=0.5\linewidth]{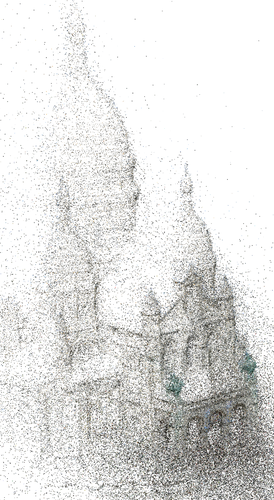} &
\includegraphics[scale=\imscl,width=0.33\linewidth,height=0.5\linewidth]{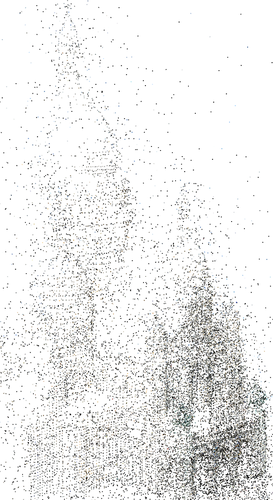} &
\includegraphics[scale=\imscl,width=0.33\linewidth,height=0.5\linewidth]{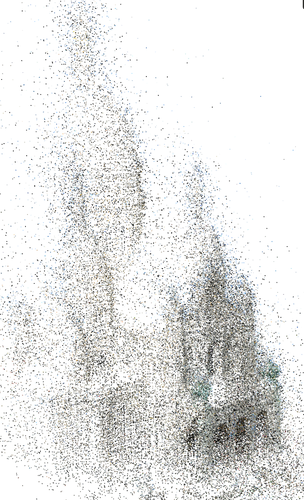} \\
\vspace{1mm} \\
(a) SIFT & (b) SuperPoint & (c) R2D2 \\
\end{tabular}
\caption{\edit{{\bf COLMAP with different local features.} We show the reconstructed point cloud for the scene ``Sacre Coeur'' using three different local features: SIFT, SuperPoint, and R2D2, using all images available (1179). The reconstructions with SIFT and R2D2 are both dense, albeit somewhat different. The reconstruction with SuperPoint is quite dense, considering it can only extract a much smaller number of features effectively, but its poses appear less accurate}.}
\label{fig:pointcloud}
\end{figure}

\begin{figure*}[bth]
\centering
\renewcommand{\imsize}{3.8cm}
\renewcommand{\hspacer}{0.3mm}
\renewcommand{\vspacer}{0.1mm}
\renewcommand{\imscl}{0.01}
\includegraphics[scale=\imscl,height=\imsize]{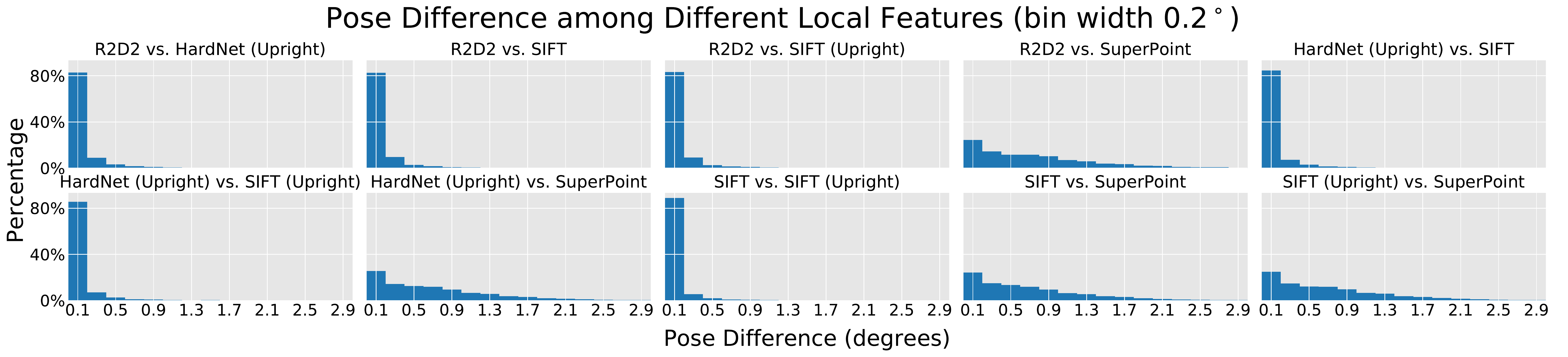}
\caption{\revision{
{\bf Histograms of pose differences between reconstructions with different local feature methods.}
We consider five different local features -- including rotation-sensitive and upright SIFT -- resulting in 10 combinations. The plots show that about 80\% percent of image pairs are within a 0.2$^o$ pose difference, with the exception of those involving SuperPoint. 
}}
\label{fig:colmap_hist}
\end{figure*}

\edit{%
We conduct a second experiment to verify that there is no bias towards using SIFT features for obtaining the ground truth.
We compare the poses obtained with SIFT to those obtained with other} local features -- note that our primary metric uses nothing but the {\em estimated poses} for evaluation.
We report results in \tbl{val_gt}. \revision{We also show the histogram of pose differences in ~\fig{colmap_hist}}.
The differences in pose %
\revision{due to the use of different local features have a median value below 0.1$^\circ$.
In fact, the pose variation of individual reconstructions with SuperPoint is of the same magnitude as the difference between reconstructions from SuperPoint and other local features: see \tbl{conv_gt}.
We conjecture that the reconstructions with SuperPoint, which does not extract a large number of keypoints, are less accurate and stable.}
This is further supported by the fact that the point cloud obtained with the entire scene generated with SuperPoint is less dense (125K 3D points) than the ones generated with SIFT (438K) or R2D2 (317k); see~\fig{pointcloud}.
\revision{In addition, we note that SuperPoint keypoints have been shown to be less accurate when it comes to precise alignment~\cite[Table~4, $\epsilon=1$]{DeTone17b}}. 
\edit{Note also that the poses from R2D2 are nearly identical to those from SIFT.}

\begin{figure*}[bthp]
\centering
\includegraphics[width=0.8\linewidth, trim= 80 250 170 15, clip]{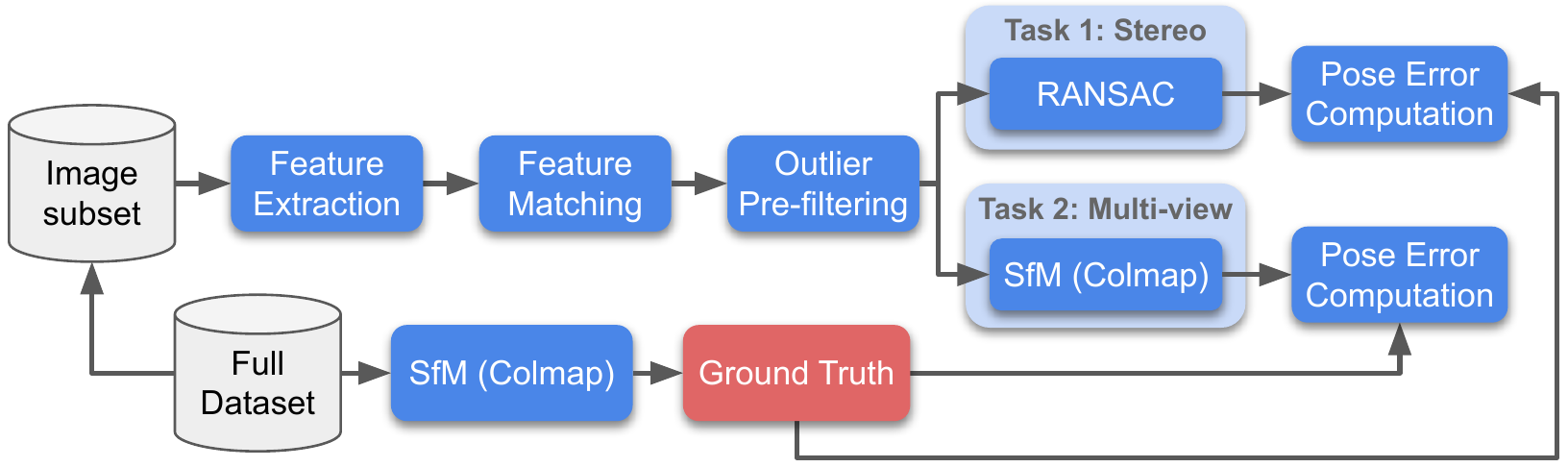}
\caption{
{\bf The benchmark pipeline} takes \edit{a subset of} $N$ images of a scene as input, extracts features for each,
and computes matches for all $M$ image pairs, $M=\frac{1}{2}N(N-1)$.
After an optional filtering step, the matches are fed to two different tasks.
Performance is measured {\em downstream}, by a
pose-based metric, common across tasks. \edit{The ground truth is extracted once, on the full set of images.}
}
\label{fig:pipeline}
\vspace{-3mm}
\end{figure*}

These observations reinforce
\edit{our trust}
in the accuracy of
\edit{our ground truth --
given a sufficient number of images, the choice of local feature is irrelevant, at least for the purpose of retrieving accurate poses.}
Our evaluation considers pose errors of up to 10$^\circ$, at a resolution of 1$^\circ$ -- significantly smaller than the fluctuations observed here, which we consider negligible.
Note that these conclusions may not hold on large-scale SfM requiring loop closures, but our dataset contains landmarks, which do not suffer from this problem.

In addition, we note that the use of dense, ground truth depth from SfM, \revision{which is arguably less accurate that camera poses}, has been verified by multiple parties, for training and evaluation, including:
CNe~\cite{Yi18a}, DFE~\cite{Ranftl18}, LF-Net~\cite{Ono18}, D2-Net~\cite{Dusmanu19}, LogPolarDesc~\cite{Ebel19}, OANet~\cite{zhang2019oanet}, and SuperGlue~\cite{sarlin20superglue} among others, suggesting it is sufficiently accurate -- several of these rely on the data used in our paper.

\edit{As a final observation, while the poses are {\em stable}, they could still be {\em incorrect}. 
This can happen on highly symmetric structures: for instance, a tower with a square or circular cross section. 
In order to prevent such errors from creeping into our evaluation, we visually inspected all the images in our test set. 
Out of 900 of them, we found 4 misregistered samples, all of them from the same scene, ``London Bridge'', which were removed from our data.
}

\section{Pipeline}
\label{sec:benchmark}

We outline our pipeline in \fig{pipeline}.
It takes as input $\numim{=}100$ images per scene.
The feature extraction module computes up to $\numfeat$ features from each image.
The
feature matching
module generates a list of putative matches for each image pair, \ie{} $\frac{1}{2}N(N-1)=4950$ combinations.
These matches can be optionally processed by an
outlier pre-filtering
module.
They are then fed to two tasks:
stereo,
and
multiview reconstruction with SfM.
We now describe each of these components in detail.

\subsection{Feature extraction}
We consider three broad families of local features.
The first includes %
full, ``classical'' pipelines, most of them handcrafted:
SIFT~\cite{Lowe04} (and RootSIFT~\cite{rootSIFT2012}), SURF~\cite{Bay06}, ORB~\cite{Rublee11}, and AKAZE~\cite{Alcantarilla13}.
We also consider FREAK~\cite{Alahi12} descriptors with BRISK~\cite{Brisk2011} keypoints. We take these from OpenCV.
For all of them, except ORB, we lower the detection threshold to extract more features, which increases performance \edit{when operating with a large feature budget}.
We also consider DoG alternatives from VLFeat \cite{vedaldi2010vlfeat}: 
(VL-)DoG, Hessian~\cite{Hessian78}, Hessian-Laplace~\cite{Mikolajczyk04}, Harris-Laplace~\cite{Mikolajczyk04}, MSER~\cite{Matas04}; and their affine-covariant versions: DoG-Affine, Hessian-Affine \cite{Mikolajczyk04,Baumberg00}, DoG-AffNet \cite{Mishkin18}, and Hessian-AffNet \cite{Mishkin18}.

The second group includes descriptors learned on DoG keypoints: L2-Net~\cite{Tian17},
Hardnet~\cite{Mishchuk17}, Geodesc~\cite{Luo18}, SOSNet~\cite{Tian19}, ContextDesc~\cite{Luo19}, and LogPolarDesc~\cite{Ebel19}.

The last group consists of pipelines learned end-to-end (e2e): Superpoint~\cite{DeTone17b}, LF-Net~\cite{Ono18}, D2-Net~\cite{Dusmanu19} (with both single- (SS) and multi-scale (MS) variants), and R2D2~\cite{Revaud19}.

Additionally, we consider Key.Net~\cite{BarrosoLaguna2019ICCVKeyNet}, a learned detector paired with HardNet \revision{and SOSNet} descriptors -- we pair it with original implementation of HardNet instead than the one provided by the authors, as it performs better\footnote{In~\cite{BarrosoLaguna2019ICCVKeyNet} the models are converted to TensorFlow -- we use the original PyTorch version.}.
\textbf{Post-IJCV update.} We have added 3 more local features to the benchmark after official publication of the paper -- DoG-AffNet-HardNet~\cite{Mishchuk17, Mishkin18}, DoG-TFeat~\cite{Balntas16b} and DoG-MKD-Concat~\cite{mukundan2018understanding}. We take them from the kornia~\cite{eriba2019kornia} library. Results of this features are included into the Tables and Figures, but not discussed in the text. 
\subsection{Feature matching}
We break this step into four stages.
Given images $\bI^i$ and $\bI^j$, $i \neq j$, we create an initial set of matches by nearest neighbor (NN) matching from $\bI^i$ to $\bI^j$, obtaining a set of matches $\mathbf{m}_{i \shortto j}$.
We optionally do the same in the opposite direction, $\mathbf{m}_{j \shortto i}$.
Lowe's ratio test \cite{Lowe04} is applied to each list to filter out non-discriminative matches, with a threshold $\ratiotest \in [0,1]$, creating ``curated'' lists $\tilde{\mathbf{m}}_{i \shortto j}$ and $\tilde{\mathbf{m}}_{j \shortto i}$.
The final set of putative matches is lists intersection, $\tilde{\mathbf{m}}_{i \shortto j} \cap \tilde{\mathbf{m}}_{j \shortto i} = \tilde{\mathbf{m}}^{\cap}_{i \leftrightarrow j}$ (known in the literature as one-to-one, mutual NN, bipartite, or cycle-consistent), or union $\tilde{\mathbf{m}}_{i \shortto j} \cup \tilde{\mathbf{m}}_{j \rightarrow i} = \tilde{\mathbf{m}}^{\cup}_{i \leftrightarrow j}$  (symmetric).
We refer to them as ``both'' and ``either'', respectively.
We also implement a simple unidirectional matching, \ie, $\tilde{\mathbf{m}}_{i \shortto j}$. 
Finally, the distance filter is optionally applied, removing matches whose distance is above a threshold.

The ``both'' strategy is similar to the ``symmetrical nearest neighbor ratio'' (sNNR)~\cite{bellavia2020NewAboutSIFT}, proposed \edit{concurrently} --
SNNR combines the nearest neighbor ratio 
\edit{in both directions into a single number by taking the harmonic mean, while our test takes the maximum of the two values.}

\subsection{Outlier pre-filtering}
Context Networks \cite{Yi18a}, or CNe for short, proposed a method to find sparse correspondences with a permutation-equivariant deep network based on PointNet \cite{Qi17a}, sparking a number of follow-up works \cite{Ranftl18,Dang18a,Zhao19,Zhang19,Sun19}.
We embed CNe into our framework.
It often works best when paired with RANSAC \cite{Yi18a,Sun19}, so we consider it as an {\em optional} pre-filtering step before RANSAC -- and apply it to both stereo and multiview.
As the published model was trained on one of our validation scenes, we re-train it on ``Notre Dame Front Facade'' and ``Buckingham Palace'', following CNe training protocol, \ie, with 2000 SIFT features, unidirectional matching, and no ratio test.
We evaluated the new model on the test set and observed that its performance is better than the one that was released by the authors.
It could be further improved by using different matching schemes, such as bidirectional matching, but we have not explored this in this paper and leave as future work.

We perform one additional, but necessary, change: CNe (like most of its successors) was originally trained to estimate the Essential matrix instead of the Fundamental matrix~\cite{Yi18a}, \ie, it assumes known intrinsics.
In order to use it within our setup, we normalize the coordinates by the size of the image instead of using ground truth calibration matrices.
This strategy has also been used in \cite{Sun19}, and has been shown to work well in practice.

\subsection{Stereo task}
\label{sec:task_stereo}
The list of tentative matches is given to a robust estimator, which estimates $\mathbf{F}_{i,j}$, the Fundamental matrix between $\bI_i$ and $\bI_j$.
In addition to (locally-optimized) RANSAC~\cite{Fischler81,LOransac2003}, as implemented in OpenCV~\cite{OpenCV}, and sklearn~\cite{sklearn}, we consider more recent algorithms \edit{with publicly available implementations}: DEGENSAC~\cite{Degensac2005}, GC-RANSAC~\cite{gcransac2018} and MAGSAC~\cite{magsac2019}. \revision{We use the original GC-RANSAC implementation\footnote{\url{https://github.com/danini/graph-cut-ransac/tree/benchmark-version}}, and not the most up-to-date version, which incorporates MAGSAC, DEGENSAC, and later changes.}

For DEGENSAC we additionally consider disabling the degeneracy check, which theoretically should be equivalent to the OpenCV and sklearn implementations -- we call this variant ``PyRANSAC''.
Given $\mathbf{F}_{i,j}$, the known intrinsics $\bK_{\{i,j\}}$ were used to compute the Essential matrix $\bE_{i,j}$, as
$\bE_{i,j} = \bK_j^T \bF_{i,j} \bK_i$. Finally, the relative rotation and translation vectors were recovered with a cheirality check with OpenCV's \texttt{recoverPose}.

\subsection{Multiview task}
\label{sec:task_multiview}
Large-scale SfM is notoriously hard to evaluate, as it requires accurate ground truth.
Since our goal is to benchmark {\em local features} and {\em matching methods}, and not SfM \edit{algorithms}, we opt for a different strategy.
We reconstruct a scene from small image subsets, which we call ``bags''.
We consider bags of 5, 10, and 25 images, which are randomly sampled from the original set of 100 images per scene -- with a co-visibility check.
We create 100 bags for bag sizes 5, 50 for bag size 10, and 25 for bag size 25 -- \ie, 175 SfM runs in total.

We use COLMAP~\cite{Schoenberger16a}, feeding it the matches computed by the previous module -- note that this comes before the robust estimation step, as COLMAP implements its own RANSAC.
If multiple reconstructions are obtained, we consider the largest one.
We also collect and report statistics such as the number of landmarks or the average track length. Both statistics and error metrics are averaged over the three bag sizes, each of which is in turn averaged over its individual bags.

\subsection{Error metrics}
\edit{Since the stereo problem is defined up to a scale factor~\cite{Hartley00}, our main error metric is based on {\em angular errors}.} 
We compute the difference, in degrees, between the {\em estimated} and {\em ground-truth} translation and rotation {\em vectors} between two cameras. 
We then threshold it over a given value for all possible \edit{-- \ie, co-visible --} pairs of images.
\edit{Doing so over different angular thresholds renders a curve.
We compute the mean Average Accuracy (mAA) by integrating this curve up to a maximum threshold, which we set to 10$^\circ$ -- this is necessary because large errors always indicate a bad pose: 30$^\circ$ is not necessarily better than 180$^\circ$, both estimates are wrong.}
\edit{Note that by computing the area under the curve we are giving more weight to methods which are more accurate at lower error thresholds, \edit{compared to using a single value at a certain designated threshold}}.

\edit{This metric was originally introduced in \cite{Yi18a}, where it was called mean Average Precision (mAP). 
We argue that ``accuracy'' \edit{is the correct terminology}, since we are \edit{simply evaluating how many of the predicted poses are ``correct'', as determined by thresholding over a given value -- \ie, our problem does not have
``false positives''.}}

\edit{The same metric is used for multiview. 
Because we do not know the scale of the scene {\em a priori}, it is not possible to measure translation error in metric terms. 
While we intend to explore this in the future, such a metric, while more interpretable, is not without problems -- for instance, the range of the distance between the camera and the scene can vary drastically from scene to scene and make it difficult to compare their results.}
To compute the mAA in pose estimation for the multiview task, we
\ky{take the mean of the average accuracy for every pair of cameras -- setting the pose error to $\infty$ for pairs containing unregistered views.}
\edit{If COLMAP returns multiple models which cannot be co-registered (which is rare) we consider only the largest of them for simplicity.}

For the stereo task, we can report this value for different co-visibility thresholds: we use $\covisibility = 0.1$ by default, which preserves most of the ``hard'' pairs. 
\edit{Note that this is not applicable to the multiview task, as all images are registered at once via bundle adjustment in SfM.}

\edit{Finally, we consider repeatability and matching score. 
Since many end-to-end methods do not report \edit{and often do not} have a clear 
measure of scale -- or support region -- we simply threshold by pixel distance, \edit{as in \cite{Rosten10}}.}
For the multiview task, we also compute the Absolute Trajectory Error (ATE)~\cite{Sturm12}, a metric widely used in SLAM. 
\edit{Since, once again,} the reconstructed model is
\edit{scale-agnostic}, we first scale the reconstructed model to that of the ground truth and then compute the ATE.
\edit{Note that ATE needs a minimum of three points to align the two models.}
\subsection{Implementation}
The benchmark code has been open-sourced\footnoteref{code}
along with every method used in the paper\footnote{{\footnotesize
\url{https://github.com/vcg-uvic/image-matching-benchmark-baselines}}}.
The implementation relies on SLURM~\cite{Yoo03} for \edit{scalable} job scheduling, \edit{which is compatible with our supercomputer clusters} -- \edit{we also provide on-the-cloud, ready-to-go images}\footnote{{\footnotesize \url{https://github.com/etrulls/slurm-gcp}}}. 
The benchmark can also run on a standard computer, sequentially.
It is computationally expensive, as it requires matching about 45k image pairs.
The most costly step -- leaving aside feature extraction, which is very method-dependent -- is typically feature matching: 2--6 seconds per image pair\footnote{Time measured on ‘n1-standard-2’ VMs on Google Cloud Compute: 2 vCPUs with 7.5 GB of RAM and no GPU.}, depending on descriptor size.
Outlier pre-filtering takes about 0.5--0.8 seconds per pair,
\edit{excluding some overhead to reformat the data into its expected format.}
RANSAC methods vary between 0.5--1 second -- as explained in \secref{ablation} we limit their number of iterations based on a compute budget, but the actual cost depends on the number of matches.
Note that these values are computed on the validation set -- for the test set experiments we increase the RANSAC budget, in order to remain compatible with the rules of \revision{the Image Matching Challenge\footnoteref{website}}.
We find COLMAP to vary drastically between set-ups.
New methods will be continuously added, and
\edit{we welcome contributions to the code base.}

\section{Details are Important}
\label{sec:ablation}

Our experiments indicate that each method needs to be carefully tuned.
In this section we outline the methodology we used to find the right hyperparameters on the validation set, and demonstrate why it is crucial to do so.

\begin{figure}
\centering
\renewcommand{\hspacer}{0.3mm}
\renewcommand{\vspacer}{0.1mm}
\renewcommand{\imscl}{0.01}
\includegraphics[scale=\imscl,width=\linewidth]{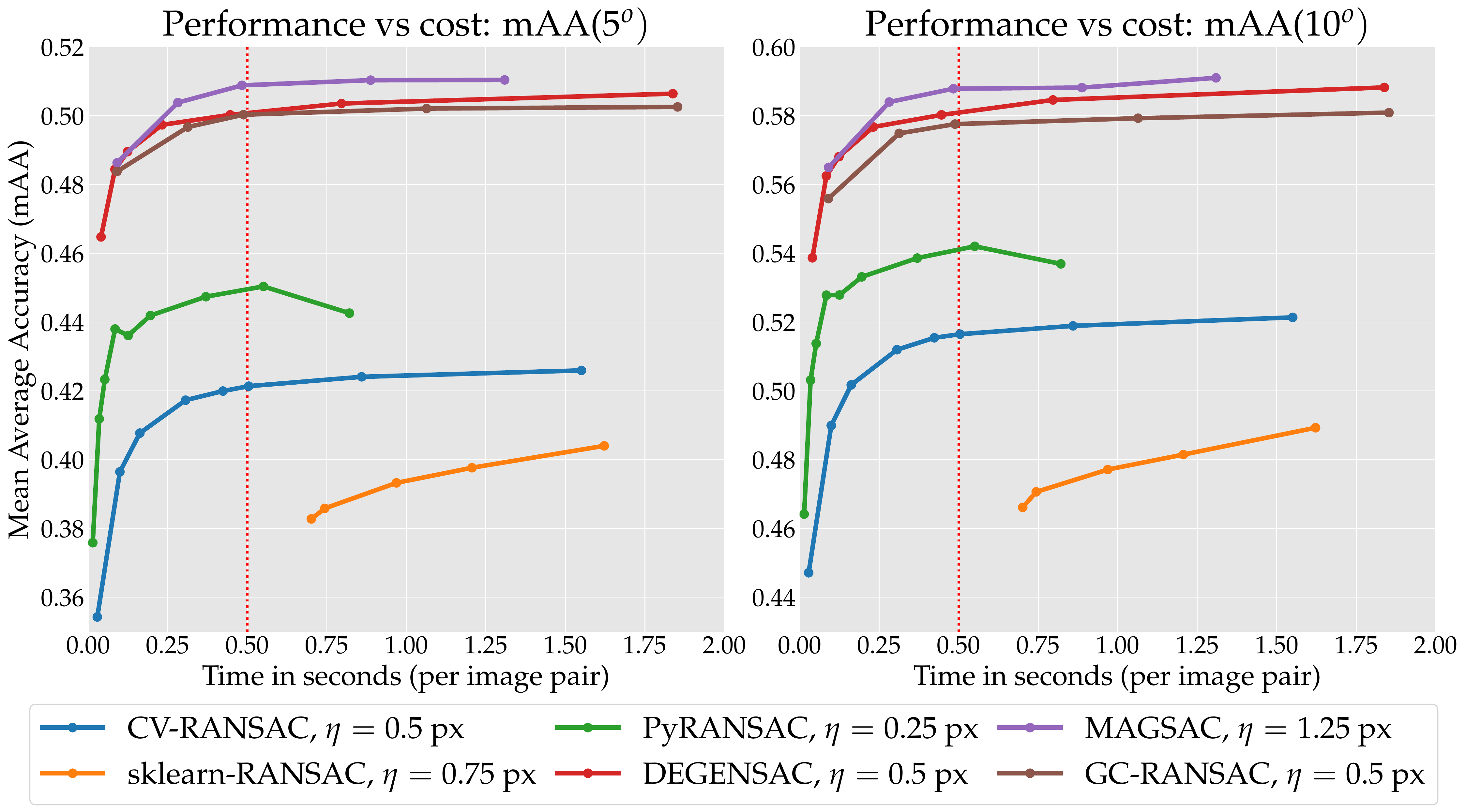}
\caption{{\bf Validation -- Performance vs. cost for RANSAC.}
\edit{We evaluate six RANSAC variants, using 8k SIFT features with ``both'' matching and a ratio test threshold of $\ratiotest{=}0.8$.
The inlier threshold $\ransacReprojTh$ and iterations limit $\ransacMaxIters$ are variables -- we plot only the best $\ransacReprojTh$ for each method, for clarity, and
set a budget of 0.5 seconds per image pair (dotted red line).
For each RANSAC variant, we pick the largest $\ransacMaxIters$ under this time ``limit'' and use it for all validation experiments.}
Computed on `n1-standard-2' VMs on Google Compute (2 vCPUs, 7.5 GB).
}
\label{fig:plots-ransac-cost-vs-map}
\end{figure}

\subsection{RANSAC: Leveling the field}
\label{sec:tuning_ransac}

Robust estimators are, \edit{in our experience,} the most sensitive part of the stereo pipeline, \edit{and thus the one we first turn to.}
All methods \edit{considered in this paper} have three parameters in common: the confidence level in their estimates, $\ransacConf$; the outlier (epipolar) threshold, $\ransacReprojTh$; and the maximum number of iterations, $\ransacMaxIters$.
\edit{We find the confidence value to be} the least sensitive, so we set it to $\ransacConf=0.999999$.

We evaluate each method with different values for $\ransacMaxIters$ and $\ransacReprojTh$, \edit{using reasonable defaults: 8k SIFT features with bidirectional matching with the ``both'' strategy and a ratio test threshold of 0.8.}
\edit{We plot the results in \fig{plots-ransac-cost-vs-map}, against their computational cost -- for the sake of clarity we only show the curve corresponding to the best reprojection threshold $\ransacReprojTh$ for each method}.

\edit{Our aim with this experiment} is to place all methods on an ``even ground'' by setting a common budget, \edit{as we need to find a way to compare them}.
We pick 0.5 seconds, where all methods have mostly converged. 
\edit{Note that these are different implementations and are obviously not directly comparable to each other, but this is a simple and reasonable approach}. 
\edit{We set this budget} by choosing
$\ransacMaxIters$ as per \fig{plots-ransac-cost-vs-map}, instead of actually {\em enforcing} a time limit, \edit{which would not be comparable across different set-ups.}
Optimal values for $\ransacMaxIters$ can vary drastically, from 10k for MAGSAC to 250k for PyRANSAC.
\edit{MAGSAC gives the best results} \edit{for this experiment, closely} followed by DEGENSAC.
We patch OpenCV
to increase the limit of iterations, which \edit{was} hardcoded to $\ransacMaxIters=1000$; this patch is now integrated into OpenCV.
This increases performance by 10-15\% relative, within our budget.
However, PyRANSAC is significantly better \edit{than OpenCV version even with this patch}, so we use it as our ``vanilla'' RANSAC instead.
The sklearn implementation is too slow for practical use.

We find that, in general, default settings can be woefully inadequate.
For example, OpenCV recommends $\ransacConf=0.99$ and $\ransacReprojTh=3$ pixels, which results in a mAA at 10$^\circ$ of \edit{0.3642} on the validation set -- a performance drop of \edit{29.3}\% relative.
\begin{figure}%
\centering
\renewcommand{\imsize}{5.2cm}
\renewcommand{\hspacer}{0.3mm}
\renewcommand{\vspacer}{0.1mm}
\renewcommand{\imscl}{0.01}
\includegraphics[scale=\imscl,width=\linewidth]{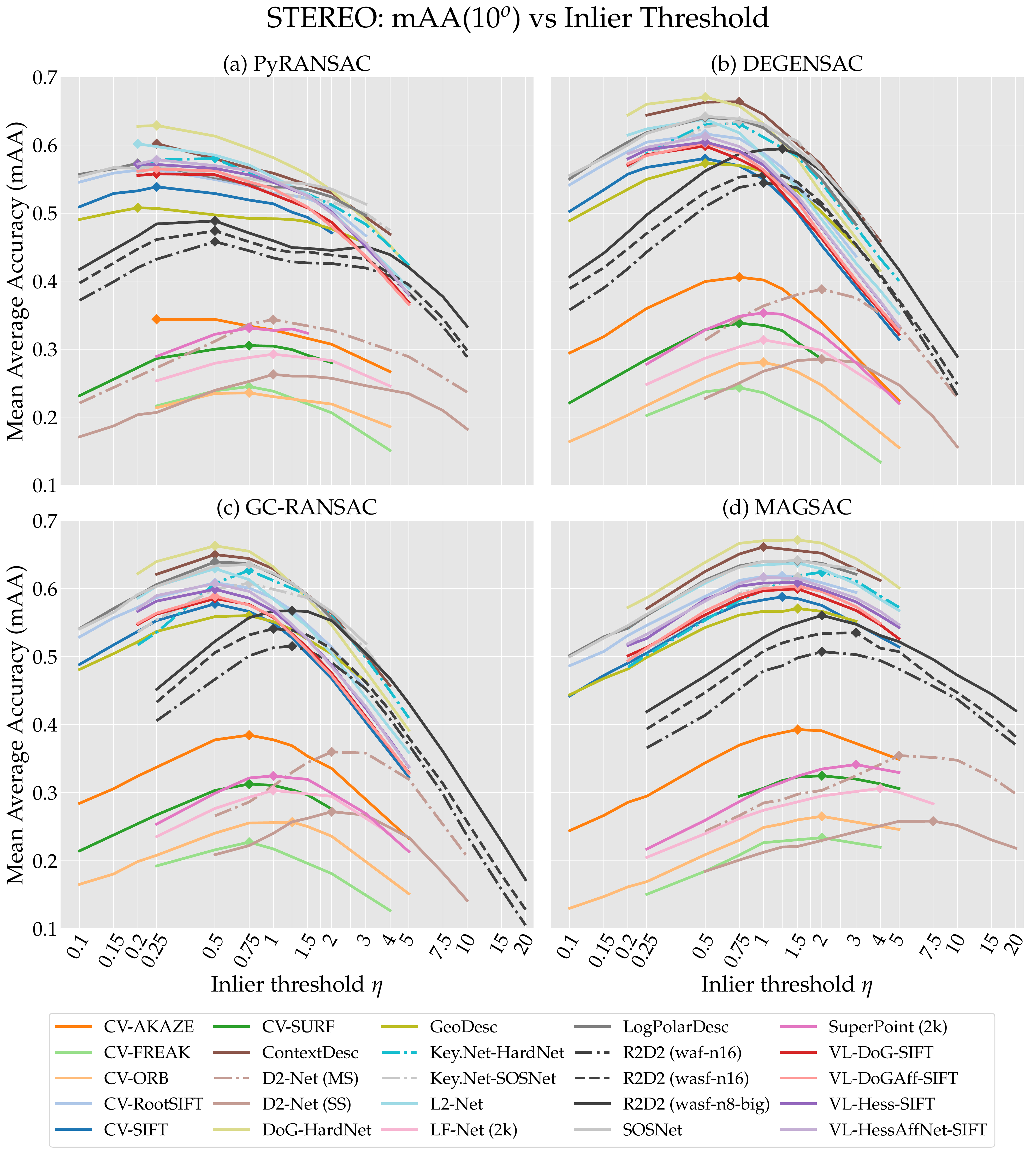}%
\caption{
{\bf Validation -- Inlier threshold for RANSAC, $\ransacReprojTh$.}
We determine $\ransacReprojTh$ for each combination, using 8k features
\edit{(2k for LF-Net and SuperPoint) 
with the ``both'' matching strategy and a reasonable value for the ratio test.}
Optimal parameters (diamonds) are listed in the \secref{appendix}.%
}
\label{fig:results_ransac_nort}
\vspace{-2mm}
\end{figure}

\subsection{RANSAC: One method at a time}
The last free parameter is the inlier threshold $\ransacReprojTh$.
We expect the optimal value for this parameter to
be different for each local feature, with looser thresholds required for methods operating on higher recall/lower precision, \edit{and end-to-end methods trained on lower resolutions}.

We report a wide array of experiments in~\fig{results_ransac_nort} that confirm our intuition: descriptors learned on DoG keypoints are clustered, while others
vary significantly.
Optimal values are also different for each RANSAC variant.
\edit{We use the ratio test with the threshold recommended by the authors of each feature, or a reasonable value if no recommendation exists, and the \edit{``both''} matching strategy -- this cuts down on the number of outliers.}

\subsection{Ratio test: One feature at a time}
\label{sec:matching}

\begin{figure}
\centering
\renewcommand{\hspacer}{0.3mm}
\renewcommand{\vspacer}{0.1mm}
\renewcommand{\imscl}{0.01}
\includegraphics[scale=\imscl,width=\linewidth]{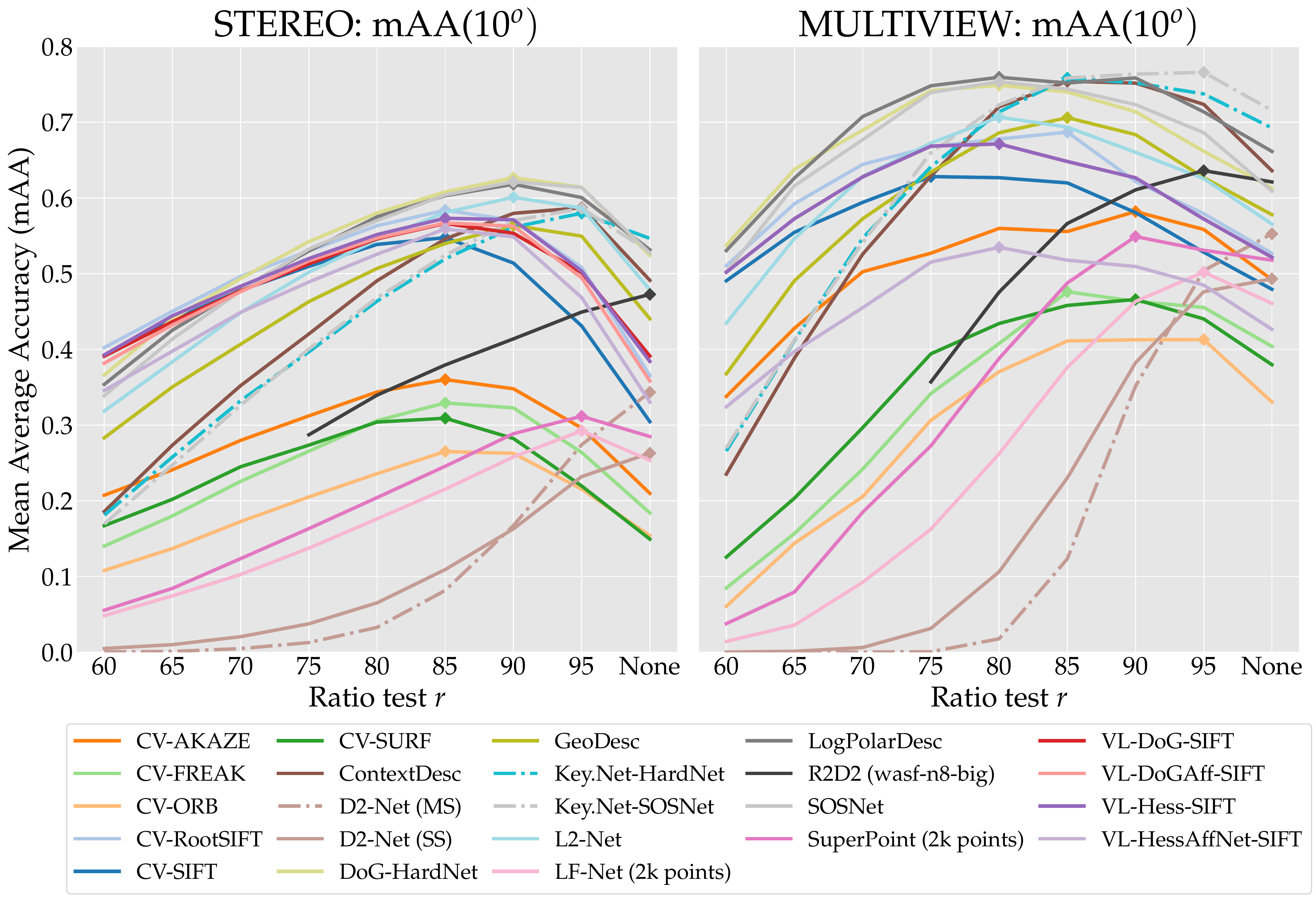} \\
\caption{
\edit{{\bf Validation -- Optimal ratio test $r$ for matching with ``both''.}
We evaluate bidirectional matching with the ``both'' strategy (the best
one), and different ratio test thresholds $r$, for each feature type.
We use 8k features (2k for SuperPoint and LF-Net).
For stereo, we use PyRANSAC.}
}
\label{fig:results_both_vs_any_and_ratio_test}
\vspace{0mm}
\end{figure}

\begin{figure}
\centering
\renewcommand{\hspacer}{0.3mm}
\renewcommand{\vspacer}{0.1mm}
\renewcommand{\imscl}{0.01}
\includegraphics[scale=\imscl,width=\linewidth]{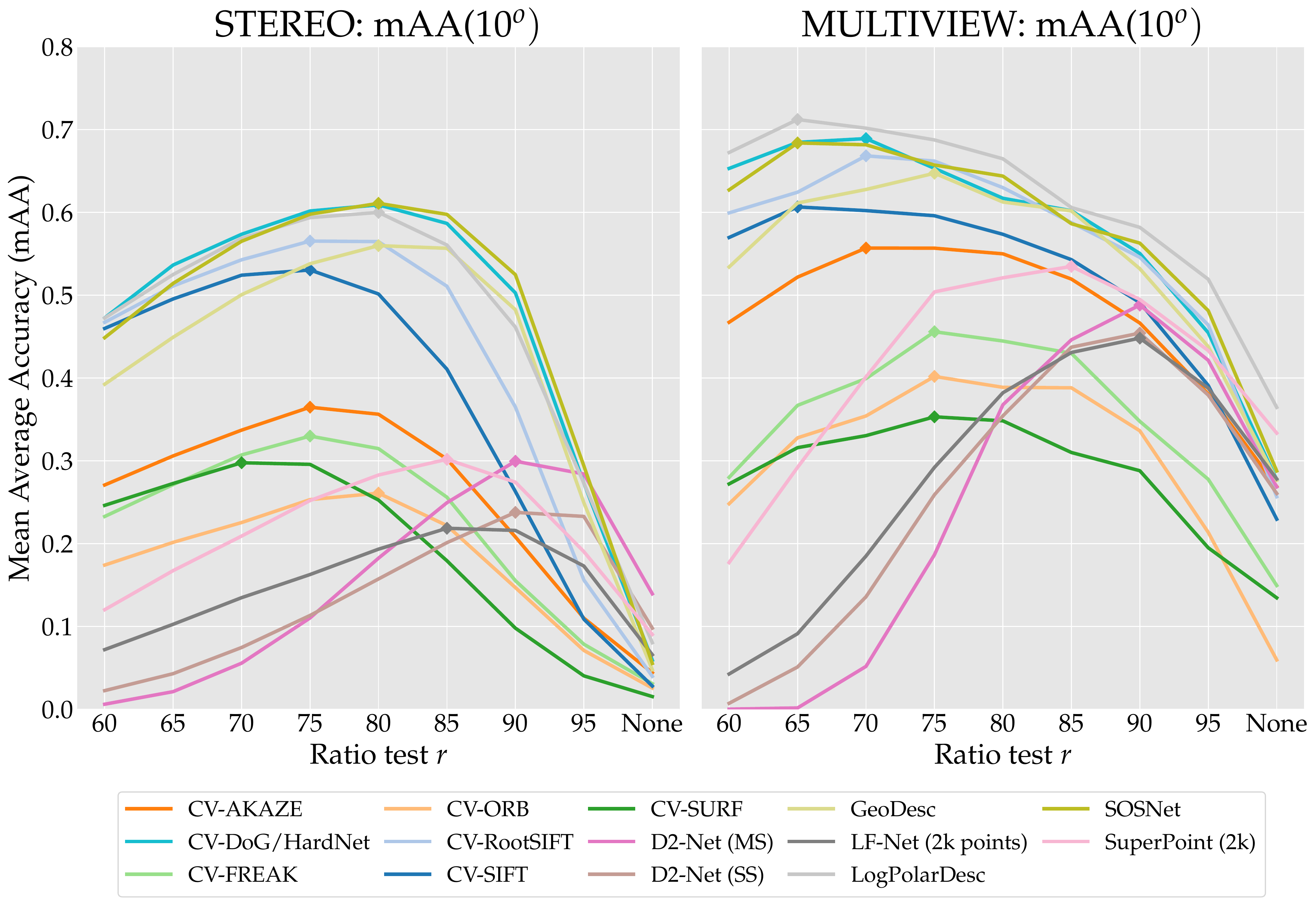} \\
\caption{
{\bf Validation -- Optimal ratio test $r$ for matching with ``either''.}
Equivalent to \fig{results_both_vs_any_and_ratio_test} but with the ``either'' matching strategy.
\edit{This strategy requires 
aggressive filtering and does not reach the performance of ``both'',
we thus explore only a subset of the methods.}
}
\label{fig:results_both_vs_any_and_ratio_test_extra}
\end{figure}

\edit{Having 
``frozen''
RANSAC, we turn to the feature matcher -- \edit{note that it comes {\em before} RANSAC, but it cannot be evaluated in isolation}.
\edit{We select PyRANSAC as a ``baseline'' RANSAC and evaluate different ratio test thresholds, separately for the stereo and multiview tasks.}
\edit{For this experiment, we use 8k features with all methods, except for those which cannot work on this regime -- SuperPoint and LF-Net. This choice will be substantiated in \secref{number-of-features}.}
We report the results for bidirectional matching with the ``both'' strategy in \fig{results_both_vs_any_and_ratio_test}, and with the ``either'' strategy in \fig{results_both_vs_any_and_ratio_test_extra}.
We find that ``both'' -- the method we have used so far -- performs best overall.
Bidirectional matching with the ``either'' strategy produces many (false) matches, increasing the computational cost in the estimator, and requires very small ratio test thresholds -- as low as $r{=}0.65$.
Our experiments with unidirectional matching indicate that is slightly worse, and it depends on the order of the images, so we did not explore it further.}

As expected, each feature requires different settings, as the distribution of their descriptors is different.
We also observe that optimal values vary significantly between stereo and multiview, even though one might expect that bundle adjustment should be able to better deal with outliers. 
We suspect \revision{that this indicates that there is room for improvement in COLMAP's implementation of RANSAC.}

Note how the ratio test is critical for performance, and one could arbitrarily select a threshold that favours one method over another, which shows the importance of proper benchmarking.
Interestingly, D2-Net is the {\em only} method that \edit{clearly} performs best without the ratio test. \edit{It also performs poorly overall in our evaluation, despite reporting state-of-the-art results in other benchmarks~\cite{Miko04b,Balntas17,Sattler12,taira2018inloc} 
-- without the ratio test, the number of tentative matches might be too high for RANSAC or COLMAP to perform well.}

Additionally, we implement the first-geometric-inconsistent ratio threshold, or FGINN \cite{Mishkin15}. 
We find that although it improves over unidirectional matching, its gains \edit{mostly} disappear against matching with ``both''. 
We report these results in \secref{fginn}.

\begin{figure}
\centering
\renewcommand{\hspacer}{0.3mm}
\renewcommand{\vspacer}{0.1mm}
\renewcommand{\imscl}{0.01}
\includegraphics[scale=\imscl,width=\linewidth]{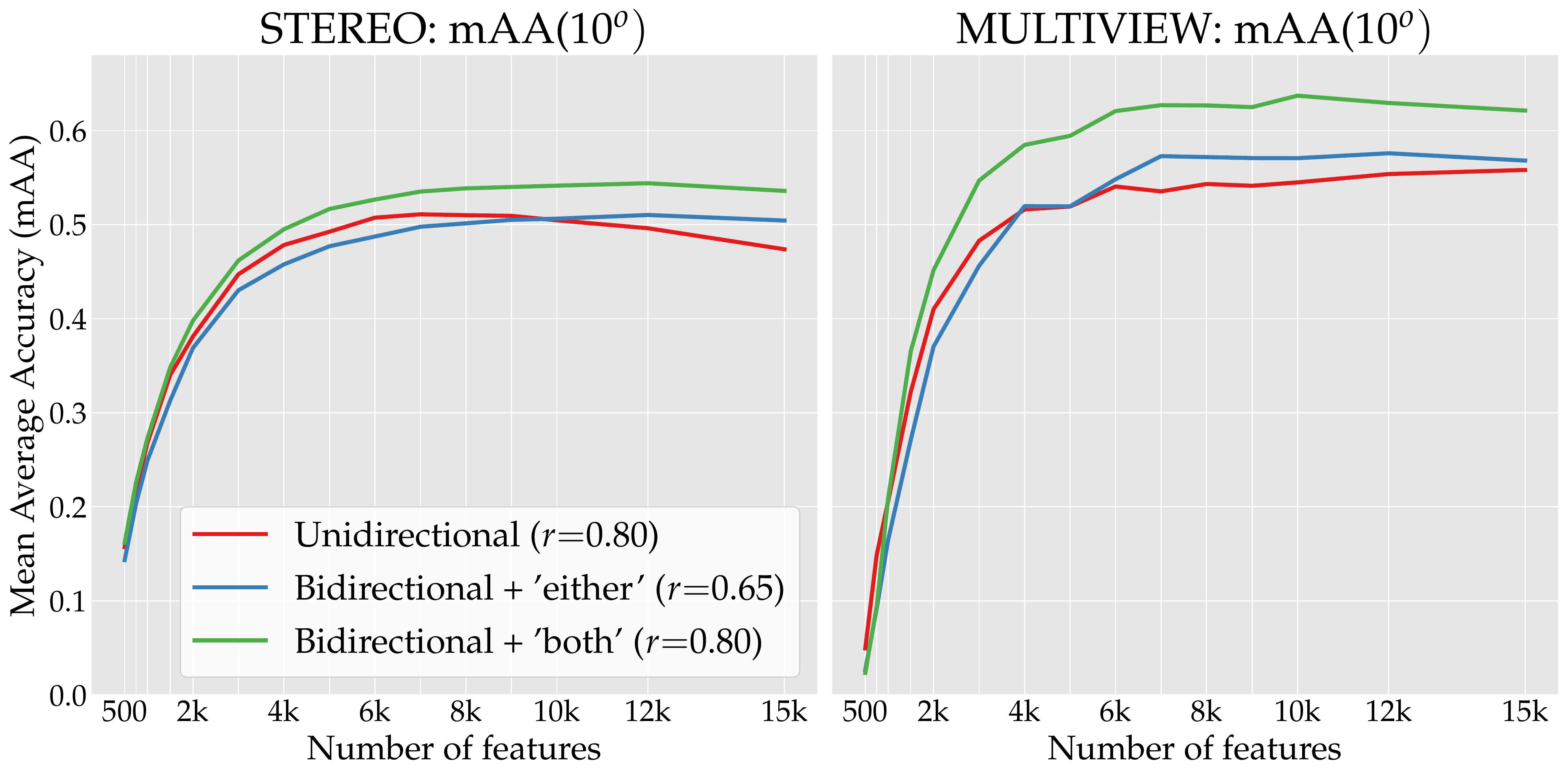} \\
\caption{
{\bf Validation -- Number of features.}
Performance on the stereo and multi-view tasks while varying the number of SIFT features, with three matching strategies, and \edit{reasonable defaults} for the ratio test $r$.
}
\label{fig:results_num_features_val}
\end{figure}

\subsection{Choosing the number of features}
\label{sec:number-of-features}

The ablation tests in this section use (up to) $K{=}8000$ feature (2k for SuperPoint and LF-Net, as they are trained to extract fewer keypoints). This number is commensurate with that used by SfM frameworks \cite{Wu13,Schoenberger16a}.
\edit{We report performance for different values of $K$ in \fig{results_num_features_val}.}
We use PyRANSAC with \edit{reasonable defaults} \edit{for all three matching strategies, with SIFT features.}

As expected, performance is strongly correlated with the number of features.
We find 8k to be a good compromise between performance and cost, and also consider 2k (actually 2048) as a `cheaper' alternative -- this also provides a fair comparison with some learned methods which only operate on that regime.
\edit{We choose these two values as valid categories for the open challenge\footnoteref{website} linked to the benchmark, and do the same on this paper for consistency.}

\subsection{Additional experiments}
\label{sec:additional-validation-experiments}

\edit{Some methods require additional considerations before evaluating them on the test set.
We briefly discuss them in this section.
Further experiments are available in \secref{appendix}.
}

\begin{figure}
\centering
\renewcommand{\imsize}{4.4cm}
\renewcommand{\hspacer}{0.1mm}
\renewcommand{\vspacer}{0.1mm}
\renewcommand{\imscl}{0.01}
\includegraphics[scale=\imscl,height=\imsize]{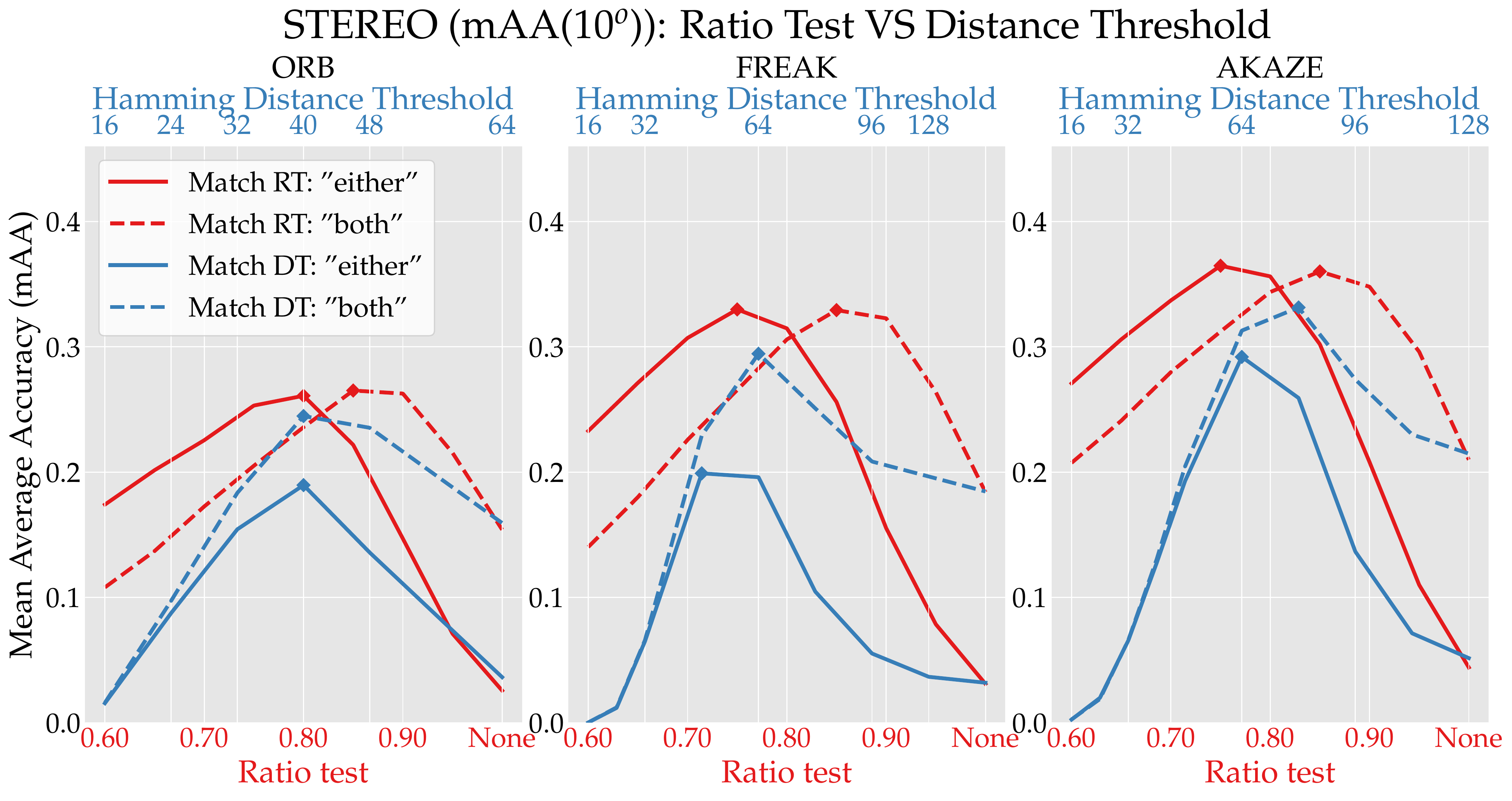}\hspace{\hspacer}%
\caption{
{\bf Validation -- Matching binary descriptors.}
We filter out non-discriminative matches with the ratio test or a distance threshold. The latter (the standard) performs worse in our experiments.%
}
\label{fig:matching_binary}
\end{figure}

\paragraph{Binary features (\fig{matching_binary}).}
\edit{We consider three binary descriptors: ORB \cite{Rublee11}, AKAZE \cite{Alcantarilla13}, and FREAK \cite{Alahi12}
}
Binary descriptor papers \edit{historically} favour a distance threshold in place of the ratio test to reject non-discriminative matches \cite{Rublee11}, although some papers have used the ratio test for ORB descriptors~\cite{saddle2019}.
We evaluate both in~\fig{matching_binary} -- \edit{as before, we use up to 8k features and matching with the ``both'' strategy}.
\edit{The ratio test works better for all three methods -- we use it instead of a distance threshold for all experiments in the paper, including those in the previous sections.}

\begin{figure}
\centering
\renewcommand{\imsize}{5.2cm}
\renewcommand{\hspacer}{0.3mm}
\renewcommand{\vspacer}{0.1mm}
\renewcommand{\imscl}{0.01}
\includegraphics[scale=\imscl,width=0.49\linewidth]{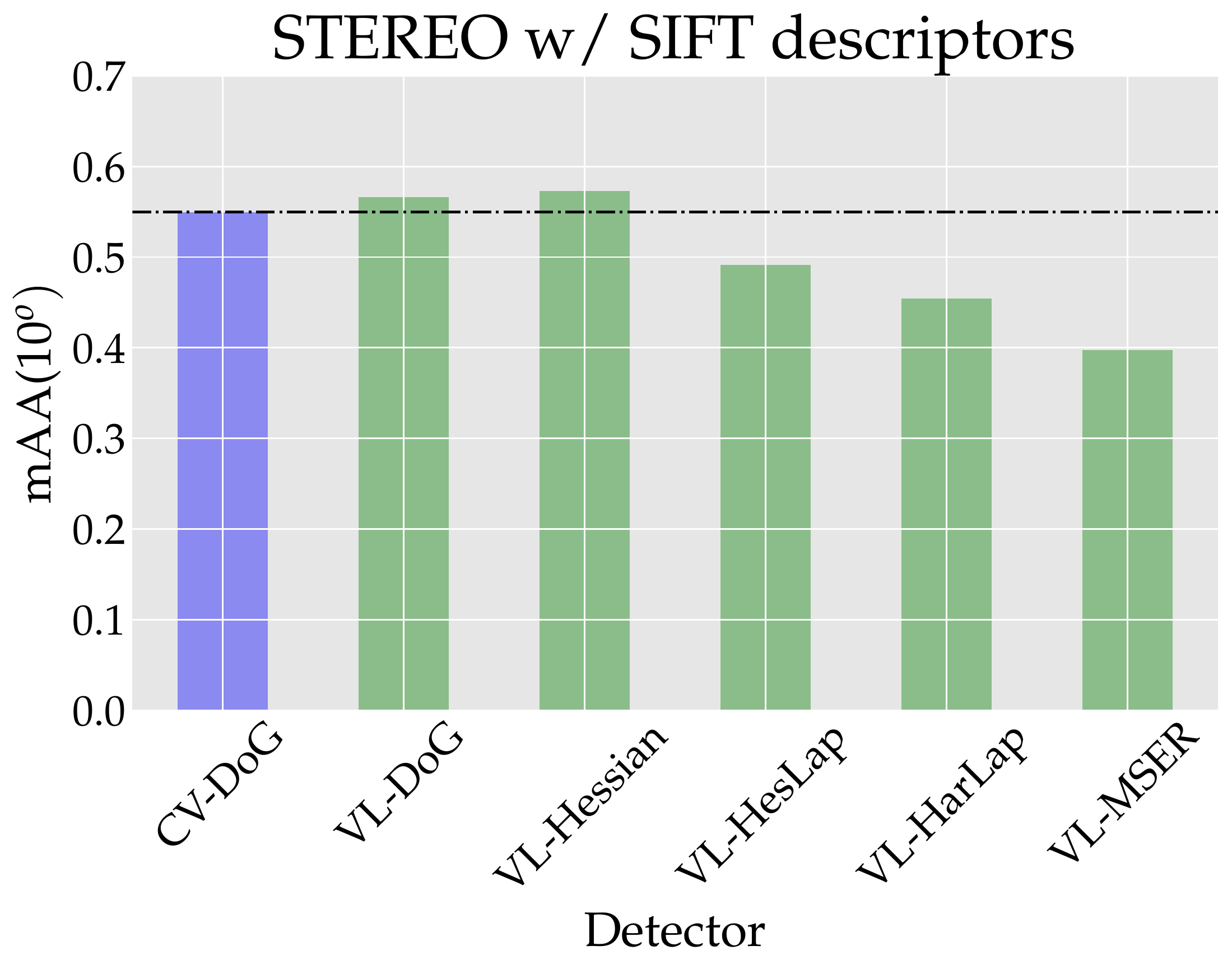}%
\hspace{\hspacer}%
\includegraphics[scale=\imscl,width=0.49\linewidth]{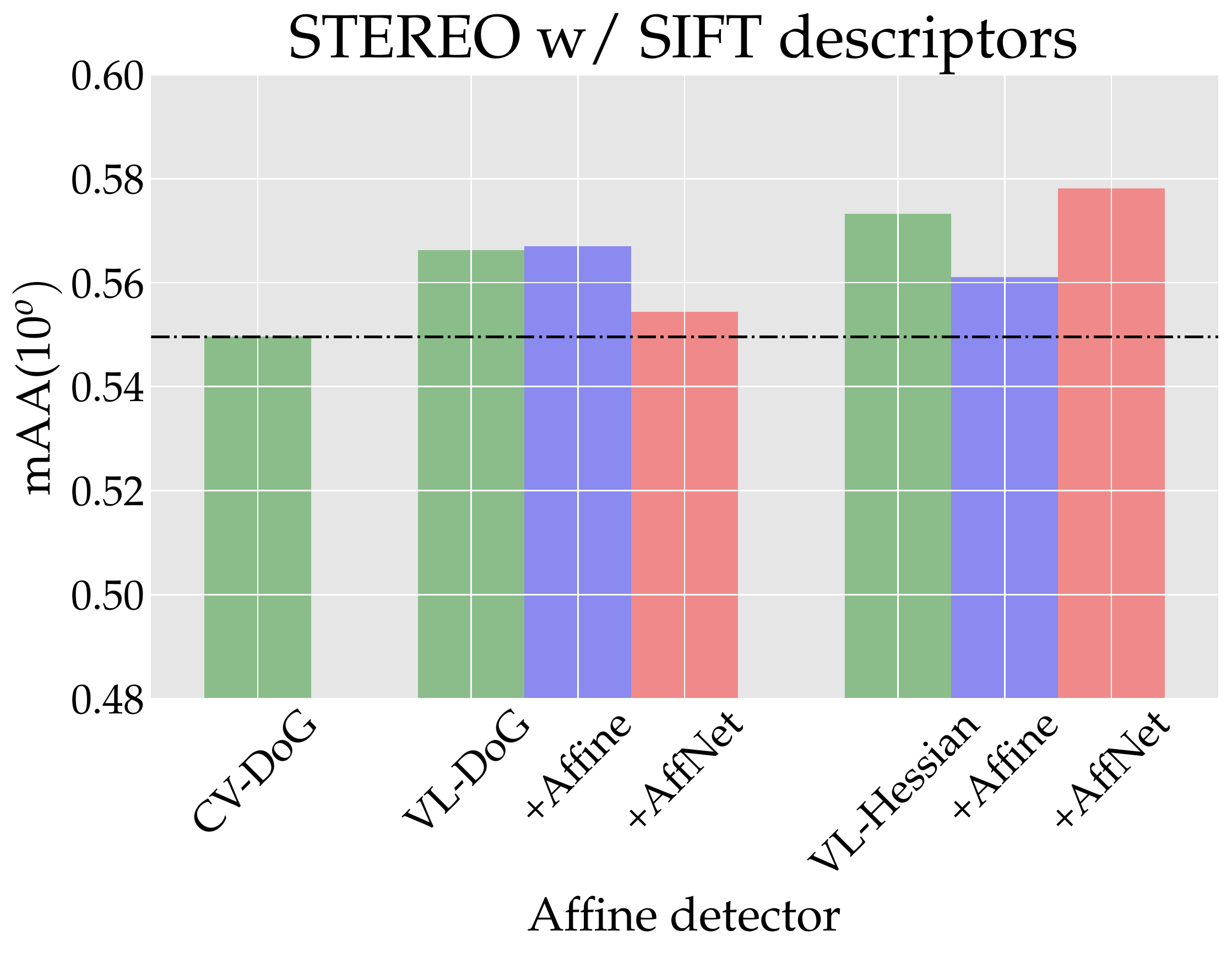}%
\caption{
{\bf Validation -- Benchmarking detectors.}
We evaluate the performance on the stereo task while pairing different detectors with SIFT descriptors.
The dashed, black line indicates OpenCV SIFT -- the baseline.
{\bf Left:} OpenCV DoG vs. VLFeat implementations of blob detectors (DoG, Hessian, HesLap) and corner detectors (Harris, HarLap), and MSER.
{\bf Right:} Affine shape estimation for DoG and Hessian keypoints, against the plain version. We consider a classical approach, Baumberg (Affine)~\cite{Baumberg00}, and the recent, learned AffNet~\cite{Mishkin18}
-- they provide a small but inconsistent boost.
}
\label{fig:results_vlfeat}
\end{figure}
\paragraph{On the influence of the detector (\fig{results_vlfeat}).}
We %
embed several popular blob and corner detectors into our pipeline, with OpenCV's DoG \cite{Lowe04} as a baseline.
We combine multiple methods, taking advantage of the VLFeat library:
\edit{Difference of Gaussians} (DoG), Hessian \cite{Hessian78}, HessianLaplace \cite{Mikolajczyk04}, HarrisLaplace \cite{Mikolajczyk04}, MSER \cite{Matas04}, DoGAffine, Hessian-Affine \cite{Mikolajczyk04,Baumberg00}, DoG-AffNet \cite{Mishkin18}, and Hessian-AffNet \cite{Mishkin18}.
We pair them with SIFT descriptors, also computed with VLFeat, as OpenCV cannot process affine keypoints, and report the results in \fig{results_vlfeat}.
VLFeat's DoG performs marginally better than OpenCV's.
Its affine version gives a small boost.
Given the small gain and the infrastructure burden of interacting with a Matlab/C library, we use OpenCV's DoG implementation for most of this paper.

\begin{figure}
\centering
\renewcommand{\imsize}{4.3cm}
\renewcommand{\hspacer}{0.3mm}
\renewcommand{\vspacer}{0.1mm}
\renewcommand{\imscl}{0.01}
\includegraphics[scale=\imscl,width=0.8\linewidth]{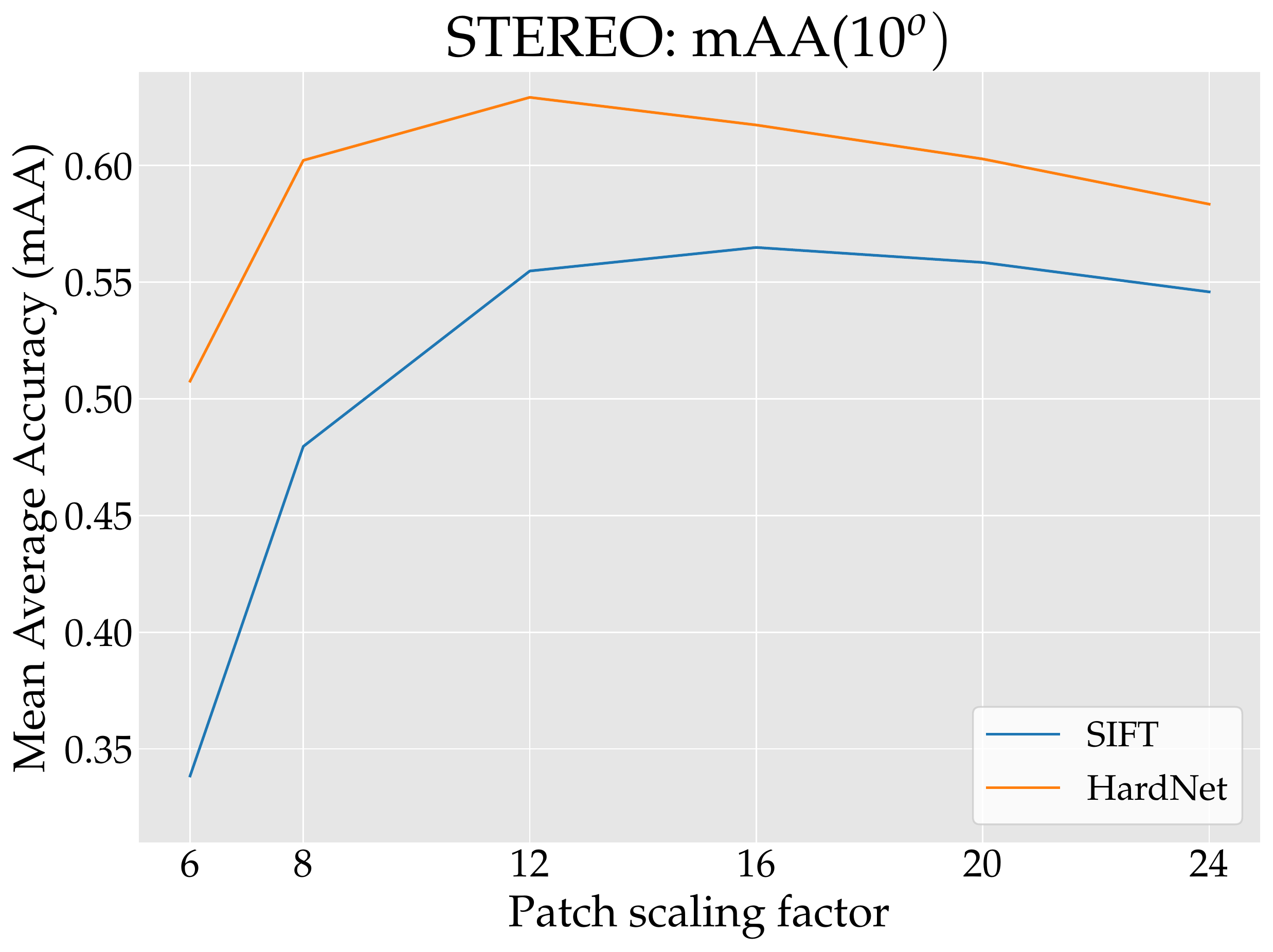}
\caption{
{\bf Validation -- Scaling the descriptor support region.}
Performance with SIFT and HardNet descriptors while applying a scaling factor $\lambda$ to the keypoint scale (note that OpenCV's default value is $\lambda{=}12$). 
\edit{We consider SIFT and HardNet. 
Default values are optimal or near-optimal.}%
}
\label{fig:results_mrsize}
\end{figure}
\paragraph{On increasing the support region (\fig{results_mrsize}).}
The size (``scale'') of the support region used to compute a descriptor can significantly affect its performance \cite{Dong15,Zagoruyko15,Aanaes02}.
We experiment with different scaling factors, using DoG with SIFT and HardNet \cite{Mishchuk17}, and find that 12$\times$ the OpenCV scale \ky{(the default value)} is already nearly optimal, confirming the findings \edit{reported} in \cite{Ebel19}.
We show these results in \fig{results_mrsize}.
Interestingly, SIFT descriptors do benefit from increasing the scaling factor from 12 to 16, \edit{but the difference is very small -- we thus use the recommended value of 12 for the rest of the paper}.
This, however, suggests that deep descriptors such as HardNet might be able to increase performance slightly by training on larger patches.

\section{Establishing the State of the Art}
\label{sec:results}

With the findings \edit{and the optimal parameters} found in \secref{ablation}, we move on to the test set, evaluating many methods with their optimal settings.
All experiments in this section use bidirectional matching with the `'both'' strategy.
We consider a large feature budget (up to 8k features) and a smaller one (up to 2k), and evaluate many detector/descriptor combinations.

We make three changes with respect to the validation experiments of the previous section.
(1) We double the RANSAC budget from 0.5 seconds (used for validation) to 1 second per image pair, and adjust the maximum number of iterations $\ransacMaxIters$ accordingly -- we made this decision to encourage participants to the challenge based on this benchmark to use built-in methods rather than run RANSAC themselves to squeeze out a little extra performance, and use the same values in the paper for consistency.
(2) We run each stereo and multiview evaluation three times and average the results, in order to decrease the potential randomness in the results -- \edit{in general, we found the variations within these three runs to be negligible}.
(3) We use brute-force to match descriptors instead of FLANN, as we observed a drop in performance. \edit{For more details, see Section~\ref{sec:flann} and Table~\ref{tbl:test_flann}}.

For stereo, we consider DEGENSAC and MAGSAC, which perform the best in the validation set, and PyRANSAC as a `baseline' RANSAC.
We report the results with 
\edit{both 8k features and 2k features in the following subsections.}
All observations are in terms of mAA, our primary metric, unless stated otherwise.

\definecolor{light-gray}{gray}{0.9}
\renewcommand{\arraystretch}{0.95}
\renewcommand{\tabcolsep}{0.8mm}
\begin{table}
\begin{center}
\begin{adjustbox}{width=\linewidth,center}
\footnotesize
\begin{tabular}{@{}lcccccccc@{}}
\toprule
\multicolumn{2}{c}{} & \multicolumn{2}{c}{PyRANSAC} & \multicolumn{2}{c}{DEGENSAC} & \multicolumn{2}{c}{MAGSAC} & \\
Method & NF & NI$^\uparrow$ & mAA(10$^o$)$^\uparrow$ & NI$^\uparrow$ & mAA(10$^o$)$^\uparrow$ & NI$^\uparrow$ & mAA(10$^o$)$^\uparrow$ & Rank \\
\midrule
CV-SIFT & 7861.1 & 167.6 & .3996 & 243.6 & .4584 & 297.4 & .4583 & 14 \\
VL-SIFT & 7880.6 & 179.7 & .3999 & 261.6 & .4655 & 326.2 & .4633 & 13 \\
VL-Hessian-SIFT & 8000.0 & 204.4 & .3695 & 290.2 & .4450 & 348.9 & .4335 & 15 \\
VL-DoGAff-SIFT & 7892.1 & 171.6 & .3984 & 250.1 & .4680 & 317.1 & .4666 & 11 \\
VL-HesAffNet-SIFT & 8000.0 & 209.3 & .3933 & 299.0 & .4679 & 350.0 & .4626 & 12 \\
\midrule
CV-$\sqrt{\text{SIFT}}$ & 7860.8 & 192.3 & .4228 & 281.7 & .4930 & 347.5 & .4941 & 10 \\
CV-SURF & 7730.0 & 107.9 & .2280 & 113.6 & .2593 & 145.3 & .2552 & 19 \\
CV-AKAZE & 7857.1 & 131.4 & .2570 & 246.8 & .3074 & 301.8 & .3036 & 17 \\
CV-ORB & 7150.2 & 123.7 & .1220 & 150.0 & .1674 & 178.9 & .1570 & 22 \\
CV-FREAK & 8000.0 & 123.3 & .2273 & 131.0 & .2711 & 196.7 & .2656 & 18 \\
\midrule
L2-Net & 7861.1 & 213.8 & .4621 & 366.0 & .5295 & 481.0 & .5252 & 5 \\
DoG-HardNet & 7861.1 & 286.5 & .4801 & 432.3 & {\bf \textcolor{Green}{.5543}} & 575.1 & .5502 & 2 \\
DoG-HardNetAmos+ & 7861.0 & 265.7 & .4607 & 398.6 & .5385 & 528.7 & .5329 & 3 \\
Key.Net-HardNet & 7997.6 & 448.1 & .3997 & 598.3 & .4986 & 815.4 & .4739 & 9 \\
Key.Net-SOSNet & 7997.6 & 275.5 & .4236 & 587.4 & .5019 & 766.4 & .4780 & 8 \\
GeoDesc & 7861.1 & 205.4 & .4328 & 348.5 & .5111 & 453.4 & .5056 & 7 \\
ContextDesc & 7859.0 & 278.2 & .4684 & 493.6 & .5098 & 544.1 & .5143 & 6 \\
DoG-SOSNet & 7861.1 & 281.6 & .4784 & 424.6 & {\bf \textcolor{red}{.5587}} & 563.3 & {\bf \textcolor{NavyBlue}{.5517}} & 1 \\
LogPolarDesc & 7861.1 & 254.4 & .4574 & 441.8 & .5340 & 591.2 & .5238 & 4 \\
\midrule
D2-Net (SS) & 5665.3 & 280.8 & .1933 & 482.3 & .2228 & 781.3 & .2032 & 21 \\
D2-Net (MS) & 6924.1 & 278.2 & .2160 & 470.6 & .2506 & 741.2 & .2321 & 20 \\
R2D2 (wasf-n8-big) & 7940.5 & 457.6 & .3683 & 842.2 & .4437 & 998.9 & .4236 & 16 \\
\midrule
DoG-AffNet-HardNet & 7834.0 & 267.9 & .4505 & 403.4 & .5447 & 516.8 & .5412 & $3^{*}$ \\
DoG-MKD-Concat & 7860.8 & 208.0 & .4061 & 305.8 & .4846 & 381.4 & .4810 & $11^{*}$  \\
DoG-TFeat & 7860.8 & 160.8 & .4008 & 234.8 & .4649 & 292.6 & .4668 & $13^{*}$ \\
\bottomrule
\end{tabular}
\end{adjustbox}
\end{center}
\caption{
{\bf Test -- Stereo results with 8k features.}
We report: {\bf (NF)} Number of Features; {\bf (NI)} Number of Inliers produced by RANSAC; and {\bf mAA(10$^o$)}.
Top three methods by mAA marked in
\textcolor{red}{red}, \textcolor{Green}{green} and \textcolor{NavyBlue}{blue}.
 $^{*}$ {\bf The last group of results is obtained after the paper publication and not described in the text of the paper. Their rank does not influence other entries ranks.}
}
\label{tbl:test_stereo}
\end{table}

\renewcommand{\arraystretch}{0.6}
\renewcommand{\tabcolsep}{1.2mm}
\begin{table}
\begin{center}
\begin{adjustbox}{width=\linewidth,center}
\footnotesize
\begin{tabular}{@{}lcccccccc@{}}
\toprule
Method & NL$^\uparrow$ & SR$^\uparrow$ & RC$^\uparrow$ & TL$^\uparrow$ & mAA(5$^o$)$^\uparrow$ & mAA(10$^o$)$^\uparrow$ & ATE$^\downarrow$ & Rank \\
\midrule
CV-SIFT & 2577.6 & 96.7 & 94.1 & 3.95 & .5309 & .6261 & .4721 & 14 \\
VL-SIFT & 3030.7 & 97.9 & 95.4 & 4.17 & .5273 & .6283 & .4669 & 13 \\
VL-Hessian-SIFT & 3209.1 & 97.4 & 94.1 & 4.13 & .4857 & .5866 & .5175 & 16 \\
VL-DoGAff-SIFT & 3061.5 & 98.0 & 96.2 & 4.11 & .5263 & .6296 & .4751 & 12 \\
VL-HesAffNet-SIFT & 3327.7 & 97.7 & 95.2 & 4.08 & .5049 & .6069 & .4897 & 15 \\
\midrule
CV-$\sqrt{\text{SIFT}}$ & 3312.1 & 98.5 & 96.6 & 4.13 & .5778 & .6765 & .4485 & 9 \\
CV-SURF & 2766.2 & 94.8 & 92.6 & 3.47 & .3897 & .4846 & .6251 & 18 \\
CV-AKAZE & 4475.9 & 99.0 & 95.4 & 3.88 & .4516 & .5553 & .5715 & 17 \\
CV-ORB & 3260.3 & 97.2 & 91.1 & 3.45 & .2697 & .3509 & .7377 & 22 \\
CV-FREAK & 2859.1 & 92.9 & 91.7 & 3.53 & .3735 & .4653 & .6229 & 20 \\
\midrule
L2-Net & 3424.9 & 98.6 & 96.2 & 4.21 & .5661 & .6644 & .4482 & 10 \\
DoG-HardNet & 4001.4 & 99.5 & {\bf \textcolor{NavyBlue}{97.7}} & {\bf \textcolor{NavyBlue}{4.34}} & {\bf \textcolor{red}{.6090}} & {\bf \textcolor{red}{.7096}} & {\bf \textcolor{red}{.4187}} & 1 \\
DoG-HardNetAmos+ & 3550.6 & 98.8 & 96.9 & 4.28 & .5879 & .6888 & .4428 & 6 \\
Key.Net-HardNet & 3366.0 & 98.9 & 96.7 & 4.32 & .5391 & .6483 & .4622 & 11 \\
Key.Net-SOSNet & {\bf \textcolor{NavyBlue}{5505.5}} & 100.0 & {\bf \textcolor{red}{98.7}} & {\bf \textcolor{Green}{4.46}} & .5989 & {\bf \textcolor{Green}{.7038}} & .4286 & 2 \\
GeoDesc & 3839.0 & 99.1 & 97.2 & 4.26 & .5782 & .6803 & .4445 & 8 \\
ContextDesc & 3732.5 & 99.3 & 97.6 & 4.22 & {\bf \textcolor{Green}{.6036}} & {\bf \textcolor{NavyBlue}{.7035}} & {\bf \textcolor{NavyBlue}{.4228}} & 3 \\
DoG-SOSNet & 3796.0 & 99.3 & 97.4 & 4.32 & {\bf \textcolor{NavyBlue}{.6032}} & .7021 & {\bf \textcolor{Green}{.4226}} & 4 \\
LogPolarDesc & 4054.6 & 99.0 & 96.4 & 4.32 & .5928 & .6928 & .4340 & 5 \\
\midrule
D2-Net (SS) & {\bf \textcolor{Green}{5893.8}} & {\bf \textcolor{Green}{99.8}} & 97.5 & 3.62 & .3435 & .4598 & .6361 & 21 \\
D2-Net (MS) & {\bf \textcolor{red}{6759.3}} & {\bf \textcolor{NavyBlue}{99.7}} & {\bf \textcolor{Green}{98.2}} & 3.39 & .3524 & .4751 & .6283 & 19 \\
R2D2 (wasf-n8-big) & 4432.9 & {\bf \textcolor{NavyBlue}{99.7}} & 97.2 & {\bf \textcolor{red}{4.59}} & .5775 & .6832 & .4333 & 7 \\
\midrule
DoG-AffNet-HardNet & 4671.3 & {\bf \textcolor{Green}{99.9}} & {\bf \textcolor{NavyBlue}{98.1}} & {\bf \textcolor{Green}{4.56}} & {\bf \textcolor{red}{.6296}} & {\bf \textcolor{red}{.7267}} & {\bf \textcolor{red}{.4021}} & $1^{*}$ \\
DoG-MKD-Concat & 3507.4 & 98.5 & 96.1 & 4.17 & .5461 & .6476 & .4668 & $11^{*}$ \\
DoG-TFeat & 2905.3 & 97.1 & 94.8 & 4.04 & .5270 & .6261 & .4873 & $14^{*}$ \\
\bottomrule
\end{tabular}
\end{adjustbox}
\end{center}
\caption{
{\bf Test -- Multiview results with 8k features.}
We report:
{\bf (NL)} Number of 3D Landmarks;
{\bf (SR)} Success Rate (\%) in the 3D reconstruction across ``bags'';
{\bf (RC)} Ratio of Cameras (\%) registered in a ``bag'';
{\bf (TL)} Track Length or number of observations per landmark;
{\bf mAA} at 5 and 10$^o$;
and {\bf (ATE)} Absolute Trajectory Error.
All metrics are averaged across different ``bag'' sizes, as explained in \secref{benchmark}.
We rank them by mAA at 10$^o$ and color-code them as in \tbl{test_stereo}.
 $^{*}$ {\bf The last group of results is obtained after the paper publication and not described in the text of the paper. Their rank does not influence other entries ranks.}
}
\label{tbl:test_multiview}
\end{table}

\subsection{Results with 8k features \edit{--- Tables~\ref{tbl:test_stereo}~and~\ref{tbl:test_multiview}}}

On the stereo task, deep descriptors extracted on DoG keypoints are at the top in terms of mAA, with SOSNet being \#1, closely followed by HardNet.
Interestingly, `HardNetAmos+'~\cite{HardNetAMOS2019}, a version trained on more datasets -- Brown~\cite{Brown10}, HPatches~\cite{Balntas17}, and AMOS-patches~\cite{HardNetAMOS2019} -- performs worse than the original models, trained only on the ``Liberty'' scene from Brown's dataset.
On the multiview task, HardNet edges out ContextDesc, SOSNet and LogpolarDesc by a small margin.

We also pair HardNet \revision{and SOSNet} with Key.Net, a learned detector, which performs worse than with DoG when extracting a large number of features, \revision{with the exception of Key.Net + SOSNet on the multiview task.}

R2D2, the best performing end-to-end method, does well on multiview (\revision{\#7}), but performs worse than SIFT on stereo -- it produces a much larger number of ``inliers'' (which may be correct or incorrect) than most other methods. This suggests that, like D2-Net, its lack of compatibility with the ratio test may be a problem when paired with sample-based robust estimators, due to a lower inlier ratio.
Note that D2-net performs poorly on our benchmark, despite state-of-the-art results on others.
On the multiview task it creates many more \ky{3D} landmarks than any other method.
\edit{Both issues may be related to its poor localization (pixel) accuracy, due to operating on downsampled feature maps.}

Out of the handcrafted methods, SIFT -- RootSIFT specifically -- \edit{remains competitive}, \edit{being}
\revision{\#10} on stereo and \revision{\#9} on multiview, within 13.1\% and 4.9\% relative of the top performing method, respectively,
\edit{while previous benchmarks report differences in performance of {\em orders of magnitude.}}
Other ``classical'' features do not fare so well. 
\edit{One interesting observation is that among these, their ranking on validation and test set is not consistent -- Hessian is better on validation than DoG, but significantly worse on the test set, especially in the multiview setup.}
\revision{While this is a special case, this nonetheless demonstrates that a small-scale benchmark can be misleading, and a method needs to be tested on a variety of scenes, which is what our test set aims to provide.}

Regarding the robust estimators, DEGENSAC and MAGSAC both perform very well, with the former edging out the latter for most local feature methods.
This may be due to the nature of the scenes, which \edit{often} contain dominant planes.

\subsection{Results with 2k features \edit{--- Tables~\ref{tbl:test_stereo_2k} and~\ref{tbl:test_multiview_2k}}}

Results change slightly on the low-budget regime, \revision{where the top two spots on both tasks are occupied by Key.Net+SOSNet and Key.Net+HardNet.} It is closely followed by LogPolarDesc (\revision{\#3 on stereo and \#4 on multiview}), \edit{a method} trained on DoG keypoints -- but using a much larger support region, resampled into log-polar patches. 
R2D2 performs very well on the multiview task (\revision{\#3}), while once again falling a bit short on the stereo task \edit{(\revision{\#8, and 14.5\%} relative below the \#1 method)}, for which it \edit{retrieves} a number of inliers significantly larger than its competitors.
The rest of the end-to-end methods do not perform so well, other than SuperPoint, which obtains competitive results on the multiview task.

The difference between classical and learned methods is more pronounced than with 8k points, with RootSIFT once again at the top, but now within 31.4\% relative of the \#1 method on stereo, and \revision{26.9\%} on multiview.
\edit{This is somewhat to be expected, given that with fewer keypoints, the quality of each individual point matters more.}

\begin{figure}
\centering
\renewcommand{\hspacer}{0.3mm}
\renewcommand{\vspacer}{0.1mm}
\renewcommand{\imscl}{0.01}
\includegraphics[scale=\imscl,width=\linewidth]{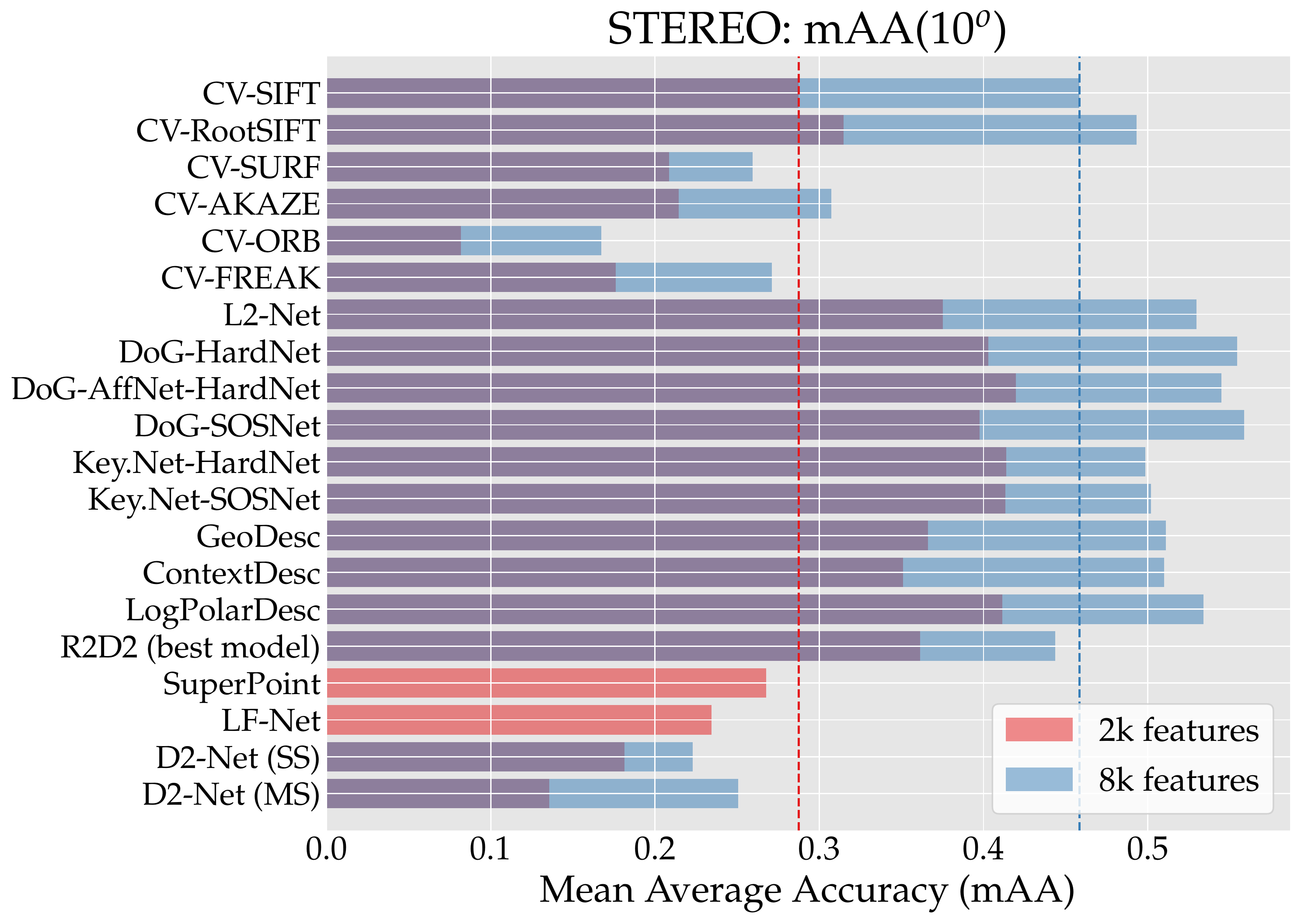}
\caption{
{\bf Test -- Stereo performance: 2k vs 8k features.}
\edit{We compare the results obtained with different methods using either 2k or 8k features -- we use DEGENSAC, which performs better than other RANSAC variants under most circumstances. 
Dashed lines indicate SIFT's performance.
For LF-Net and SuperPoint we do not include results with 8k features, as we failed to obtain meaningful results.
For R2D2, we use the best model for each setting.}
}
\label{fig:rest_2k_vs_8k}
\vspace{-3mm}
\end{figure}

\begin{figure}
\centering
\renewcommand{\hspacer}{0.3mm}
\renewcommand{\vspacer}{0.1mm}
\renewcommand{\imscl}{0.01}
\includegraphics[scale=\imscl,width=\linewidth]{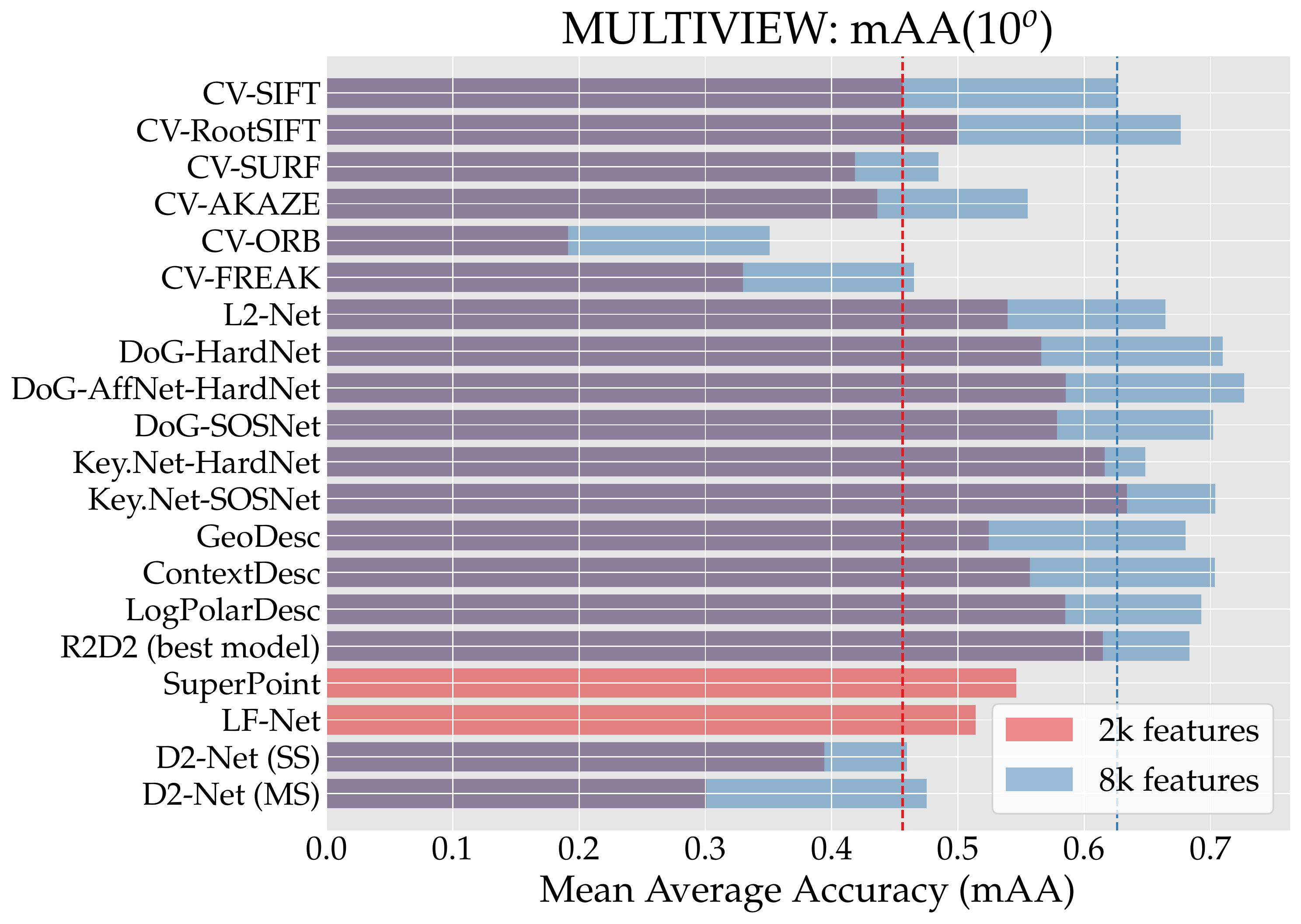}
\caption{
{\bf Test -- Multiview performance: 2k vs 8k features.}
\edit{Same as \fig{rest_2k_vs_8k}, for multiview.
}
}
\label{fig:rest_2k_vs_8k_multiview}
\vspace{-3mm}
\end{figure}

\subsection{2k features vs 8k features \edit{--- Figs.~\ref{fig:rest_2k_vs_8k}~and~\ref{fig:rest_2k_vs_8k_multiview}}}

We compare the results between the low- and high-budget regimes in \fig{rest_2k_vs_8k}, for stereo (with DEGENSAC), and \fig{rest_2k_vs_8k_multiview}, for multiview.
Note how methods can behave quite differently. 
\edit{Those} based on DoG \edit{significantly benefit from an increased feature budget}, whereas those learned end-to-end \edit{may} require re-training -- this is exemplified by the difference in performance between 2k and 8k for Key.Net+Hardnet, specially on multiview, which is very narrow despite quadrupling the budget.
Overall, learned detectors -- KeyNet, SuperPoint, R2D2, LF-Net -- show relatively better results on multiview setup than on stereo. Our hypothesis is that they have good robustness, but low localization precision, \revision{which is later corrected during bundle adjustment}.
\definecolor{light-gray}{gray}{0.9}
\renewcommand{\arraystretch}{0.95}
\renewcommand{\tabcolsep}{0.8mm}
\begin{table}
\begin{center}
\begin{adjustbox}{width=\linewidth,center}
\footnotesize
\begin{tabular}{@{}lcccccccc@{}}
\toprule
\multicolumn{2}{c}{} & \multicolumn{2}{c}{PyRANSAC} & \multicolumn{2}{c}{DEGENSAC} & \multicolumn{2}{c}{MAGSAC} & \\
Method & NF & NI$^\uparrow$ & mAA(10$^o$)$^\uparrow$ & NI$^\uparrow$ & mAA(10$^o$)$^\uparrow$ & NI$^\uparrow$ & mAA(10$^o$)$^\uparrow$ & Rank \\
\midrule
CV-SIFT & 2048.0 & 84.9 & .2489 & 79.0 & .2875 & 99.2 & .2805 & 12 \\
\midrule
CV-$\sqrt{\text{SIFT}}$ & 2048.0 & 84.2 & .2724 & 88.3 & .3149 & 106.8 & .3125 & 10 \\
CV-SURF & 2048.0 & 37.9 & .1725 & 72.7 & .2086 & 87.0 & .2081 & 15 \\
CV-AKAZE & 2048.0 & 96.1 & .1780 & 91.0 & .2144 & 115.5 & .2127 & 14 \\
CV-ORB & 2031.8 & 56.3 & .0610 & 63.5 & .0819 & 71.5 & .0765 & 19 \\
CV-FREAK & 2048.0 & 62.5 & .1461 & 65.6 & .1761 & 78.4 & .1698 & 17 \\
\midrule
L2-Net & 1936.3 & 66.1 & .3131 & 92.4 & .3752 & 114.7 & .3691 & 6 \\
DoG-HardNet & 1936.3 & 111.9 & .3508 & 117.7 & .4029 & 150.5 & .4033 & 5 \\
Key.Net-HardNet & 2048.0 & 134.4 & .3272 & 174.8 & {\bf \textcolor{red}{.4139}} & 228.4 & .3897 & 1 \\
Key.Net-SOSNet & 2048.0 & 88.0 & .3279 & 171.9 & {\bf \textcolor{Green}{.4132}} & 212.9 & .3928 & 2 \\
GeoDesc & 1936.3 & 98.9 & .3127 & 103.9 & .3662 & 129.7 & .3640 & 7 \\
ContextDesc & 2048.0 & 118.8 & .2965 & 124.1 & .3510 & 146.4 & .3485 & 9 \\
DoG-SOSNet & 1936.3 & 111.1 & .3536 & 132.1 & .3976 & 149.6 & .4092 & 4 \\
LogPolarDesc & 1936.3 & 118.8 & .3569 & 124.9 & {\bf \textcolor{NavyBlue}{.4115}} & 161.0 & .4064 & 3 \\
\midrule
D2-Net (SS) & 2045.6 & 107.6 & .1157 & 134.8 & .1355 & 259.3 & .1317 & 18 \\
D2-Net (MS) & 2038.2 & 149.3 & .1524 & 188.4 & .1813 & 302.9 & .1703 & 16 \\
LF-Net & 2020.3 & 100.2 & .1927 & 106.5 & .2344 & 141.0 & .2226 & 13 \\
SuperPoint & 2048.0 & 120.1 & .2577 & 126.8 & .2964 & 127.3 & .2676 & 11 \\
R2D2 (wasf-n16) & 2048.0 & 191.0 & .2829 & 215.6 & .3614 & 215.6 & .3614 & 8 \\
\midrule
DoG-AffNet-HardNet & 2047.8 & 105.6 & .3589 & 152.1 & {\bf \textcolor{red}{.4197}} & 195.2 & {\bf \textcolor{Green}{.4175}} & $1^{*}$ \\
\bottomrule
\end{tabular}
\end{adjustbox}
\end{center}
\caption{
{\bf Test -- Stereo results with 2k features.}
Same as \tbl{test_stereo}.
 $^{*}$ {\bf The last group of results is obtained after the paper publication and not described in the text of the paper. Their rank does not influence other entries ranks.}
}
\label{tbl:test_stereo_2k}
\end{table}

\renewcommand{\arraystretch}{0.6}
\renewcommand{\tabcolsep}{1.2mm}
\begin{table}
\begin{center}
\begin{adjustbox}{width=\linewidth,center}
\footnotesize
\begin{tabular}{@{}lcccccccc@{}}
\toprule
Method & NL$^\uparrow$ & SR$^\uparrow$ & RC$^\uparrow$ & TL$^\uparrow$ & mAA(5$^o$)$^\uparrow$ & mAA(10$^o$)$^\uparrow$ & ATE$^\downarrow$ & Rank \\
\midrule
CV-SIFT & 1081.2 & 87.6 & 87.4 & 3.70 & .3718 & .4562 & .6136 & 13 \\
\midrule
CV-$\sqrt{\text{SIFT}}$ & 1174.7 & 90.3 & 89.4 & 3.82 & .4074 & .4995 & .5589 & 12 \\
CV-SURF & 1186.6 & 90.2 & 88.6 & 3.55 & .3335 & .4184 & .6701 & 15 \\
CV-AKAZE & 1383.9 & 94.7 & 90.9 & 3.74 & .3393 & .4361 & .6422 & 14 \\
CV-ORB & 683.3 & 74.9 & 73.0 & 3.21 & .1422 & .1914 & .8153 & 19 \\
CV-FREAK & 1075.2 & 87.2 & 86.3 & 3.52 & .2578 & .3297 & .7169 & 17 \\
\midrule
L2-Net & 1253.3 & 94.7 & 92.6 & 3.96 & .4369 & .5392 & .5419 & 9 \\
DoG-HardNet & 1338.2 & 96.3 & 93.7 & 4.03 & .4624 & .5661 & .5093 & 6 \\
Key.Net-HardNet & 1276.3 & 97.8 & {\bf \textcolor{NavyBlue}{95.7}} & {\bf \textcolor{red}{4.49}} & {\bf \textcolor{Green}{.5050}} & {\bf \textcolor{Green}{.6161}} & {\bf \textcolor{Green}{.4902}} & 2 \\
Key.Net-SOSNet & 1475.5 & {\bf \textcolor{Green}{99.3}} & {\bf \textcolor{red}{96.5}} & {\bf \textcolor{Green}{4.42}} & {\bf \textcolor{red}{.5229}} & {\bf \textcolor{red}{.6340}} & {\bf \textcolor{red}{.4853}} & 1 \\
GeoDesc & 1133.6 & 93.6 & 91.3 & 4.02 & .4246 & .5244 & .5455 & 10 \\
ContextDesc & 1504.9 & 95.6 & 93.3 & 3.92 & .4529 & .5568 & .5327 & 7 \\
DoG-SOSNet & 1317.4 & 96.0 & 93.8 & 4.05 & .4739 & .5784 & .5194 & 5 \\
LogPolarDesc & 1410.2 & 96.0 & 93.8 & 4.05 & .4794 & .5849 & .5090 & 4 \\
\midrule
D2-Net (SS) & {\bf \textcolor{red}{2357.9}} & {\bf \textcolor{NavyBlue}{98.9}} & 94.7 & 3.39 & .2875 & .3943 & .7010 & 16 \\
D2-Net (MS) & {\bf \textcolor{Green}{2177.3}} & 98.2 & 93.4 & 3.01 & .1921 & .3007 & .7861 & 20 \\
LF-Net & 1385.0 & 95.6 & 90.4 & 4.14 & .4156 & .5141 & .5738 & 11 \\
SuperPoint & 1184.3 & 95.6 & 92.4 & {\bf \textcolor{NavyBlue}{4.34}} & .4423 & .5464 & .5457 & 8 \\
R2D2 (wasf-n16) & 1228.4 & {\bf \textcolor{red}{99.4}} & {\bf \textcolor{Green}{96.2}} & 4.29 & {\bf \textcolor{NavyBlue}{.5045}} & {\bf \textcolor{NavyBlue}{.6149}} & {\bf \textcolor{NavyBlue}{.4956}} & 3 \\
\midrule
DoG-AffNet-HardNet & {\bf \textcolor{NavyBlue}{1788.7}} & 98.7 & {\bf \textcolor{NavyBlue}{95.7}} & 4.19 & .4771 & .5854 & .5114 & $4^{*}$\\ 
\bottomrule
\end{tabular}
\end{adjustbox}
\end{center}
\caption{
{\bf Test -- Multiview results with 2k features.}
Same as \tbl{test_multiview}.
 $^{*}$ {\bf The last group of results is obtained after the paper publication and not described in the text of the paper. Their rank does not influence other entries ranks.}
}
\label{tbl:test_multiview_2k}
\end{table}


\subsection{Outlier pre-filtering with deep networks \edit{--- \tbl{test_cne}}}

Next, we study the performance of CNe~\cite{Yi18a} for outlier rejection, paired with PyRANSAC, DEGENSAC, and MAGSAC.
Its training data does not use the ratio test, so we omit it here too -- note that because of this, it expects a relatively large number of input matches. 
We thus evaluate it only for the 8k feature setting, \edit{while using the ``both'' matching strategy}.

\definecolor{light-gray}{gray}{0.9}
\renewcommand{\arraystretch}{0.9}
\renewcommand{\tabcolsep}{0.3mm}
\begin{table}
\begin{center}
\begin{adjustbox}{width=\linewidth,center}
\footnotesize
\begin{tabular}{@{}lcccccccc@{}}
\toprule
& \multicolumn{6}{c}{Stereo Task} & \multicolumn{2}{c}{\multirow{2}{*}{Multi-view Task}} \\
& \multicolumn{2}{c}{PyRANSAC} & \multicolumn{2}{c}{DEGENSAC} & \multicolumn{2}{c}{MAGSAC} & \\
Method & mAA(10$^o$)$^\uparrow$ & $\Delta$(\%)$^\uparrow$ & mAA(10$^o$)$^\uparrow$ & $\Delta$(\%)$^\uparrow$ & mAA(10$^o$)$^\uparrow$ & $\Delta$(\%)$^\uparrow$ & mAA(10$^o$)$^\uparrow$ & $\Delta$(\%)$^\uparrow$ \\
\midrule
CV-SIFT & .4086 & +2.24 & .4751 & +3.65 & .4694 & +2.42 & .6815 & +8.85 \\
\midrule
CV-$\sqrt{\text{SIFT}}$ & .4205 & -0.53 & .4927 & -0.06 & .4848 & -1.87 & {\bf \textcolor{NavyBlue}{.6978}} & +3.16 \\
CV-SURF & .2490 & +9.18 & .3071 & +18.42 & .2954 & +15.75 & .5750 & +18.67 \\
CV-AKAZE & .2857 & +11.18 & .3417 & +11.18 & .3316 & +9.23 & .6026 & +8.51 \\
CV-ORB & .1323 & +8.49 & .1856 & +10.87 & .1748 & +11.34 & .4171 & +18.88 \\
CV-FREAK & .2532 & +11.36 & .3204 & +18.18 & .3053 & +14.93 & .5574 & +19.79 \\
\midrule
L2-Net & .4377 & -5.27 & .5012 & -5.35 & .4937 & -5.99 & .6951 & +4.62 \\
DoG-HardNet & .4427 & -7.80 & {\bf \textcolor{Green}{.5156}} & -6.98 & .5056 & -8.11 & {\bf \textcolor{Green}{.7061}} & -.50 \\
Key.Net-HardNet & .3081 & -22.92 & .4226 & -15.23 & .4012 & -15.36 & .6620 & +2.11 \\
GeoDesc & .4239 & -2.05 & .4924 & -3.67 & .4807 & -4.93 & .6956 & +2.25 \\
ContextDesc & .3976 & -15.11 & .4482 & -12.09 & .4535 & -11.83 & .6900 & -1.91 \\
DoG-SOSNet & .4439 & -7.21 & {\bf \textcolor{red}{.5187}} & -7.15 & {\bf \textcolor{NavyBlue}{.5073}} & -8.04 & {\bf \textcolor{red}{.7103}} & +1.18 \\
LogPolarDesc & .4259 & -6.89 & .4898 & -8.27 & .4808 & -8.22 & .6871 & -.82 \\
\midrule
D2-Net (SS) & .1231 & -36.32 & .1717 & -22.95 & .1608 & -20.86 & .4639 & +0.89 \\
D2-Net (MS) & .0998 & -53.78 & .1370 & -45.33 & .1316 & -43.29 & .4132 & -13.02 \\
R2D2 (wasf-n8-big) & .2218 & -39.78 & .3141 & -29.21 & .3032 & -28.43 & .6229 & -8.83 \\
\bottomrule
\end{tabular}
\end{adjustbox}
\end{center}
\caption{
{\bf Test -- Outlier pre-filtering with CNe (8k features).}
We report mAP at 10$^o$ with CNe, on stereo and multi-view, and its increase in performance w.r.t. \tbl{test_multiview} -- positive $\Delta$ meaning CNe helps.
When using CNe, we disable the ratio test.
}
\label{tbl:test_cne}
\end{table}

Our experiments with SIFT, \edit{the local feature used to train CNe}, are encouraging: CNe aggressively filters out about 80\% of the matches in a single forward pass, boosting mAA at 10$^\circ$ by \edit{2-4\%} relative for stereo task and 8\% for multiview task.
In fact, it is surprising that nearly all classical methods benefit from it, with gains of up to \et{20\%} relative.
By contrast, it damages performance with most learned descriptors, even those operating on DoG keypoints, and particularly for methods learned end-to-end, such as D2-Net and R2D2.
We hypothesize this might be because the models performed better on the ``classical'' keypoints it was trained with
-- \cite{Sun19} reports that re-training them for a specific feature helps.

\renewcommand{\arraystretch}{1}
\renewcommand{\tabcolsep}{0.9mm}
\begin{table}
\begin{center}
\tiny
\begin{tabular}{@{}lcccccccc@{}}
\toprule
& \multicolumn{2}{c}{CV-$\sqrt{\text{SIFT}}$} & \multicolumn{2}{c}{HardNet} & \multicolumn{2}{c}{SOSNet} & \multicolumn{2}{c}{LogPolarDesc} \\
& NI$^\uparrow$ & mAA(10$^o$)$^\uparrow$ & NI$^\uparrow$ & mAA(10$^o$)$^\uparrow$ & NI$^\uparrow$ & mAA(10$^o$)$^\uparrow$ & NI$^\uparrow$ & mAA(10$^o$)$^\uparrow$ \\
\midrule
Standard & 281.7 & 0.4930 & 432.3 & 0.5543 & 424.6 & 0.5587 & 441.8 & 0.5340 \\
\midrule
Upright & 270.0 & 0.4878 & 449.2 & 0.5542 & 432.9 & 0.5554 & 461.8 & 0.5409 \\
$\Delta$ (\%) & -4.15 & -1.05 & +3.91 & -0.02 & +1.95 & -0.59 & +4.53 & +1.29 \\
\midrule
Upright++ & 358.9 & 0.5075 & 527.6 & 0.5728 & 508.4 & 0.5738 & 543.2 & 0.5510 \\
$\Delta$ (\%) & +27.41 & +2.94 & +22.04 & +3.34 & +19.74 & +2.70 & +22.95 & +3.18 \\
\bottomrule
\end{tabular}
\end{center}
\caption{
{\bf Test -- Stereo performance with upright descriptors (8k features).}
\edit{We report {\bf (NI)} the number of inliers and {\bf mAA} at 10$^o$ for the stereo task, using DEGENSAC. As DoG may return multiple orientations for the same point~\cite{Lowe04} (up to 30\%), we report: {\bf (top)} with orientation estimation; {\bf (middle)} setting the orientation to zero while removing duplicates; and {\bf (bottom)} adding new points until hitting the 8k-feature budget.}
}
\label{tbl:test_upright}
\end{table}
\renewcommand{\arraystretch}{1}
\renewcommand{\tabcolsep}{0.9mm}
\begin{table}
\begin{center}
\tiny
\begin{tabular}{@{}lcccccccc@{}}
\toprule
& \multicolumn{2}{c}{CV-$\sqrt{\text{SIFT}}$} & \multicolumn{2}{c}{HardNet} & \multicolumn{2}{c}{SOSNet} & \multicolumn{2}{c}{LogPolarDesc} \\
& NL$^\uparrow$ & mAA(10$^o$)$^\uparrow$ & NL$^\uparrow$ & mAA(10$^o$)$^\uparrow$ & NL$^\uparrow$ & mAA(10$^o$)$^\uparrow$ & NL$^\uparrow$ & mAA(10$^o$)$^\uparrow$ \\
\midrule
Standard & 3312.1 & 0.6765 & 4001.4 & 0.7096 & 3796.0 & 0.7021 & 4054.6 & 0.6928 \\
\midrule
Upright & 3485.1 & 0.6572 & 3594.6 & 0.6962 & 4025.1 & 0.7054 & 3737.4 & 0.6934 \\
$\Delta$ (\%) & +5.22 & -2.85 & -10.17 & -1.89 & +6.04 & +0.47 & -7.82 & +0.09 \\
\midrule
Upright++ & 4404.6 & 0.6792 & 4250.4 & 0.7231 & 3988.6 & 0.7129 & 4414.1 & 0.7109 \\
$\Delta$ (\%) & +32.99 & +0.40 & +6.22 & +1.90 & +5.07 & +1.54 & +8.87 & +2.61 \\
\bottomrule
\end{tabular}
\end{center}
\caption{
{\bf Test -- Multiview performance with upright descriptors (8k features).}
Analogous to \tbl{test_upright}, on the multiview task. We report the number of landmarks (NL) instead of the number of inliers (NI).
}
\label{tbl:test_upright_multiview}
\end{table}
\renewcommand{\arraystretch}{1}
\renewcommand{\tabcolsep}{0.9mm}
\begin{table}
\begin{center}
\tiny
\begin{tabular}{@{}lcccccccc@{}}
\toprule
& \multicolumn{2}{c}{CV-$\sqrt{\text{SIFT}}$} & \multicolumn{2}{c}{HardNet} & \multicolumn{2}{c}{SOSNet} & \multicolumn{2}{c}{D2Net} \\
& NI$^\uparrow$ & mAA(10$^o$)$^\uparrow$ & NI$^\uparrow$ & mAA(10$^o$)$^\uparrow$ & NI$^\uparrow$ & mAA(10$^o$)$^\uparrow$ & NI$^\uparrow$ & mAA(10$^o$)$^\uparrow$ \\
\midrule
Exact & 281.7 & 0.4930 & 432.0 & 0.5532 & 424.3 & 0.5575 & 470.6 & 0.2506 \\
\midrule
FLANN & 274.6 & 0.4879 & 363.3 & 0.5222 & 339.8 & 0.5179 & 338.9 & 0.2046 \\
$\Delta$ (\%) & -2.52 & -1.03 & -15.90 & -5.60 & -19.92 & -7.10 & -27.99 & -18.36 \\
\bottomrule
\end{tabular}
\end{center}
\caption{
\edit{{\bf Test -- Stereo performance with OpenCV FLANN, using kd-tree approximate nearest neighbors (8k features).}
kd-tree parameters are: 4 trees, 128 checks. We report {\bf (NI)} the number of inliers and {\bf mAA} at 10$^o$ for the stereo task, using DEGENSAC. }}
\label{tbl:test_flann}
\end{table}
\subsection{On the effect of 
local feature 
orientation \edit{estimation --- Tables~\ref{tbl:test_upright}~and~\ref{tbl:test_upright_multiview}}}

In contrast with classical methods, which estimate the orientation of each keypoint, modern, end-to-end pipelines \cite{DeTone17b,Dusmanu19,Revaud19} often skip this step, assuming that the images are roughly aligned (upright), with the descriptor shouldering the increased invariance requirements.
As our images meet this condition, we experiment with setting the orientation of keypoints to a fixed value (zero).
DoG often returns multiple orientations for the same keypoint, so we consider two variants: one where we simply remove keypoints which become duplicates after setting the orientation to a constant value \edit{(Upright)}, and a second one where we fill out the budget with new keypoints \edit{(Upright++)}.
We list the results in \tbl{test_upright}, for stereo, and \tbl{test_upright_multiview}, for multiview.
Performance increases across the board \edit{with Upright++} -- albeit by a small margin.

\subsection{On the effect of approximate nearest neighbor matching \edit{--- \tbl{test_flann}}}
\label{sec:flann}

\edit{While it is known that approximate nearest neighbor search algorithms have non-\edit{perfect} recall~\cite{Muja09}, it is not clear how their usage influences downstream performance.
We thus compare exact (brute-force) nearest neighbor search with a popular choice for approximate nearest neighbor search, FLANN~\cite{Muja09}, as implemented in OpenCV.
We experimented with different parameters and found that 4 trees and 128 checks provide a reasonable trade-off between precision and runtime.
\edit{%
We report results with and without FLANN in \tbl{test_flann}.
}
The performance drop varies for different methods: from a moderate 1$\%$ for RootSIFT, to 5-7$\%$ HardNet and SOSNet, and 18\% for D2-Net.}

\begin{figure}
\centering
\renewcommand{\imsize}{5.3cm}
\renewcommand{\hspacer}{0.3mm}
\renewcommand{\vspacer}{0.1mm}
\renewcommand{\imscl}{0.01}
\includegraphics[scale=\imscl,width=\linewidth]{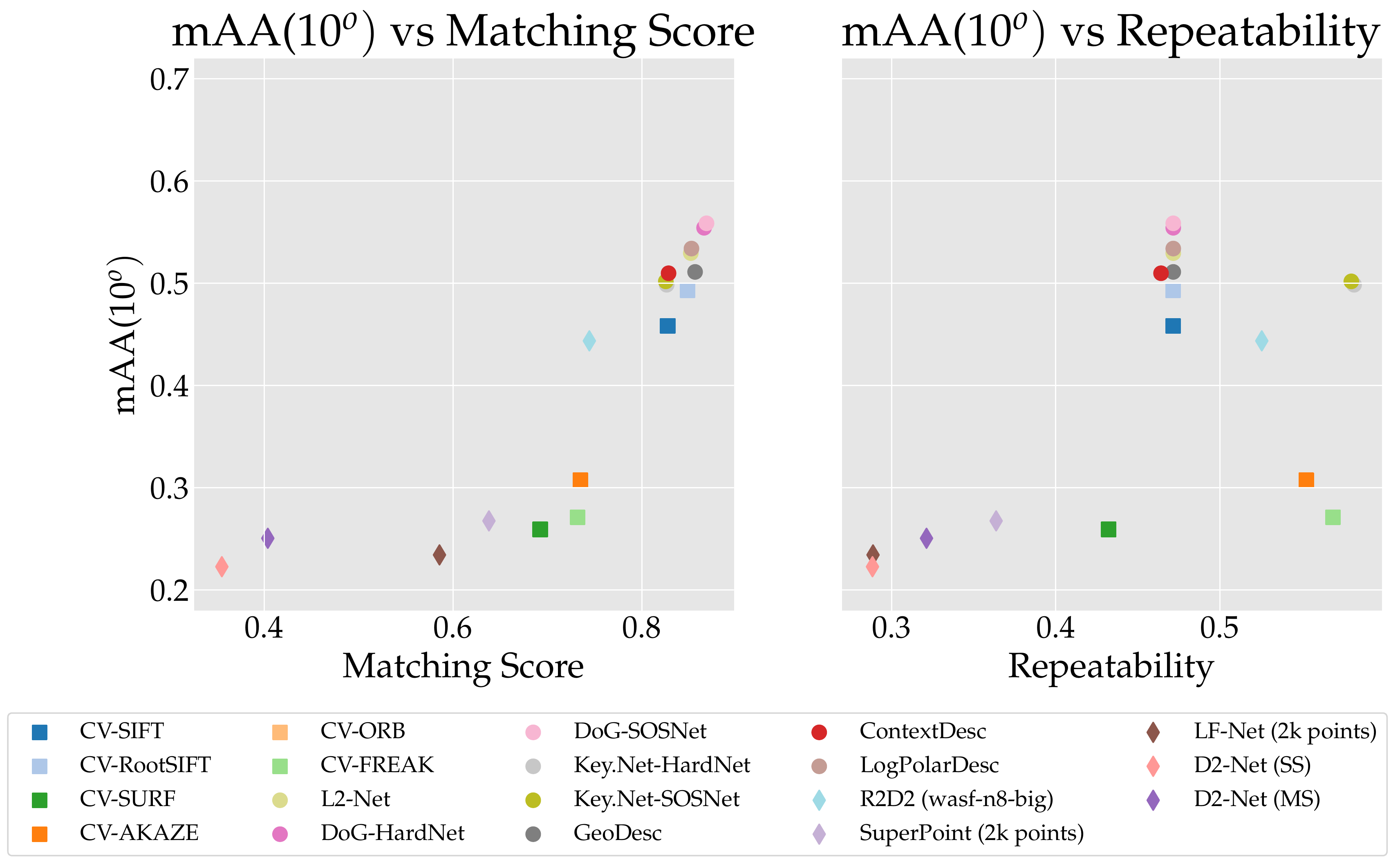}
\caption{{\bf Test -- Downstream vs. traditional metrics \revision{(8k features)}.}
We cross-reference stereo mAA at 10$^\circ$ with repeatability and matching score, with a 3-pixel threshold.
}
\label{fig:results_correlation}
\end{figure}

\subsection{Pose mAA vs. traditional metrics \edit{--- \fig{results_correlation}}}
\label{sec:metrics}

\begin{figure}
\centering
\includegraphics[width=\linewidth]{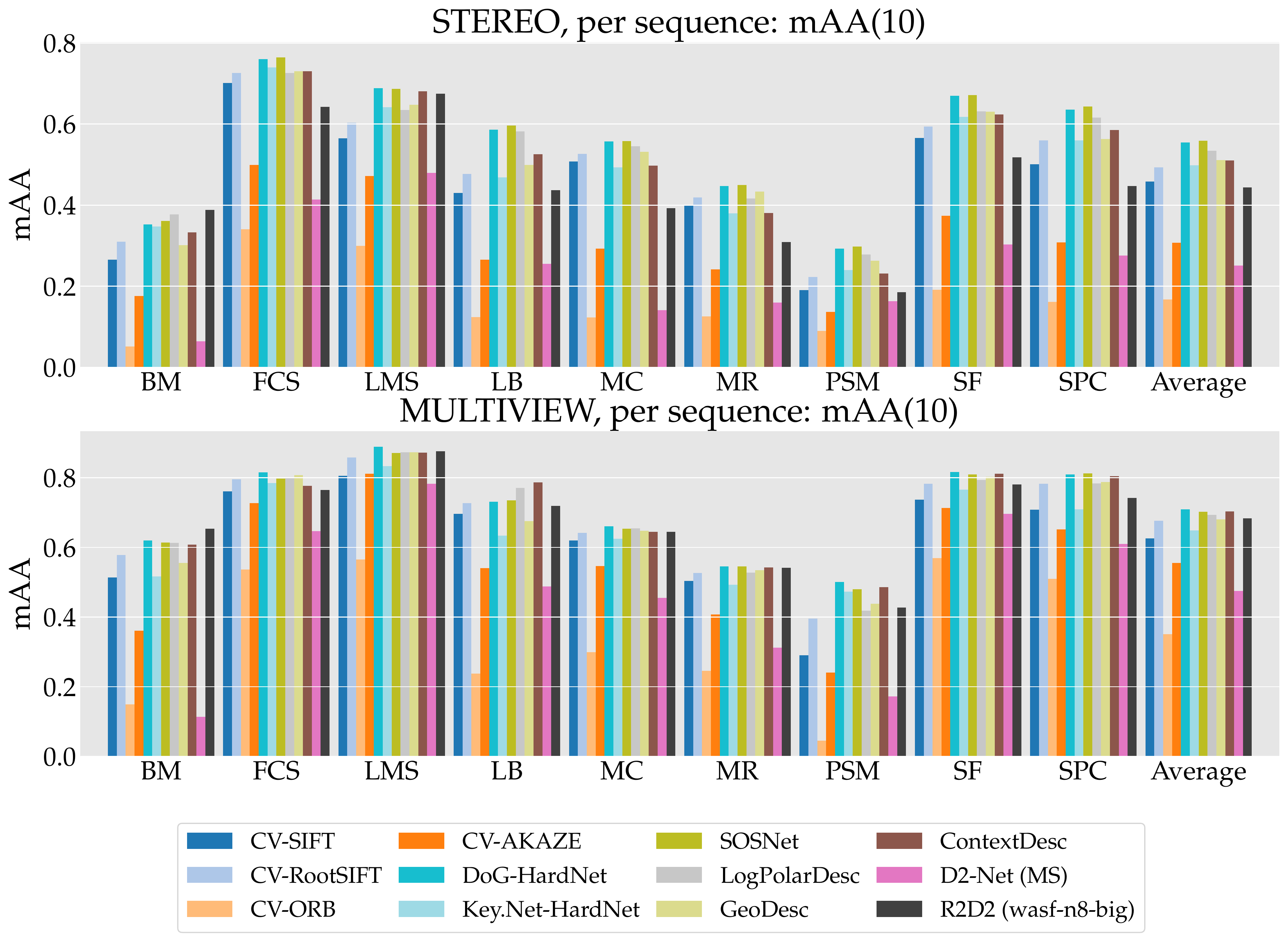}
\caption{
{\bf Test -- Breakdown by scene \revision{(8k features)}.}
\edit{For the stereo task, we use DEGENSAC.} Note how performance can vary drastically between scenes, and the relative rank of a given local feature fluctuates as well.
Please refer to \tbl{sequences_trainval} for a legend. 
}
\label{fig:sequences}
\vspace{-3mm}
\end{figure}

\edit{\edit{To examine the relationship between our pose metric and traditional metrics, we compare mAA against}
repeatability and matching score 
on the stereo task, with DEGENSAC.
While the matching score seems to correlate with mAA, repeatability is harder to interpret.
\edit{However, note that even for the matching score, which shows correlation, higher value does not guarantee high mAA -- see \eg RootSIFT vs ContextDesc.}
We \revision{remind} the reader that, as explained in \secref{benchmark}, our implementation differs from the classical formulation, as many methods do not have a strict notion of a support region.
We compute these metrics at a 3-pixel threshold, and provide more granular results in \secref{appendix}.}

\edit{\edit{As shown, }all methods based on DoG are clustered, as they operate on the same keypoints. 
Key.Net obtains the best repeatability, but performs worse than DoG in terms of mAA with the same descriptors (HardNet). 
AKAZE and FREAK perform surprisingly well in terms of repeatability -- \#2 and \#3, respectively -- but obtain a low mAA, which may be related to their descriptors, which are binary.
R2D2 shows good repeatability but a poor matching score and is outperformed by DoG-based features.}

\subsection{Breakdown by scene \edit{--- \fig{sequences}}}
Results may vary drastically from scene to scene, as shown in \fig{sequences}.
A given method may also perform better on some than others -- for instance, D2-Net nears the state of the art on ``Lincoln Memorial Statue'', but is 5x times worse on ``British Museum''.
AKAZE and ORB show similar behaviour.
This can provide valuable insights on limitations and failure cases.

\begin{figure}
\centering
\renewcommand{\imsize}{5.2cm}
\renewcommand{\hspacer}{0.3mm}
\renewcommand{\vspacer}{0.1mm}
\renewcommand{\imscl}{0.01}
\includegraphics[width=\linewidth]{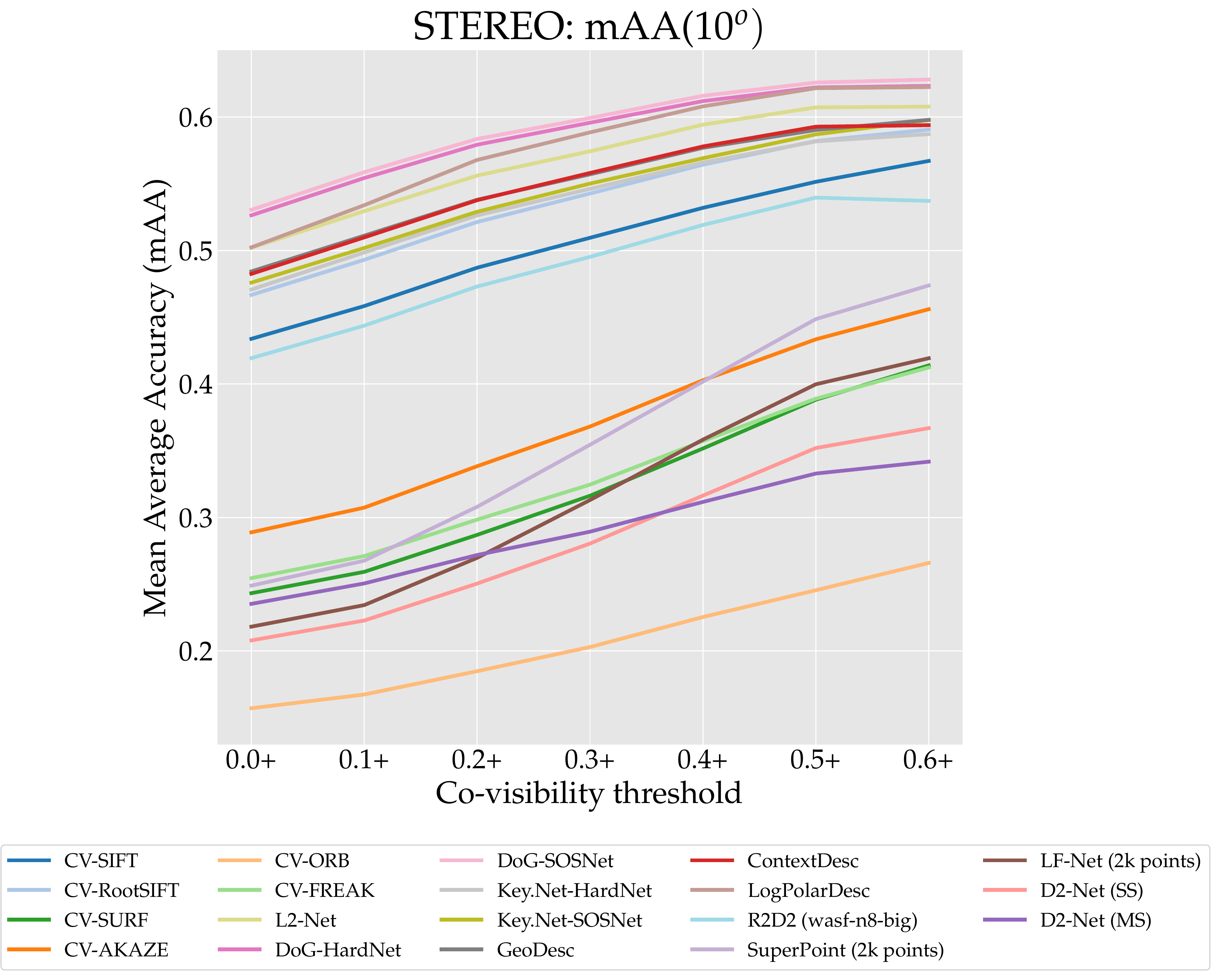}
\caption{{\bf Test -- Local features vs. co-visibility \revision{(8k features)}.}
\edit{We plot mAA at 10$^\circ$ on the stereo task -- using DEGENSAC -- at different co-visibility thresholds for different local feature types. 
Bin ``0+'' consists of all possible image pairs, including potentially unmatchable ones.
Bin ``0.1+'' includes pairs with a minimum co-visibility value of 0.1 -- this is the default value we use for all other experiments in this paper -- and so forth.
Results are mostly consistent, with end-to-end methods performing better at higher than lower co-visibility.}
}
\label{fig:results_visibility}
\end{figure}
\begin{figure}
\centering
\renewcommand{\imsize}{5.2cm}
\renewcommand{\hspacer}{0.3mm}
\renewcommand{\vspacer}{0.1mm}
\renewcommand{\imscl}{0.01}
\includegraphics[width=\linewidth]{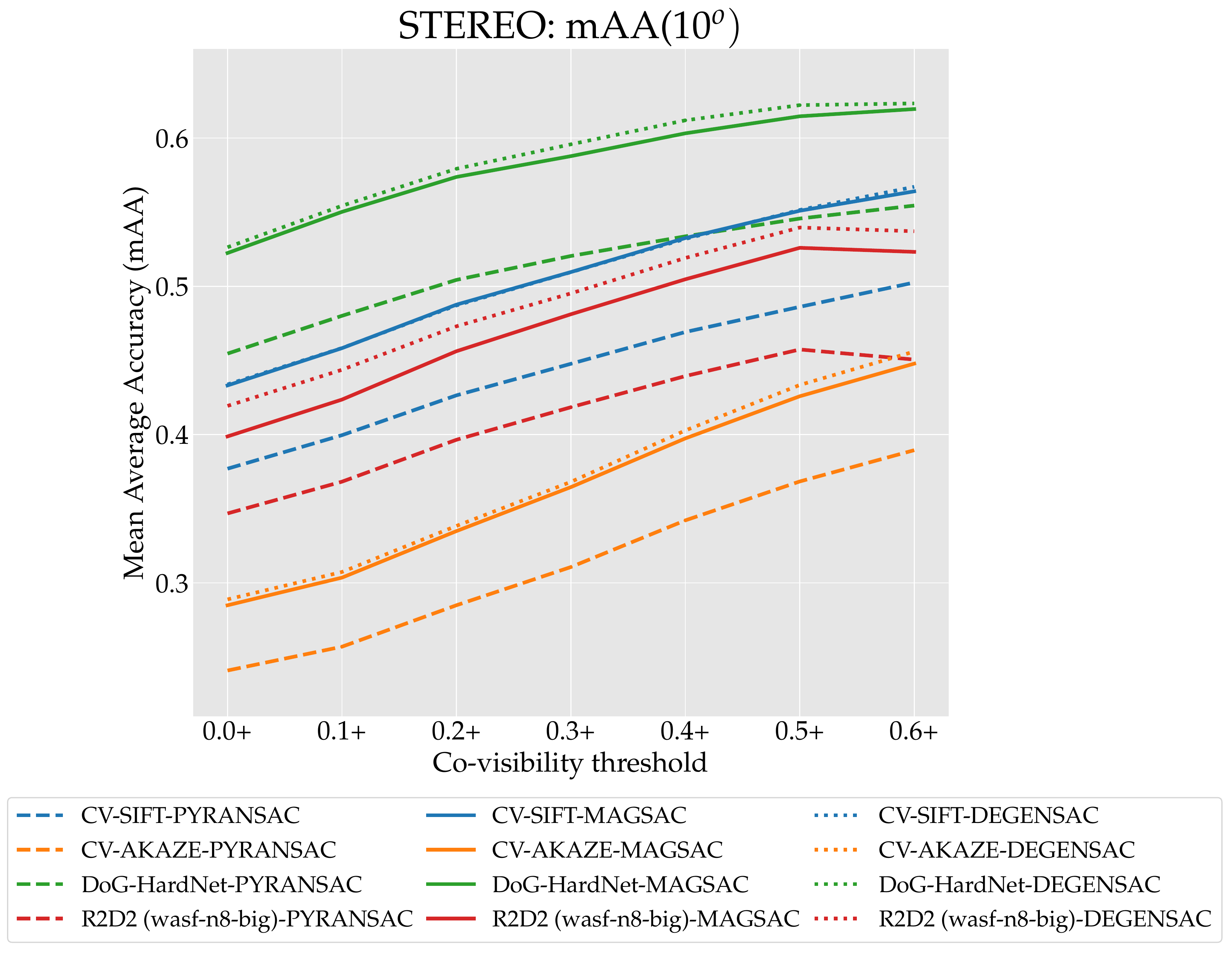}
\caption{{\bf Test -- RANSAC vs. co-visibility \revision{(8k features)}.} \edit{We plot mAA at 10$^\circ$ on the stereo task at different co-visibility thresholds for different RANSAC variants, binning the results as in \fig{results_visibility}. 
We pair them with different local features methods.
The difference in performance between RANSAC variants seems consistent across pairs, regardless of their difficulty.}
}
\label{fig:results_visibility_ransac}
\end{figure}

\subsection{Breakdown by co-visibility 
\edit{--- Figs.~\ref{fig:results_visibility}~and~\ref{fig:results_visibility_ransac}}}

\edit{\edit{Next, in \fig{results_visibility} we evaluate stereo performance at different co-visibility thresholds, for several local feature methods, using DEGENSAC.}
Bins are encoded as $v$+, with $v$ the co-visibility threshold, and include all image pairs with a co-visibility value larger or equal than $v$. 
This means that the first bin may contain unmatchable images -- we use 0.1+ for all other experiments in the paper. 
We do not report values above 0.6+ as 
\edit{there are only a handful of them and thus the results are very noisy.}

Performance for all local features and RANSAC variants increases with the co-visibility threshold, as expected.
Results are consistent, with end-to-end methods performing better at higher than \edit{co-visibility} than lower, and single-scale D2-Net outperforming its multi-scale counterpart at 0.4+ and above, where the images are more likely aligned in terms of scale.

We also break down different RANSAC methods in \fig{results_visibility_ransac}, along with different local fetures, including AKAZE (binary), SIFT (also handcrafted), HardNet (learned descriptor on DoG points) and R2D2 (learned end-to-end). 
We do not observe significant variations \edit{in the trend of each curve as we \edit{swap} RANSAC and local feature methods}.
DEGENSAC and MAGSAC show very similar performance.}

\begin{figure}
\centering
\renewcommand{\hspacer}{0.3mm}
\renewcommand{\vspacer}{0.1mm}
\renewcommand{\imscl}{0.01}
\includegraphics[scale=\imscl,width=\linewidth]{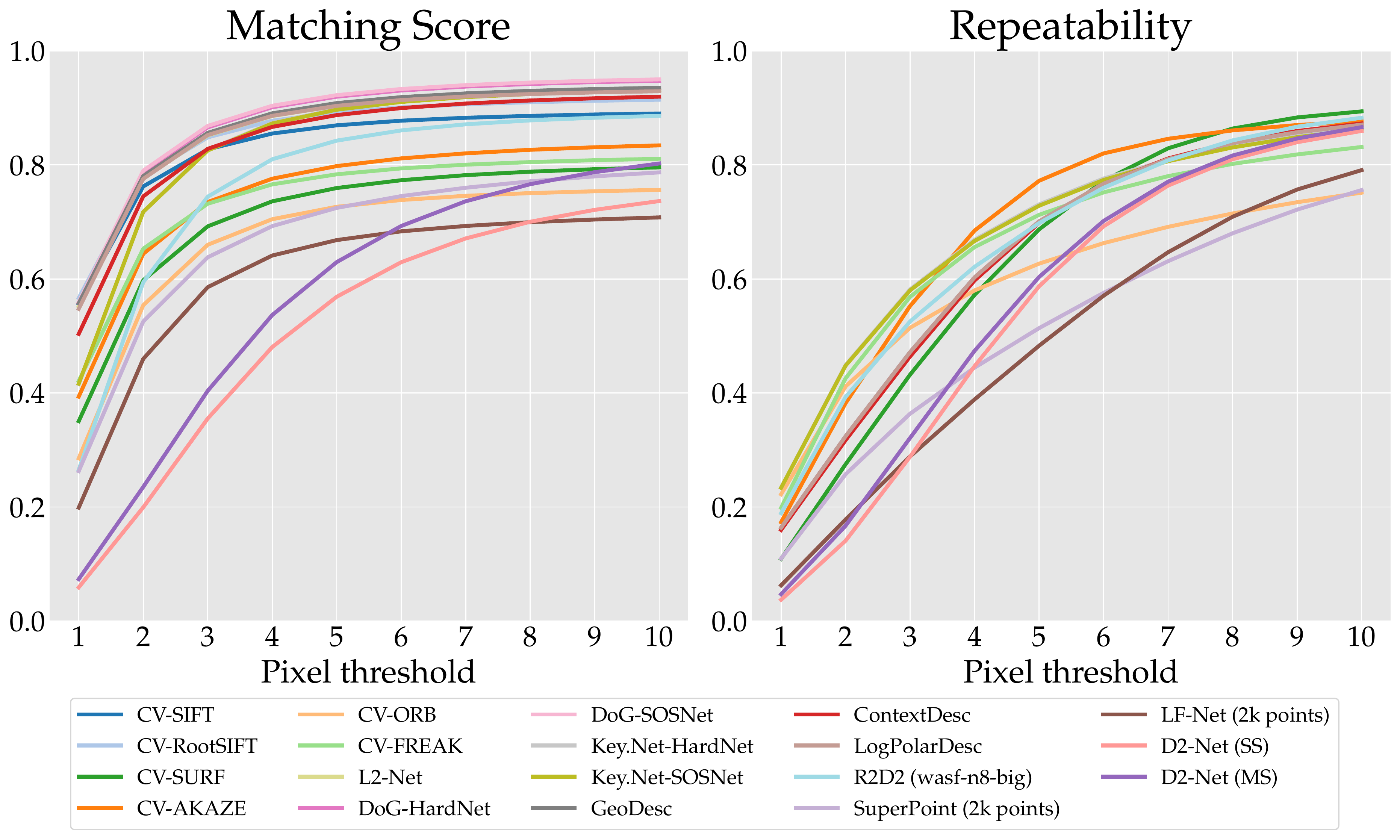}\\
\caption{{\bf Test -- Classical metrics, by pixel threshold \revision{(8k features)}.} Repeatability and matching score computed at different pixel thresholds are shown.}
\label{fig:results_repeatability}
\end{figure}
\subsection{%
\edit{Classical metrics vs.}
\edit{pixel}
threshold --- \fig{results_repeatability}}

\edit{In \fig{results_correlation} we plot repeatability and matching score against mAA at a fixed error threshold of 3 pixels. 
In \fig{results_repeatability} we show them at different pixel thresholds. 
End-to-end methods tend to perform better at higher pixel thresholds, which is expected -- D2-Net in particular extracts keypoints from downsampled feature maps. 
These results are computed from the depth maps estimated by COLMAP, which are not pixel-perfect, so the results for very low thresholds are not completely trustworthy.

Note that repeatability is typically lower than matching score, which might be counter-intuitive as the latter is more strict -- it requires two features to be nearest-neighbors in descriptor space in addition to being physically close (after reprojection). 
We compute repeatability with the raw set of keypoints, whereas the matching score is computed with optimal matching settings -- bidirectional matching with the ``both'' strategy and the ratio test. This results in a much smaller pool -- 8k features are typically narrowed down to \edit{200--400} matches (see \tbl{results_supp_inliers}).
This better isolates the performance of the detector and the descriptor where it matters.}

\begin{figure}
\centering
\includegraphics[width=\linewidth]{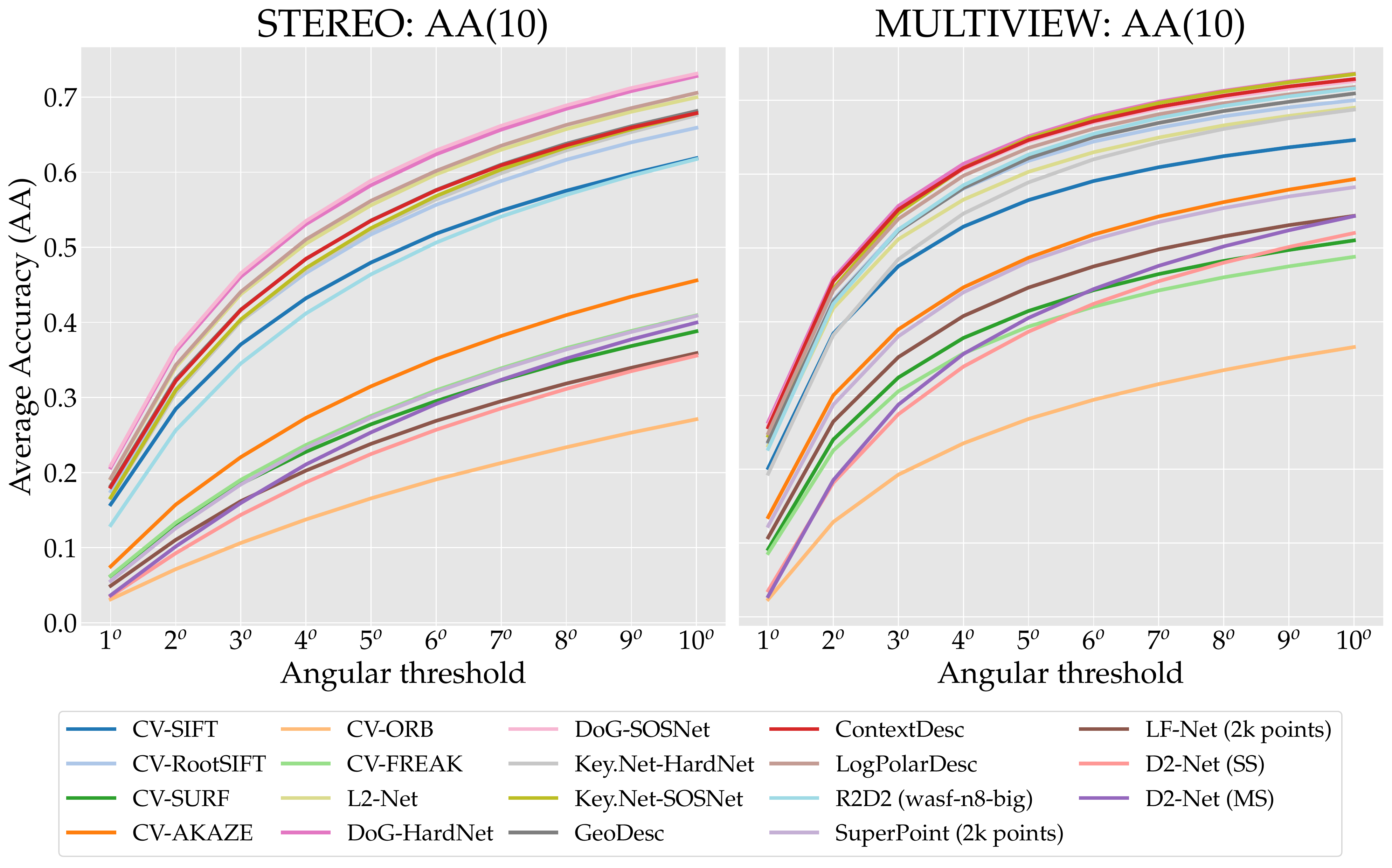}
\caption{
\textbf{Test -- Breakdown by angular threshold, for local features \revision{(8k features)}.}
\edit{Accuracy in pose estimation is evaluated at every error threshold. Mean Average Accuracy used in the rest of paper is the area under this curve. Ranks are consistent across thresholds.}
}
\label{fig:map_granular_features}
\end{figure}
\begin{figure}
\centering
\includegraphics[width=\linewidth]{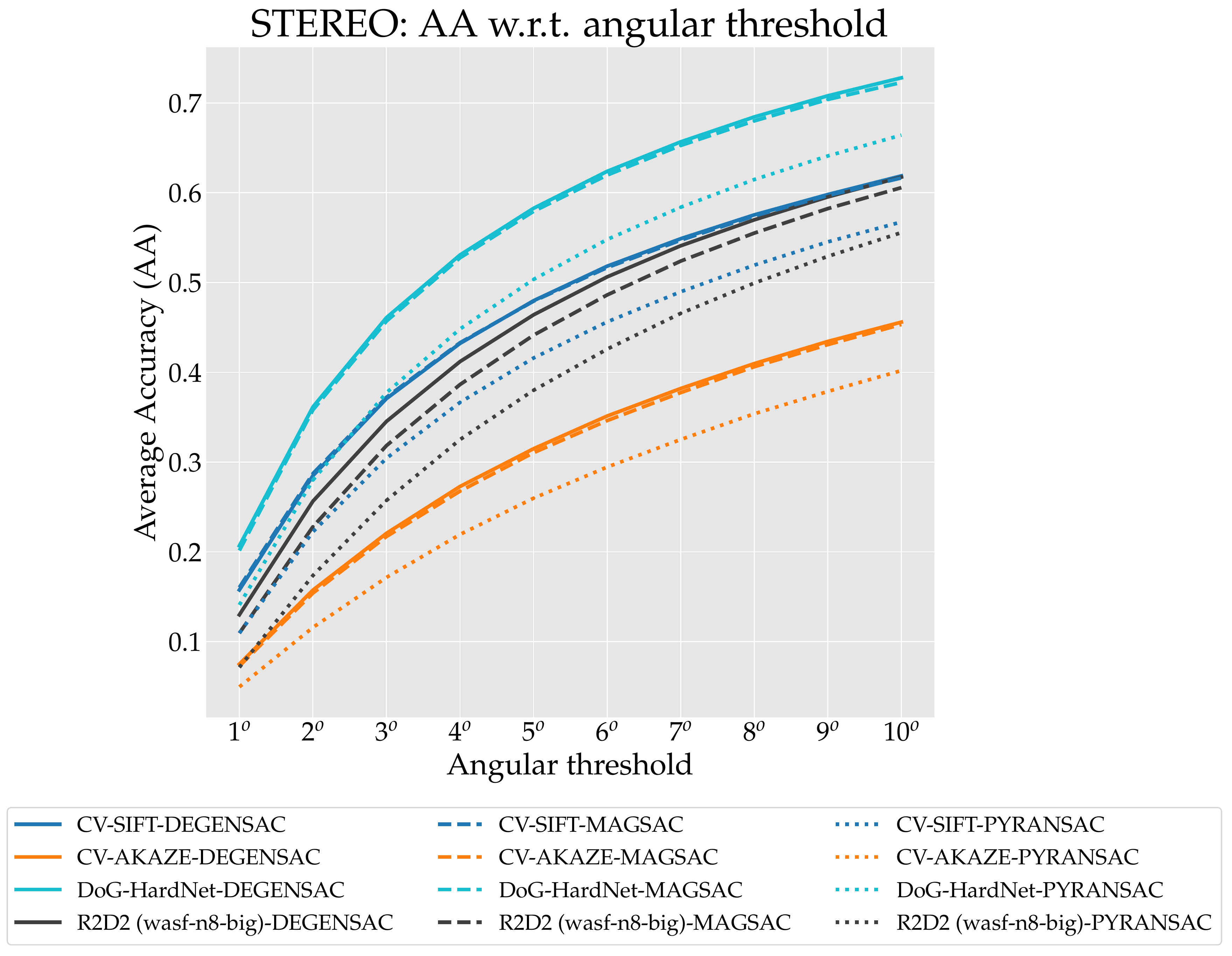}
\caption{
{\bf Test -- Breakdown by angular threshold, for RANSAC \revision{(8k features)}.}
\edit{We plot performance on the stereo task at different angular error thresholds, for \edit{four} different local features and three RANSAC variants: MAGSAC (solid line), DEGENSAC (dashed line), and PyRANSAC (dotted line).
We observe a similar behaviour across all thresholds.}
}
\label{fig:map_granular_geom}
\end{figure}
\subsection{Breakdown by 
\edit{angular}
threshold --- Figs.~\ref{fig:map_granular_features}~and~\ref{fig:map_granular_geom}}

We summarize pose accuracy by mAA at 10$^\circ$ in order to have a single, 
easy-to-interpret number.
In this section we show how performance varies across different error thresholds -- we look at the {\em average accuracy} at every angular error threshold, rather than the mean Average Accuracy. \fig{map_granular_features} plots performance on stereo and multiview for different local feature types, showing that the ranks remain consistent across thresholds.
\fig{map_granular_geom} shows how it affects different RANSAC variants, with four different local features.
Again, ranks do not change. DEGENSAC and MAGSAC perfom nearly identically for all features, except for R2D2.
The consistency in the ranks demonstrate that summarizing the results with a single number, mAA at 10$^\circ$, is reasonable.
\subsection{Qualitative results --- Figs.~\ref{fig:supp_stereo},~\ref{fig:supp_stereo2},~\ref{fig:supp_multiview},~and~\ref{fig:supp_multiview2}}
Figs.~\ref{fig:supp_stereo}~and~\ref{fig:supp_stereo2} show qualitative results for the stereo task. 
We draw the inliers produced by DEGENSAC and color-code them using the ground truth depth maps, from green (0) to yellow (5 pixels off) if they are correct, and in red if they are incorrect (more than 5 pixels off). 
Matches including keypoints which fall on occluded pixels are drawn in blue. 
Note that while the depth maps are somewhat noisy and not pixel-accurate, they are sufficient for this purpose.
\edit{Notice how the best performing methods have more correct matches that are well spread across the overlapping region.}

\fig{supp_multiview} shows qualitative results for the multiview task, for handcrafted detectors. 
We illustrate it by drawing keypoints used in the SfM reconstruction in blue and the rest in red. 
It showcases the importance of the detector, specially on unmatchable regions such as the sky. 
ORB and Hessian keypoints are too concentrated in the high-contrast regions, failing to provide evenly-distributed features.
In contrast, SURF fails to filter-out background and sky points. 
\edit{\fig{supp_multiview2} shows results for learned methods: DoG+HardNet, Key.Net+HardNet, SuperPoint, R2D2, and D2-Net (multiscale). 
They have different detection patterns: Key.Net resembles a ``cleaned'' version of DoG, while R2D2 seems to be evenly distributed. D2Net looks rather noisy, while SuperPoint fires precisely on corner points on structured parts, and sometimes form a regular grid on sky-like homogeneous regions, which might be due to \ky{the method} running out of locations to place points at -- note however that the best results were obtained with larger NMS (non-maxima suppression) thresholds}.

\section{Further results and considerations}
\label{sec:appendix}

In this section we provide additional results on the validation set.
\edit{These include}
a study of the typical outlier ratios under optimal settings in \secref{appendix-inliers}, matching with FGINN in \secref{fginn}, image-preprocessing techniques for feature extraction in \secref{appendix-clahe}, and a breakdown of the optimal settings in \secref{optimal-settings}, provided \edit{to serve as a} reference.

\renewcommand{\arraystretch}{1}
\renewcommand{\tabcolsep}{1.2mm}
\begin{table}
\small
\begin{center}
\begin{tabular}{@{}lcccc@{}}
\toprule
Method & \# matches & \# inliers & Ratio (\%) & mAA(10$^\circ$) \\
\midrule
CV-SIFT & 328.3 & 113.0 & 34.4 & 0.548 \\
\midrule
CV-$\sqrt{\text{SIFT}}$& 331.6& 131.4& 39.6& 0.584 \\
CV-SURF& 221.5& 77.4& 35.0& 0.309 \\
CV-AKAZE& 369.1& 143.5& 38.9& 0.360 \\
CV-ORB& 193.4& 74.7& 38.6& 0.265 \\
CV-FREAK & 216.6 & 75.5 & 34.8 & 0.329 \\
\midrule
DoG-HardNet& 433.1& 173.1& 40.0& 0.627 \\
Key.Net-HardNet & 644.1 & 166.6 & 25.9 & 0.580 \\
L2Net& 368.0& 144.0& 39.1& 0.601 \\
GeoDesc& 322.6& 133.4& 41.3& 0.564 \\
ContextDesc & 560.8 & 165.9 & 29.6 & 0.587 \\
SOSNet& 391.7& 161.8& 41.3& 0.621 \\
LogPolarDesc& 522.8& 175.2& 33.5& 0.618 \\
\midrule
SuperPoint (2k points)& 202.0& 62.6& 31.0& 0.312 \\
LF-Net (2k points)& 165.9& 73.2& 44.1& 0.293 \\
D2-Net (SS)& 657.3& 148.9& 22.7& 0.263 \\
D2-Net (MS)& 601.7& 161.7& 26.9& 0.343 \\
R2D2 (wasf-n8-big) & 901.5 & 477.3 & 52.9 & 0.473 \\
\bottomrule
\end{tabular}
\end{center}
\caption{
\edit{{\bf Validation -- Number of inliers with optimal settings.} We use 8k features with optimal parameters, and  PyRANSAC as a robust estimator. The number of inliers varies significantly between methods, despite tuning the matcher and the ratio test, and lower inlier ratios tend to correlate with low performance.}
}
\label{tbl:results_supp_inliers}
\end{table}
\subsection{Number of inliers per step --- \tbl{results_supp_inliers}}
\label{sec:appendix-inliers}

\edit{We list the number of input matches and their \emph{resulting} inliers for the stereo task, in \tbl{results_supp_inliers}.
\edit{As before, we remind the reader that these inliers are what each method reports, \ie, the \edit{matches} that are actually used to estimate the poses.}
We list the number of input matches, the number of inliers produced by each method (which may still contain outliers), their ratio, and the mAA at 10$^\circ$.
We use PyRANSAC with optimal settings for each method, the ratio test, and bidirectional matching with the ``both'' strategy.

We see that inlier-to-outlier ratios hover around 35--40\% for all features relying on classical detectors. 
Key.Net with HardNet descriptors sees a significant drop in inlier ratio and mAA, when compared to its DoG counterpart.
D2-Net similarly has inlier ratios around 25\%. 
R2D2 has the largest inlier ratio by far at 53\%, but is outperformed by many other methods in terms of mAA, suggesting that many of these are not actual inliers.
In general, we observe that the methods which produce a large number of matches, such as Key.Net (600+), D2-Net (600+) or R2D2 (900+) are less accurate in terms of pose estimation.}

\begin{figure}
\centering
\renewcommand{\imsize}{4.3cm}
\renewcommand{\hspacer}{0.3mm}
\renewcommand{\vspacer}{0.1mm}
\renewcommand{\imscl}{0.01}
\includegraphics[scale=\imscl,width=0.99\linewidth]{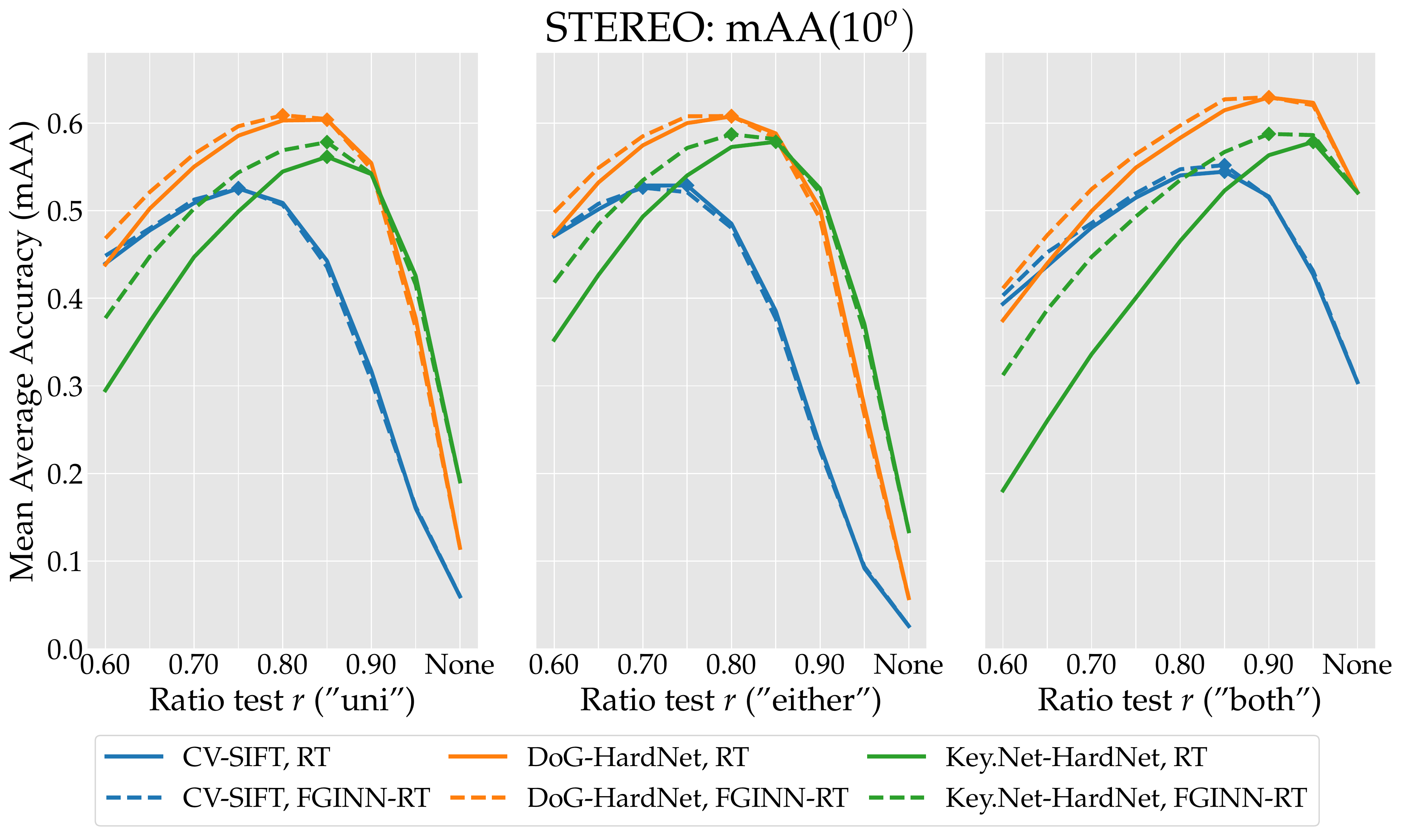}
\caption{{\bf Validation -- FGINN vs. ratio test  \revision{(8k features)}.} We evaluate the ratio test with FGINN~\cite{Mishkin15} (dashed line), and the standard ratio test (solid line). With FGINN, the valid range for $r$ -- the ratio test threshold -- is significantly wider, but the best performance with the ``both'' matching strategy is not significantly better than for the standard ratio test.
}
\label{fig:results_fginn}
\end{figure}
\subsection{Feature matching with an advanced ratio test --- \fig{results_fginn}}
\label{sec:fginn}

We also compare the benefits of applying first-geometrically-inconsistent-neighbor-ratio~(FGINN) \cite{Mishkin15} to DoG/SIFT, DoG/HardNet and Key.Net/HardNet, against Lowe's standard ratio test~\cite{Lowe04}.
FGINN performs the ratio-test with second-nearest neighbors that are ``far enough'' from the tentative match (10 pixels in \cite{Mishkin15}).
In other words, it loosens the test to allow for nearby-thus-similar points. 
We test it for 3 matching strategies: unidirectional (``uni''), ``both'' and ``either''.
We report the results in \fig{results_fginn}.
As shown, FGINN provides minor improvements over the standard ratio test in case of unidirectional matching, and not as much when ``both'' is used. 
It also behaves differently compared to the standard strategy, in that performance at stricter thresholds degrades less.

\begin{figure}
\centering
\renewcommand{\imsize}{5.2cm}
\renewcommand{\hspacer}{0.3mm}
\renewcommand{\vspacer}{0.1mm}
\renewcommand{\imscl}{0.01}
\includegraphics[scale=\imscl,width=0.49\linewidth]{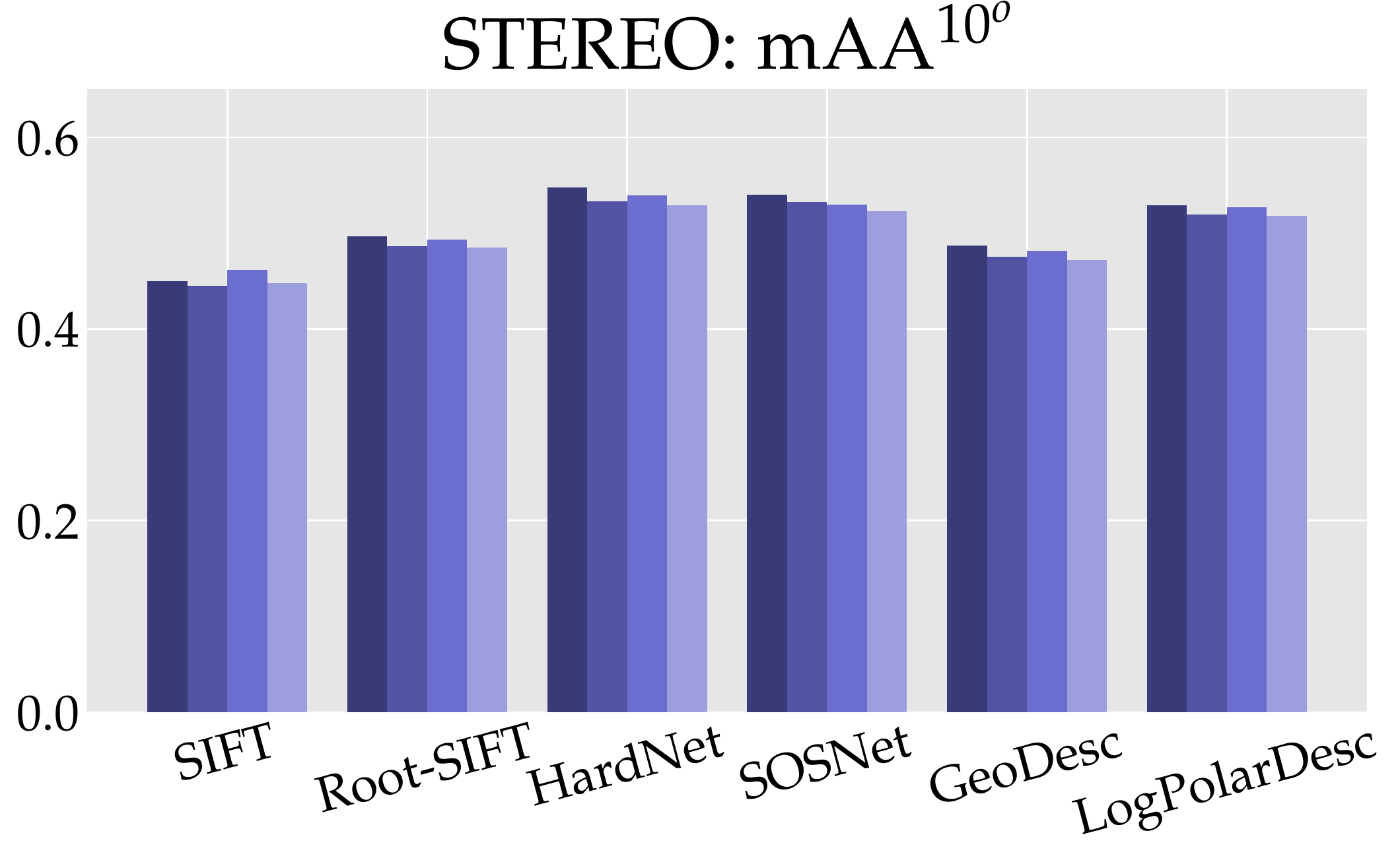}
\includegraphics[scale=\imscl,width=0.49\linewidth]{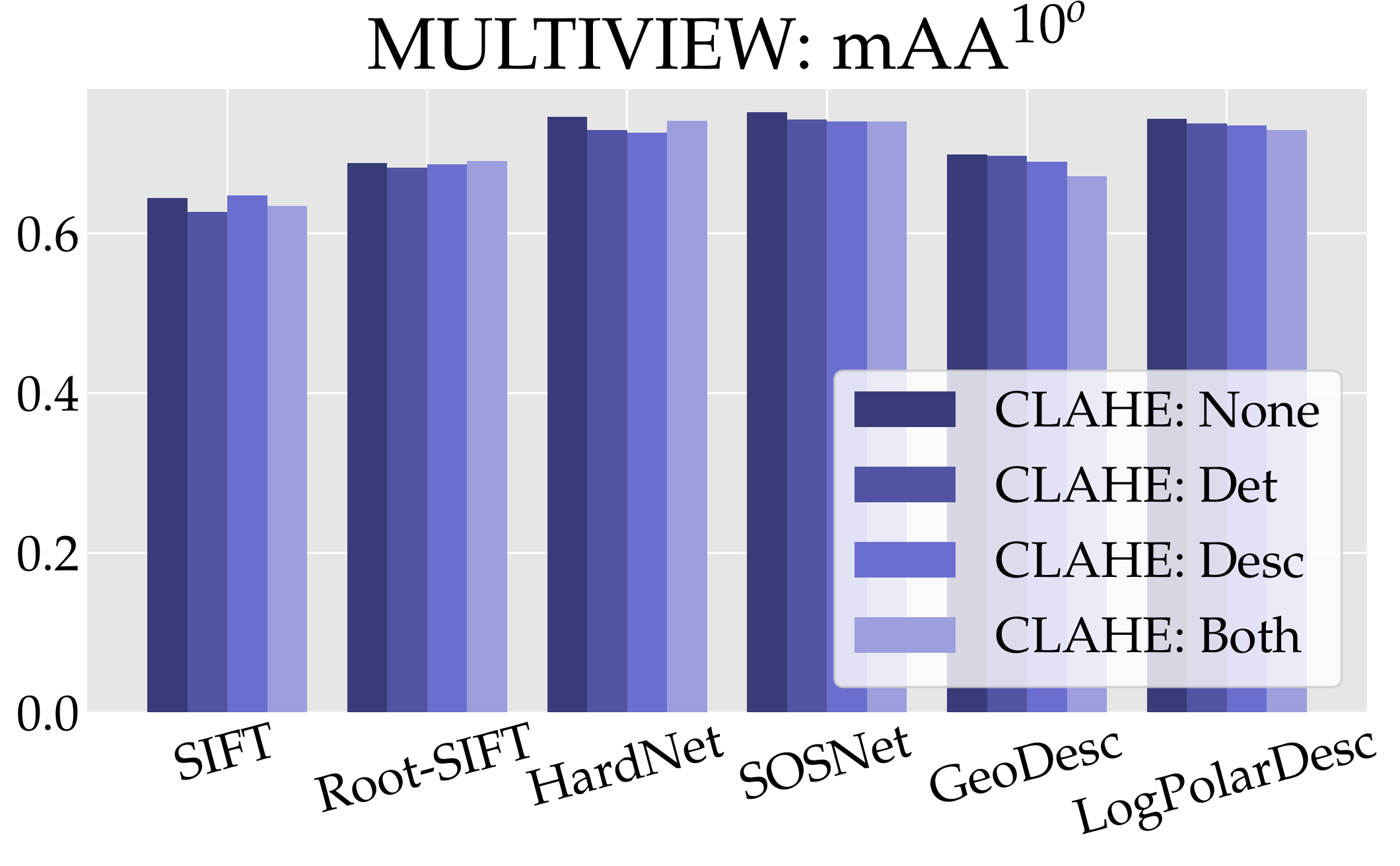}%
\caption{
\edit{{\bf Validation -- Image pre-processing with CLAHE \revision{(8k features)}.} 
We experiment with contrast normalization for keypoint and descriptor extraction.
Results are very similar with or without it.}}
\label{fig:results_clahe}
\end{figure}
\subsection{Image pre-processing --- \fig{results_clahe}}
\label{sec:appendix-clahe}

Contrast normalization is key to invariance against illumination changes -- local feature methods typically apply some normalization strategy over small patches~\cite{Lowe04,Mishchuk17}.
Therefore, we experiment with contrast-limited adaptive histogram equalization (CLAHE) \cite{Pizer87}, as implemented in OpenCV.
We apply it prior to feature detection and/or description with SIFT and several learned descriptors, and display the results in \fig{results_clahe}.
Performance decreases for all learned methods, presumably because they are not trained for it. 
Contrary to our initial expectations, SIFT does not benefit much from it either: the only increase in performance comes from applying it for descriptor extraction, at 2.5\% relative for stereo task and 0.56\% relative for multi-view. 
This might be due to the small number of night-time images in our data. It also falls in line with the observations in~\cite{Dong_2015_CVPR}, which show that SIFT descriptors are actually optimal under certain assumptions.
\definecolor{light-gray}{gray}{0.9}
\renewcommand{\arraystretch}{0.95}
\renewcommand{\tabcolsep}{1.5mm}
\begin{table}
\begin{center}
\scriptsize
\begin{tabular}{@{}lcccccc@{}}
\toprule
& $\ransacReprojTh_{\text{PyR}}$ & $\ransacReprojTh_{\text{DEGEN}}$ & $\ransacReprojTh_{\text{GCR}}$ & $\ransacReprojTh_{\text{MAG}}$ & $r_{\text{stereo}}$ & $r_{\text{multiview}}$ \\
\midrule
CV-SIFT & 0.25 & 0.5 & 0.5 & 1.25 & 0.85 & 0.75 \\
\midrule
CV-$\sqrt{\text{SIFT}}$ & 0.25 & 0.5 & 0.5 & 1.25 & 0.85 & 0.85 \\
CV-SURF & 0.75 & 0.75 & 0.75 & 2 & 0.85 & 0.90 \\
CV-AKAZE & 0.25 & 0.75 & 0.75 & 1.5 & 0.85 & 0.90 \\
CV-ORB & 0.75 & 1 & 1.25 & 2 & 0.85 & 0.95 \\
CV-FREAK & 0.5 & 0.5 & 0.75 & 2 & 0.85 & 0.85 \\
\midrule
VL-DoG-SIFT & 0.25 & 0.5 & 0.5 & 1.5 & 0.85 & 0.80 \\
VL-DoGAff-SIFT & 0.25 & 0.5 & 0.5 & 1.5 & 0.85 & 0.80 \\
VL-Hess-SIFT & 0.2 & 0.5 & 0.5 & 1.5 & 0.85 & 0.80 \\
VL-HessAffNet-SIFT & 0.25 & 0.5 & 0.5 & 1& 0.85 & 0.80 \\
\midrule
CV-DoG/HardNet & 0.25 & 0.5 & 0.5 & 1.5 & 0.90 & 0.80 \\
KeyNet/Hardnet & 0.5 & 0.75 & 0.75 & 2 & 0.95 & 0.85 \\
KeyNet/SOSNet & 0.25 & 0.75 & 0.75 & 1.5 & 0.95 & 0.95  \\
CV-DoG/L2Net & 0.2 & 0.5 & 0.5 & 1.5 & 0.90 & 0.80 \\
CV-DoG/GeoDesc & 0.2 & 0.5 & 0.75 & 1.5 & 0.90 & 0.85 \\
ContextDesc & 0.25 & 0.75 & 0.5 & 1 & 0.95 & 0.85 \\
CV-DoG/SOSNet & 0.25 & 0.5 & 0.75 & 1.5 & 0.90 & 0.80 \\
CV-DoG/LogPolarDesc & 0.2 & 0.5 & 0.5 & 1.5 & 0.90 & 0.80 \\
\midrule
D2-Net (SS) & 1 & 2 & 2 & 7.5 & --- & --- \\
D2-Net (MS) & 1 & 2 & 2 & 5 & --- & --- \\
R2D2 (wasf-n8-big) & 0.75 & 1.25 & 1.25 & 2 & --- & 0.95 \\
\midrule
CV-DoG/TFeat & 0.25 & 0.5 & -- & 1.25 & 0.85 & 0.80 \\
CV-DoG/MKD-Concat & 0.25 & 0.5 & -- & 1.25 & 0.85 & 0.80 \\
CV-DoGAffNet/HardNet & 0.25 & 0.5 & -- & 1.25 & 0.85 & 0.85 \\
\bottomrule
\end{tabular}
\end{center}
\caption{
{\bf Optimal hyper-parameters with 8k features.}
\edit{We summarize the optimal hyperparameters -- the maximum number of RANSAC iterations $\ransacReprojTh$ and the ratio test threshold $r$ -- for each combination of methods.
The number of RANSAC iterations $\ransacMaxIters$ is set to 250k for PyRANSAC, 50k for DEGENSAC, and 10k for both GC-RANSAC and MAGSAC (for the experiments on the validation set).
We use bidirectional matching with the ``both'' strategy.}
}
\label{tbl:optimal_parameters_8k}
\end{table}

\definecolor{light-gray}{gray}{0.9}
\renewcommand{\arraystretch}{0.95}
\renewcommand{\tabcolsep}{1.5mm}
\begin{table}
\begin{center}
\footnotesize
\begin{tabular}{@{}lccccc@{}}
\toprule
& $\ransacReprojTh_{\text{PyR}}$ & $\ransacReprojTh_{\text{DEGEN}}$ & $\ransacReprojTh_{\text{MAG}}$ & $r_{\text{stereo}}$ & $r_{\text{multiview}}$ \\
\midrule
CV-SIFT & 0.75 & 0.5 & 2 & 0.90 & 0.90 \\
\midrule
CV-$\sqrt{\text{SIFT}}$ & 0.5 & 0.5 & 1.25 & 0.90 & 0.90 \\
CV-SURF & 0.25 & 1 & 3 & 0.90 & 0.95 \\
CV-AKAZE & 1 & 0.75 & 2 & 0.90 & 0.95 \\
CV-ORB & 1 & 1.25 & 3 & 0.90 & 0.90 \\
CV-FREAK & 0.75 & 0.75 & 2 & 0.9 & 0.95 \\
\midrule
CV-DoG/HardNet & 0.5 & 0.5 & 1.5 & 0.95 & 0.90 \\
KeyNet/HardNet & 0.5 & 0.75 & 2.5 & 0.95 & 0.90 \\
KeyNet/SOSNet & 0.5 & 0.75 & 1.5 & 0.95 & 0.95 \\
CV-DoG/L2Net & 0.25 & 0.5 & 1.5 & 0.90 & 0.90 \\
CV-DoG/GeoDesc & 0.5 & 0.5 & 1.5 & 0.95 & 0.90 \\
ContextDesc& 0.75 & 0.75 & 2 & 0.95 & 0.95  \\
CV-DoG/SOSNet & 0.5 & 0.75 & 1.5 & 0.95 & 0.90 \\
CV-DoG/LogPolarDesc & 0.5 & 0.5 & 1.5 & 0.95 & 0.90 \\
\midrule
D2-Net (SS) & 1.5 & 2 & 7.5 & --- & --- \\
D2-Net (MS) & 1.5 & 2 & 10 & --- & --- \\
SuperPoint & 0.75 & 1 & 3 & 0.95 & 0.90 \\
LF-Net & 1 & 1 & 4 & 0.95 & 0.95 \\
R2D2 (wasf-n16) & 1.5 & 1.25 & 2 & --- & --- \\
\midrule
CV-DoGAffNet/HardNet & 0.25 & 0.5 &  1.25 & 0.95 & 0.95 \\
\bottomrule
\end{tabular}
\end{center}
\caption{
{\bf Optimal hyper-parameters with 2k features.}
\edit{Equivalent to \tbl{optimal_parameters_8k}. We do not evaluate GC-RANSAC, as it is always outperformed by DEGENSAC and MAGSAC, but keep PyRANSAC as a baseline RANSAC.}
}
\label{tbl:optimal_parameters_2k}
\end{table}

\subsection{Optimal settings breakdown --- Tables~\ref{tbl:optimal_parameters_8k}~and~\ref{tbl:optimal_parameters_2k}}
\label{sec:optimal-settings}

For the sake of clarity, we summarize the optimal hyperparameter combinations from Figs. \ref{fig:plots-ransac-cost-vs-map}, \ref{fig:results_ransac_nort} and \ref{fig:results_both_vs_any_and_ratio_test} in \tbl{optimal_parameters_8k} (for 8k features) and \tbl{optimal_parameters_2k} (for 2k features). 
We set the confidence value to $\ransacConf{=}0.999999$ for all RANSAC variants. 
\edit{Notice how it is better to have more features and a stricter ratio test threshold to filter them out, than having fewer features from the beginning.}

\begin{figure*}
\centering
\renewcommand{\arraystretch}{0.5}
\renewcommand{\tabcolsep}{0.2mm}
\renewcommand{\imsize}{3.44cm}
\renewcommand{\hspacer}{0.3mm}
\renewcommand{\vspacer}{0.1mm}
\renewcommand{\imscl}{0.01}
\small
\begin{center}
\begin{tabular}{@{}ccccc@{}}
\includegraphics[scale=\imscl,width=\imsize]{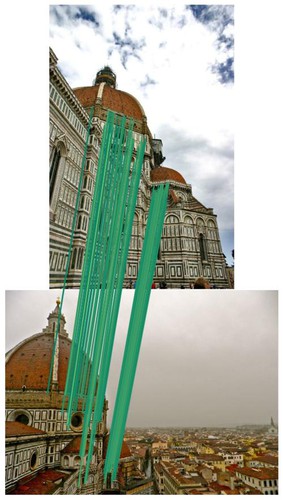} &
\includegraphics[scale=\imscl,width=\imsize]{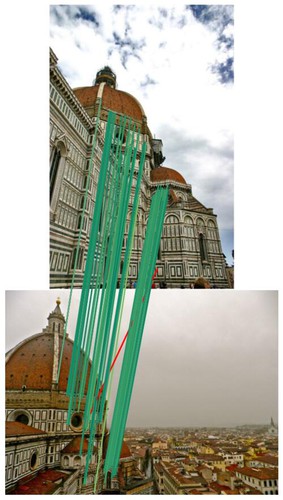} &
\includegraphics[scale=\imscl,width=\imsize]{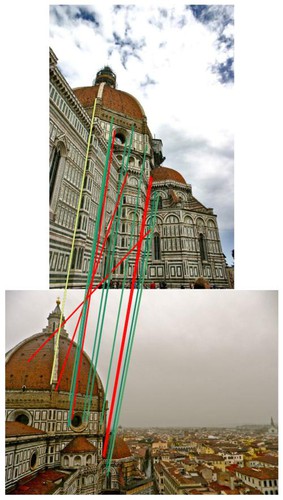} &
\includegraphics[scale=\imscl,width=\imsize]{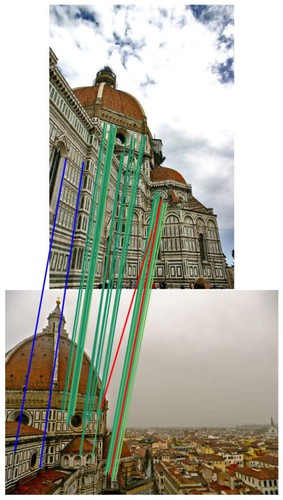} &
\includegraphics[scale=\imscl,width=\imsize]{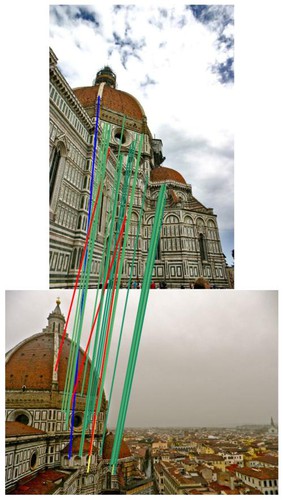}\\

\includegraphics[scale=\imscl,width=\imsize]{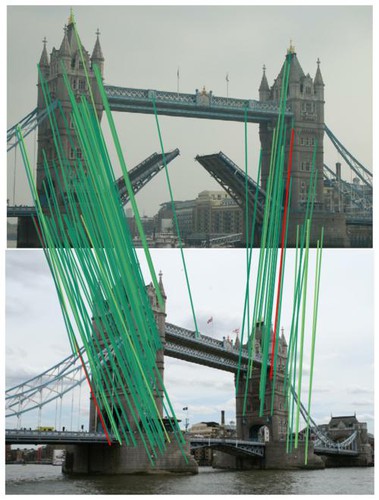} &
\includegraphics[scale=\imscl,width=\imsize]{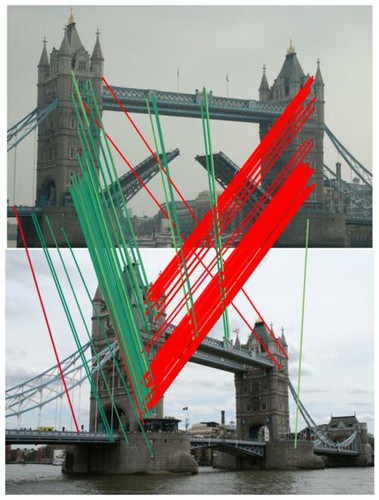} &
\includegraphics[scale=\imscl,width=\imsize]{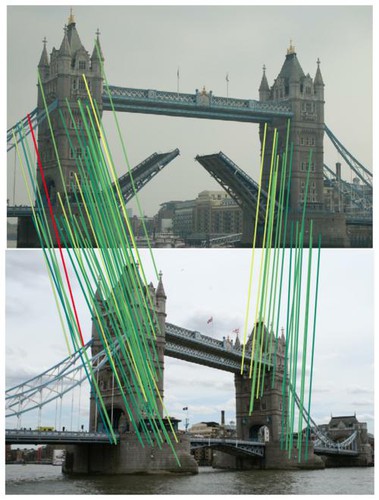} &
\includegraphics[scale=\imscl,width=\imsize]{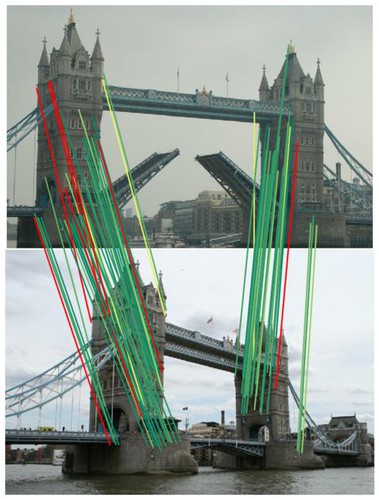} &
\includegraphics[scale=\imscl,width=\imsize]{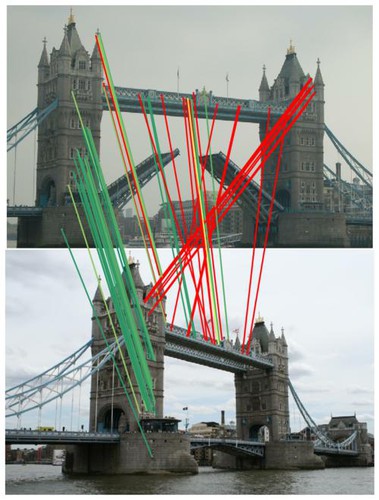}\\

\includegraphics[scale=\imscl,width=\imsize]{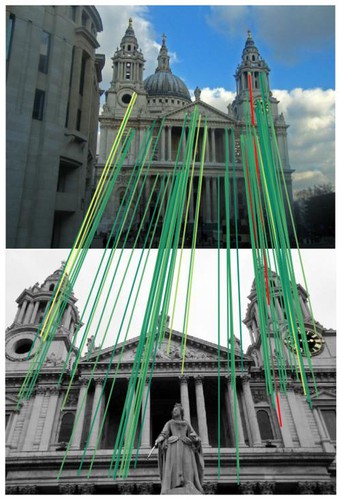} &
\includegraphics[scale=\imscl,width=\imsize]{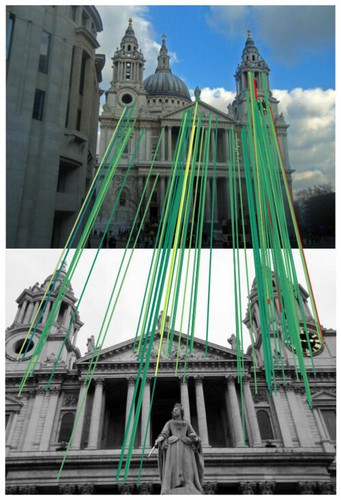} &
\includegraphics[scale=\imscl,width=\imsize]{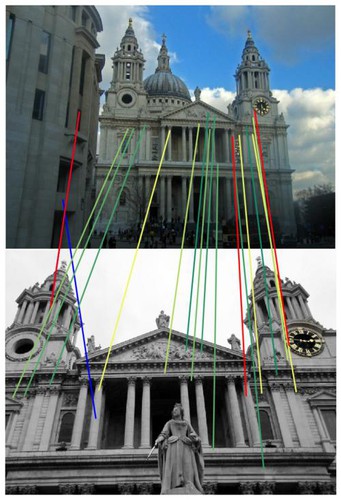} &
\includegraphics[scale=\imscl,width=\imsize]{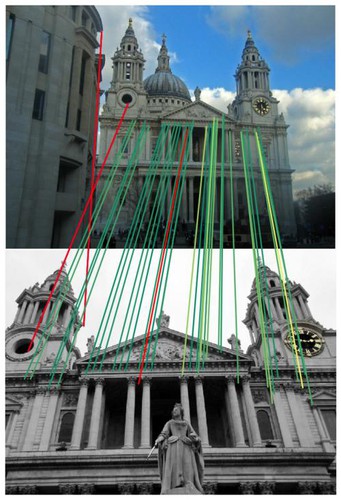} &
\includegraphics[scale=\imscl,width=\imsize]{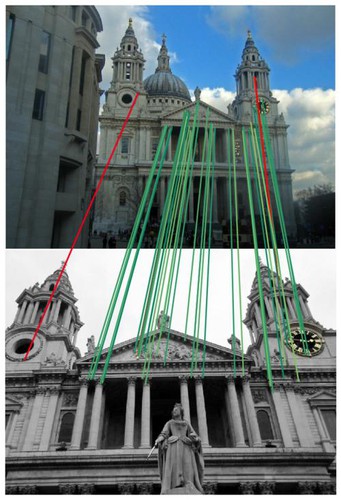}\\

\includegraphics[scale=\imscl,width=\imsize]{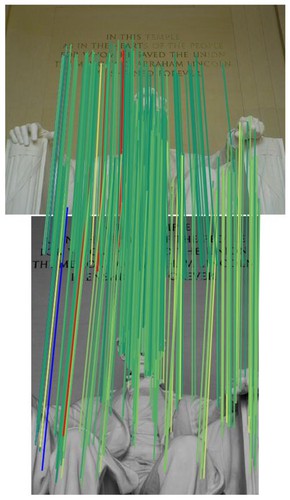} &
\includegraphics[scale=\imscl,width=\imsize]{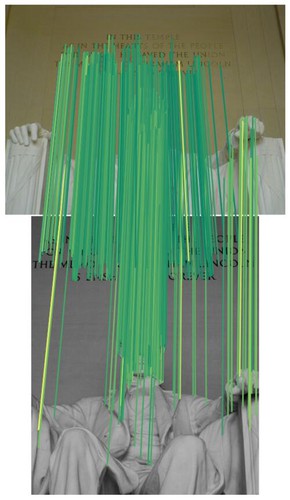} &
\includegraphics[scale=\imscl,width=\imsize]{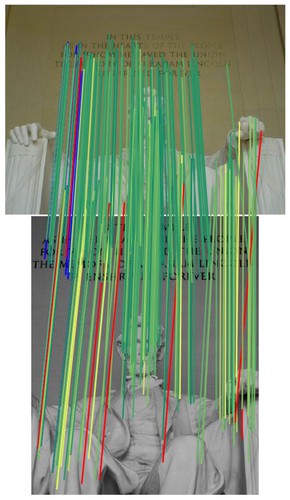} &
\includegraphics[scale=\imscl,width=\imsize]{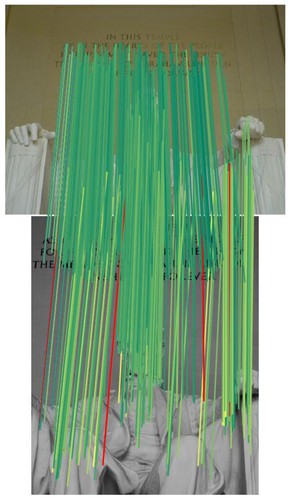} &
\includegraphics[scale=\imscl,width=\imsize]{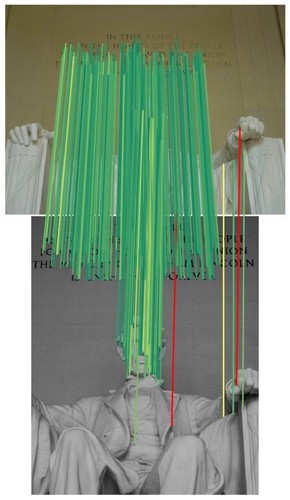}\\

(a) RootSIFT & (b) Hessian-SIFT & (c) SURF & (d) AKAZE & (e) ORB \\

\end{tabular}
\end{center}
\caption{{\bf Qualitative results for the stereo task -- \edit{``Classical''} features.} We plot the matches predicted by each \edit{local feature, with DEGENSAC}. \edit{Matches above a 5-pixel error threshold are drawn in \textcolor{red}{{\bf red}}, and those below are color-coded by their error, from 0 (\textcolor{ForestGreen}{{\bf green}}) to 5 pixels (\textcolor{Goldenrod}{{\bf yellow}}). Matches for which we do not have depth estimates are drawn in \textcolor{blue}{{\bf blue}}.}}
\label{fig:supp_stereo}
\end{figure*}
\begin{figure*}
\centering
\renewcommand{\arraystretch}{0.5}
\renewcommand{\tabcolsep}{0.2mm}
\renewcommand{\imsize}{3.44cm}
\renewcommand{\hspacer}{0.3mm}
\renewcommand{\vspacer}{0.1mm}
\renewcommand{\imscl}{0.01}
\small
\begin{center}
\begin{tabular}{@{}ccccc@{}}
\includegraphics[scale=\imscl,width=\imsize]{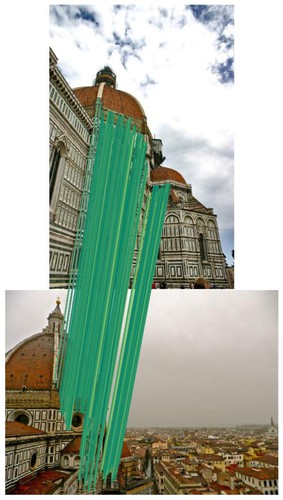} &
\includegraphics[scale=\imscl,width=\imsize]{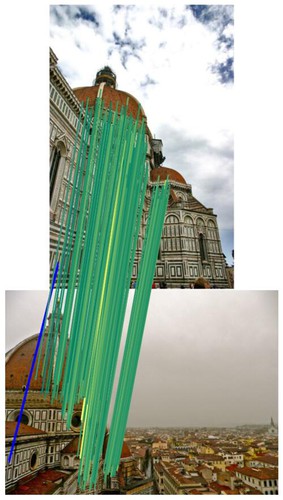} &
\includegraphics[scale=\imscl,width=\imsize]{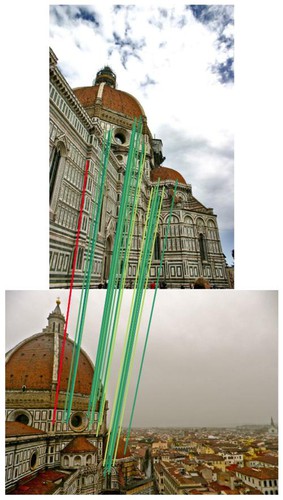} &
\includegraphics[scale=\imscl,width=\imsize]{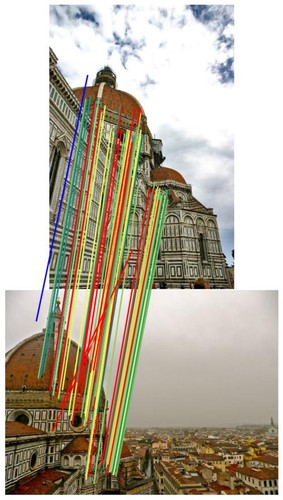} &
\includegraphics[scale=\imscl,width=\imsize]{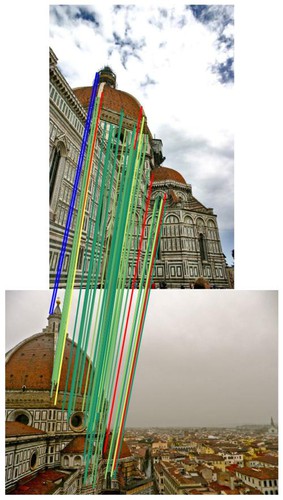}\\

\includegraphics[scale=\imscl,width=\imsize]{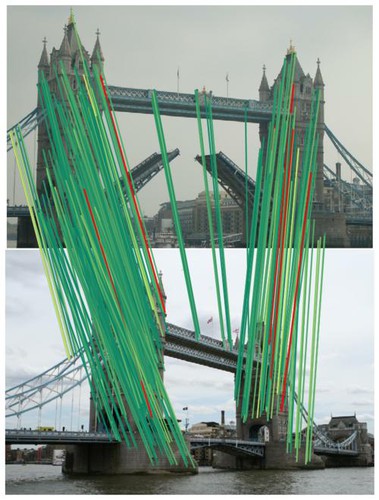} &
\includegraphics[scale=\imscl,width=\imsize]{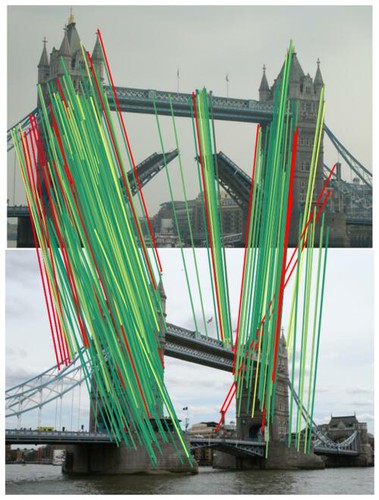} &
\includegraphics[scale=\imscl,width=\imsize]{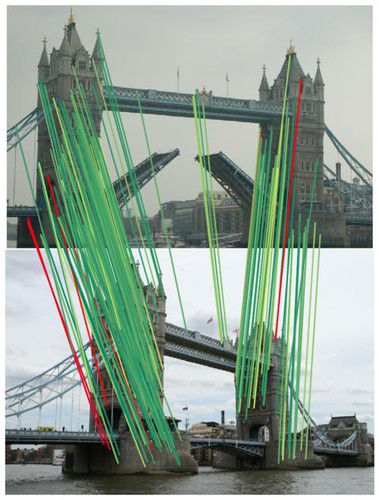} &
\includegraphics[scale=\imscl,width=\imsize]{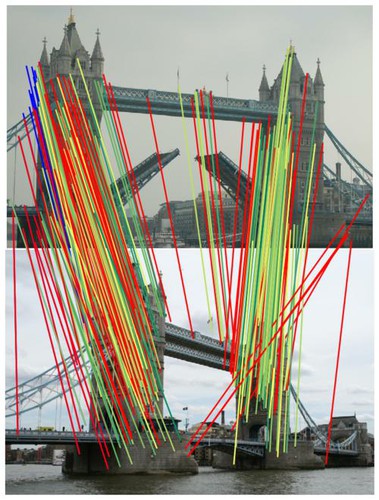} &
\includegraphics[scale=\imscl,width=\imsize]{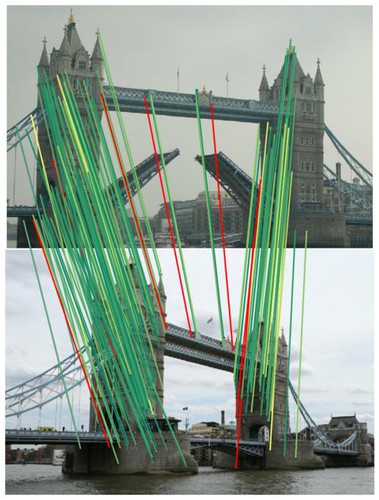}\\

\includegraphics[scale=\imscl,width=\imsize]{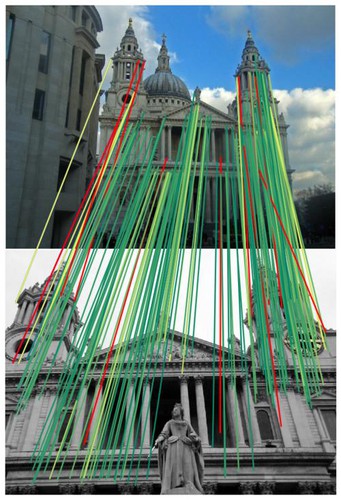} &
\includegraphics[scale=\imscl,width=\imsize]{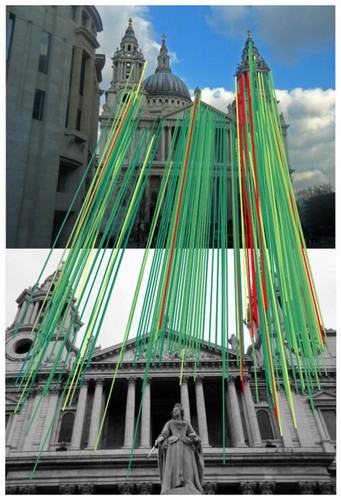} &
\includegraphics[scale=\imscl,width=\imsize]{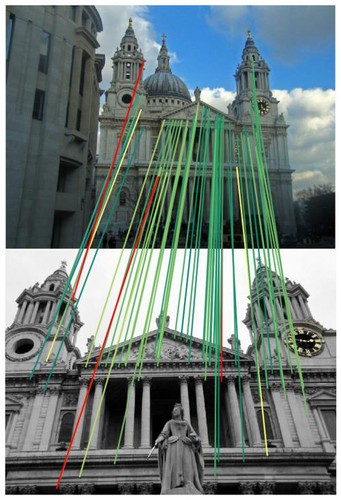} &
\includegraphics[scale=\imscl,width=\imsize]{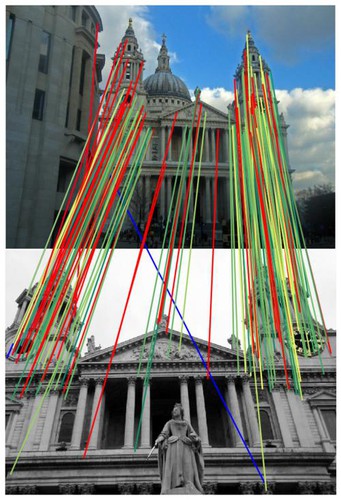} &
\includegraphics[scale=\imscl,width=\imsize]{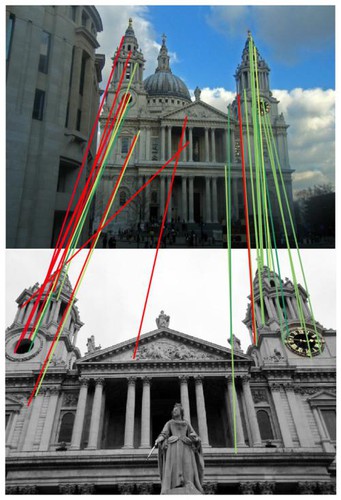}\\

\includegraphics[scale=\imscl,width=\imsize]{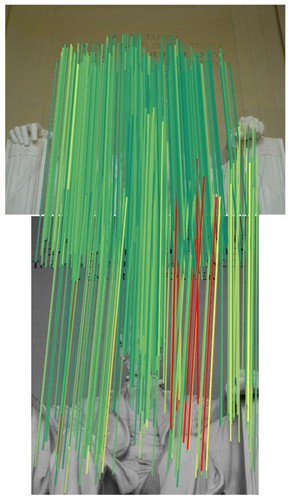} &
\includegraphics[scale=\imscl,width=\imsize]{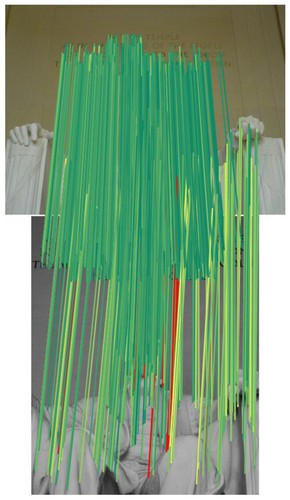} &
\includegraphics[scale=\imscl,width=\imsize]{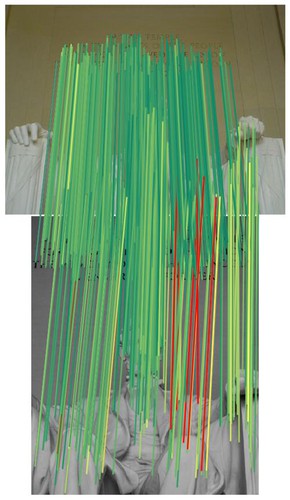} &
\includegraphics[scale=\imscl,width=\imsize]{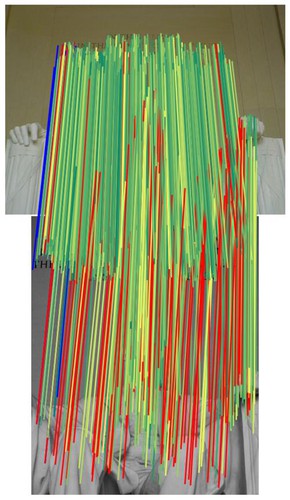} &
\includegraphics[scale=\imscl,width=\imsize]{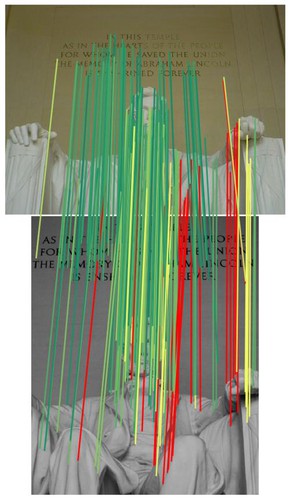}\\
(a) DoG+HardNet & (b) KeyNet+HardNet & (c) R2D2 & (d) D2-Net (MS) & (e) SuperPoint (2k) \\

\end{tabular}
\end{center}
\caption{{\bf Qualitative results for the stereo task -- Learned features.}
Color-coded as in \fig{supp_stereo}.}
\label{fig:supp_stereo2}
\end{figure*}
\begin{figure*}
\centering
\renewcommand{\arraystretch}{0.5}
\renewcommand{\tabcolsep}{0.2mm}
\renewcommand{\imsize}{3.44cm}
\renewcommand{\hspacer}{0.3mm}
\renewcommand{\vspacer}{0.1mm}
\renewcommand{\imscl}{0.01}
\small
\begin{center}
\begin{tabular}{@{}ccccc@{}}

\includegraphics[scale=\imscl,width=\imsize]{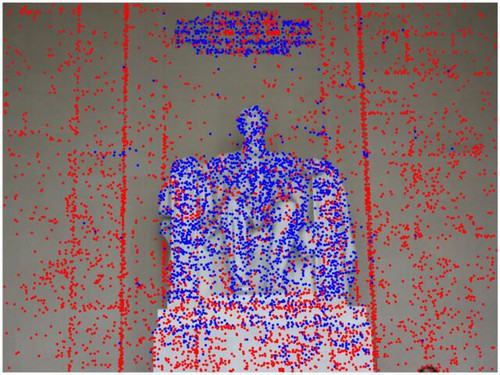} &
\includegraphics[scale=\imscl,width=\imsize]{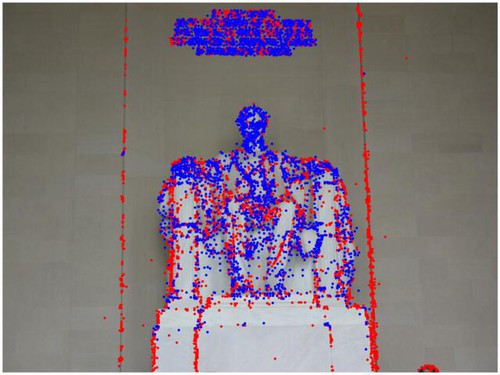} &
\includegraphics[scale=\imscl,width=\imsize]{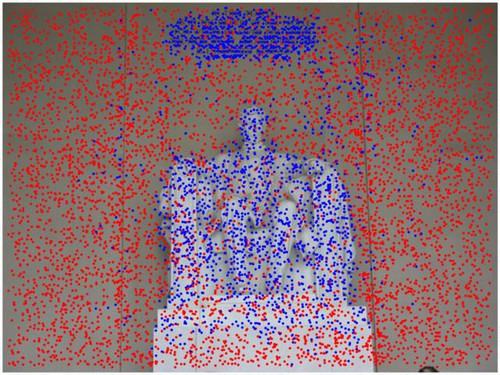} &
\includegraphics[scale=\imscl,width=\imsize]{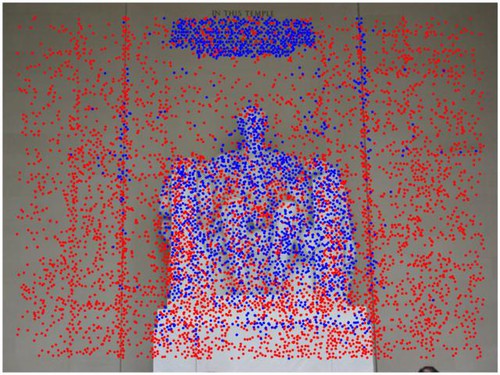} &
\includegraphics[scale=\imscl,width=\imsize]{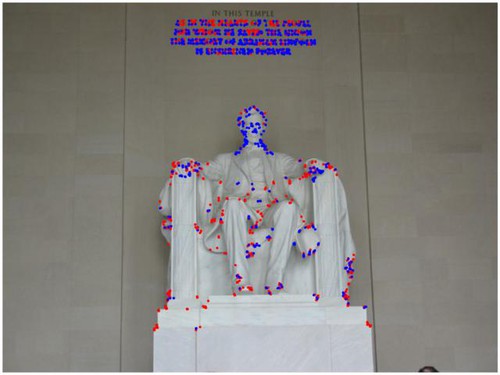} \\

\includegraphics[scale=\imscl,width=\imsize]{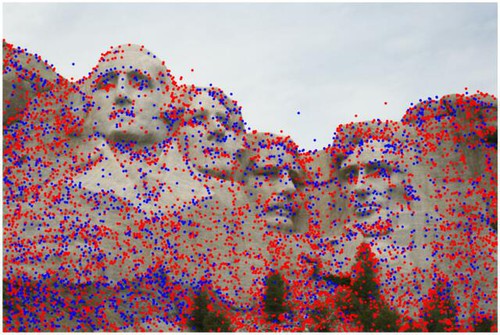} &
\includegraphics[scale=\imscl,width=\imsize]{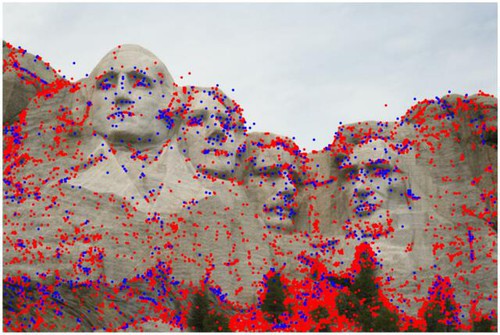} &
\includegraphics[scale=\imscl,width=\imsize]{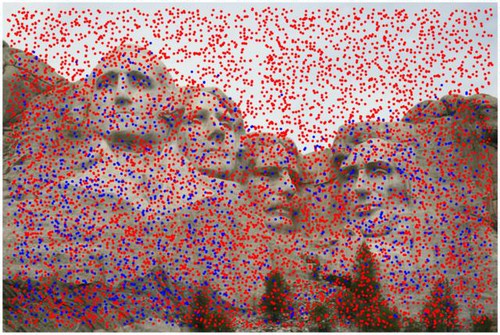} &
\includegraphics[scale=\imscl,width=\imsize]{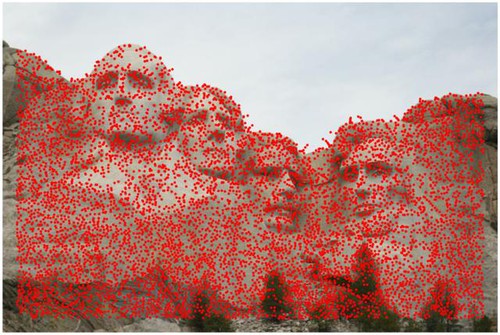} &
\includegraphics[scale=\imscl,width=\imsize]{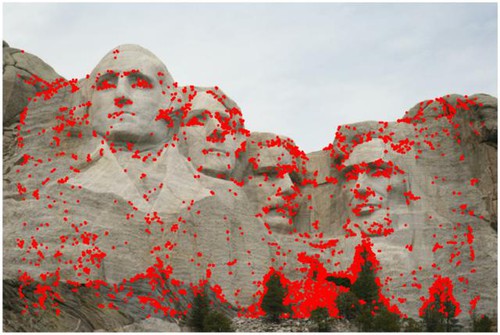} \\

\includegraphics[scale=\imscl,width=\imsize]{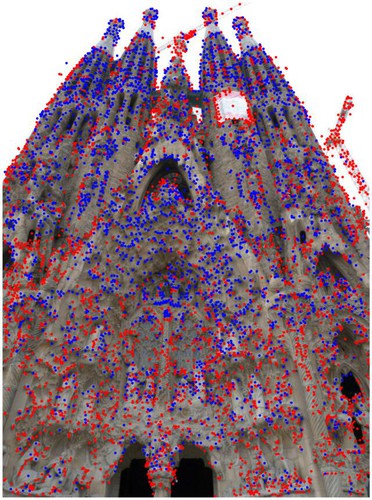} &
\includegraphics[scale=\imscl,width=\imsize]{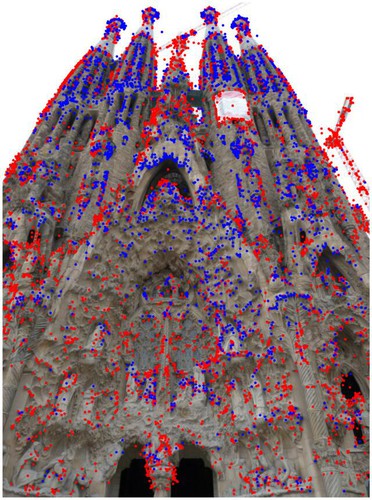} &
\includegraphics[scale=\imscl,width=\imsize]{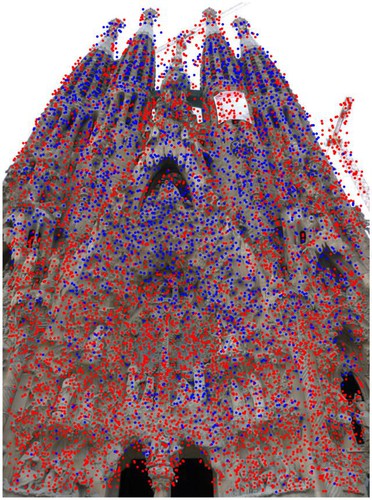} &
\includegraphics[scale=\imscl,width=\imsize]{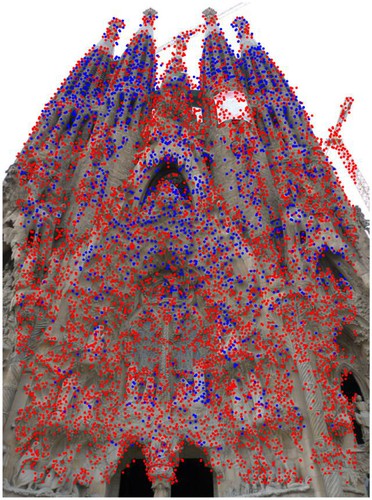} &
\includegraphics[scale=\imscl,width=\imsize]{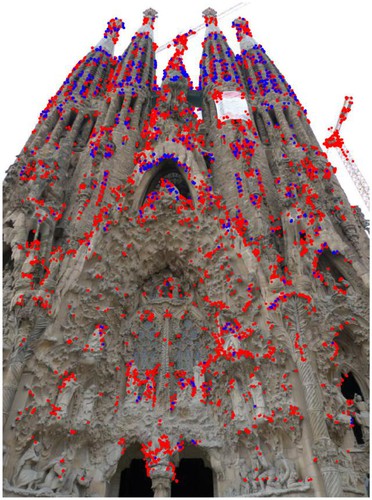} \\

\includegraphics[scale=\imscl,width=\imsize]{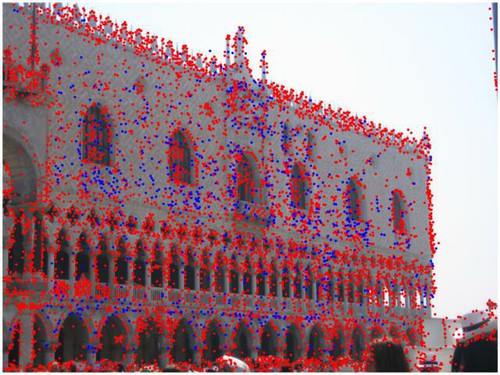} &
\includegraphics[scale=\imscl,width=\imsize]{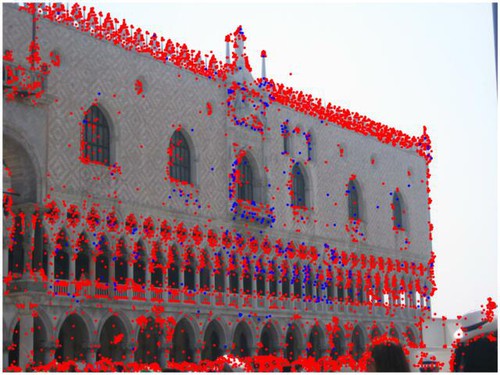} &
\includegraphics[scale=\imscl,width=\imsize]{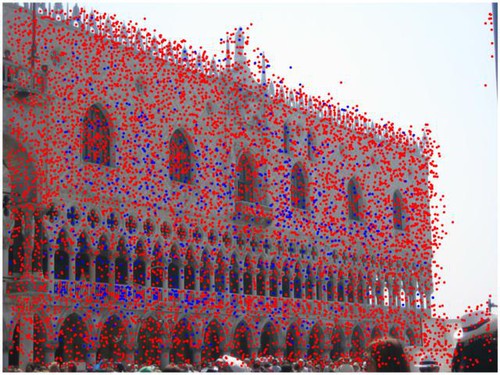} &
\includegraphics[scale=\imscl,width=\imsize]{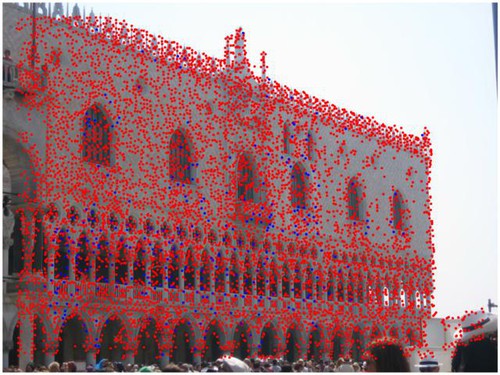} &
\includegraphics[scale=\imscl,width=\imsize]{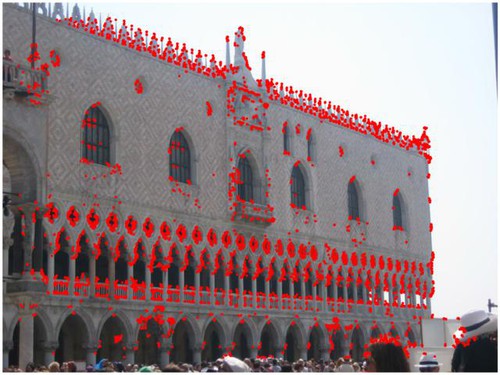} \\

\includegraphics[scale=\imscl,width=\imsize]{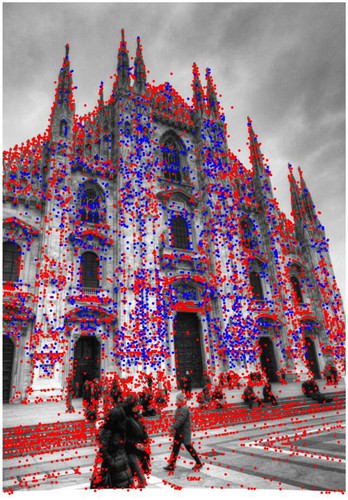} &
\includegraphics[scale=\imscl,width=\imsize]{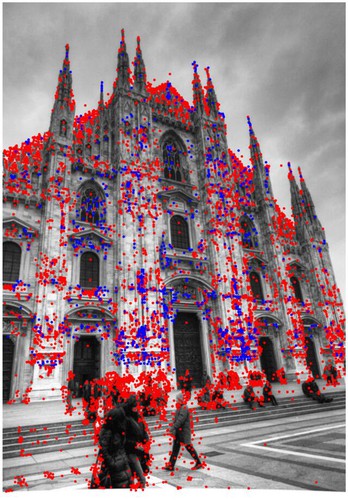} &
\includegraphics[scale=\imscl,width=\imsize]{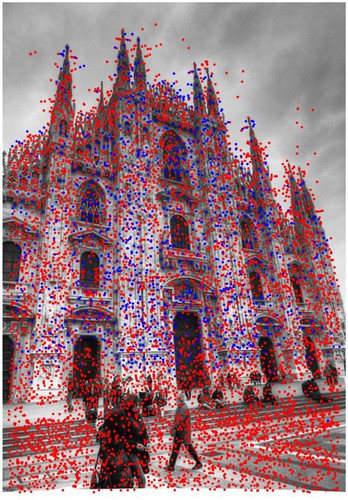} &
\includegraphics[scale=\imscl,width=\imsize]{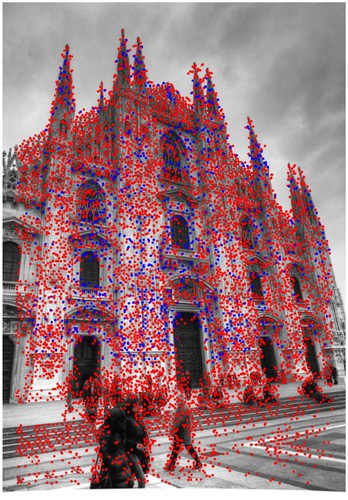} &
\includegraphics[scale=\imscl,width=\imsize]{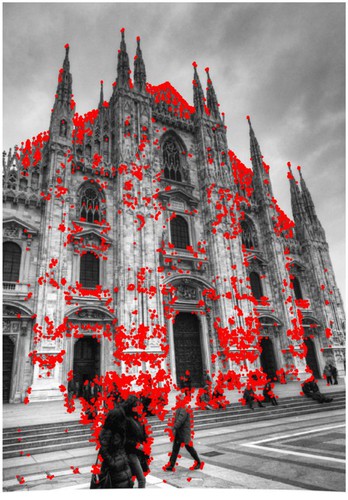} \\
(a) RootSIFT & (b) Hessian-SIFT & (c) SURF & (d) AKAZE & (e) ORB \\
\end{tabular}
\end{center}
\caption{{\bf Qualitative results for the multi-view task -- \edit{``Classical''} features.}
We show images and the keypoints detected on them for different methods, with the points reconstructed by COLMAP in \textcolor{blue}{{\bf blue}}, and the ones that are not used in the 3D model in \textcolor{red}{{\bf red}} -- more blue points indicate denser 3D models. These results correspond the multiview-task for a 25-image subset.
}
\label{fig:supp_multiview}
\end{figure*}
\begin{figure*}
\centering
\renewcommand{\arraystretch}{0.5}
\renewcommand{\tabcolsep}{0.2mm}
\renewcommand{\imsize}{3.44cm}
\renewcommand{\hspacer}{0.3mm}
\renewcommand{\vspacer}{0.1mm}
\renewcommand{\imscl}{0.01}
\small
\begin{center}
\begin{tabular}{@{}ccccc@{}}

\includegraphics[scale=\imscl,width=\imsize]{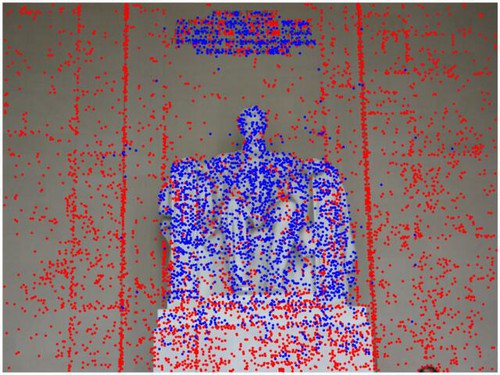} &
\includegraphics[scale=\imscl,width=\imsize]{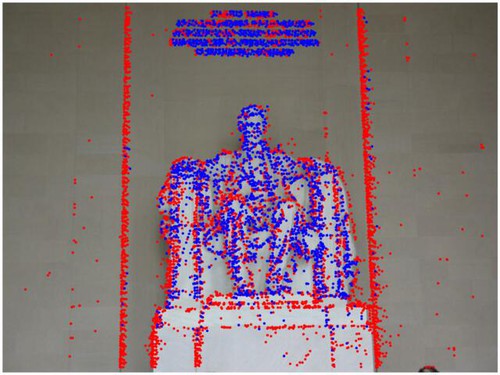} &
\includegraphics[scale=\imscl,width=\imsize]{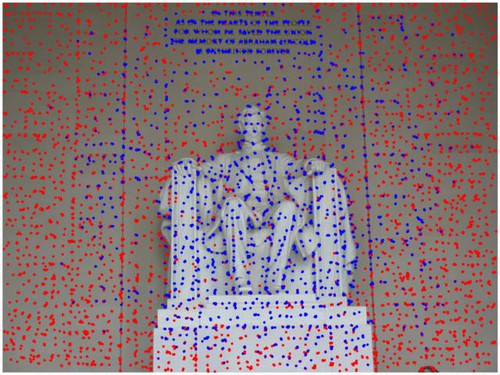} &
\includegraphics[scale=\imscl,width=\imsize]{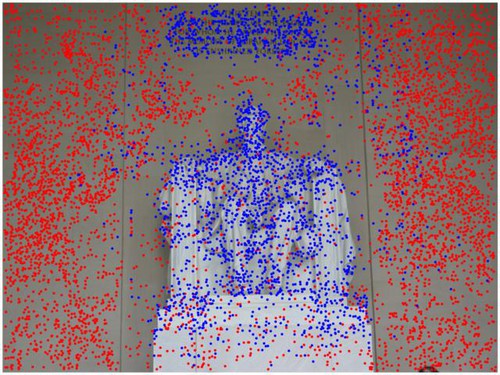} &
\includegraphics[scale=\imscl,width=\imsize]{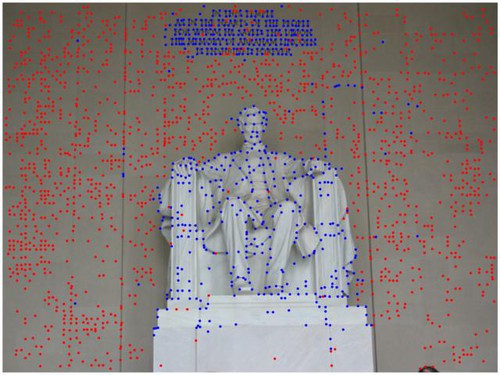} \\

\includegraphics[scale=\imscl,width=\imsize]{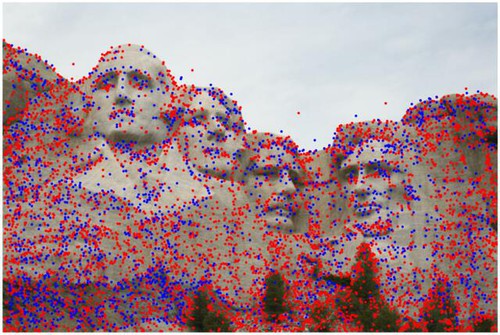} &
\includegraphics[scale=\imscl,width=\imsize]{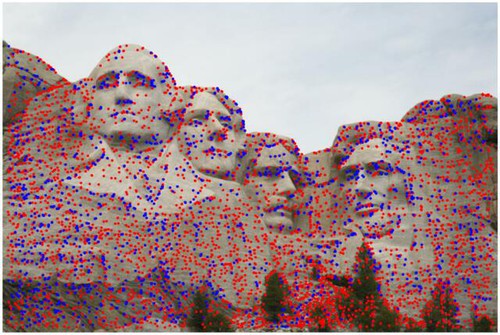} &
\includegraphics[scale=\imscl,width=\imsize]{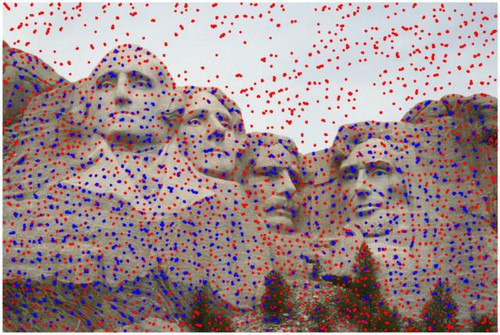} &
\includegraphics[scale=\imscl,width=\imsize]{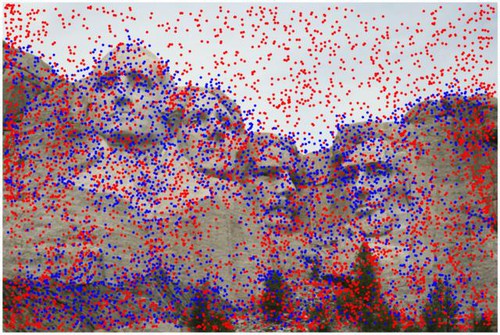} &
\includegraphics[scale=\imscl,width=\imsize]{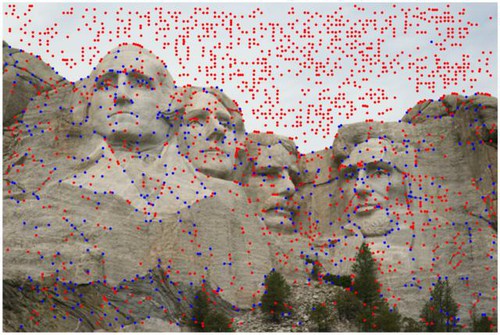} \\

\includegraphics[scale=\imscl,width=\imsize]{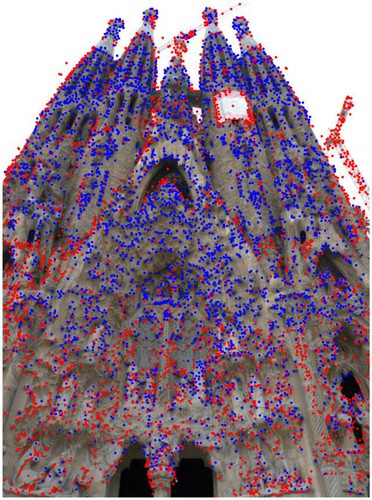} &
\includegraphics[scale=\imscl,width=\imsize]{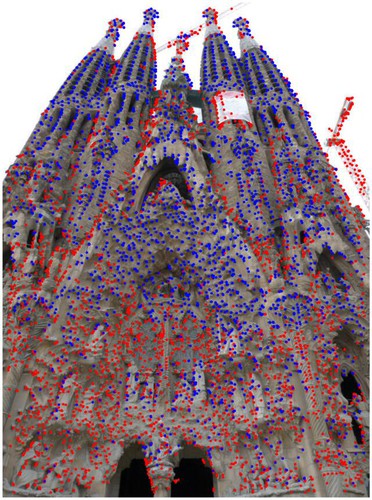} &
\includegraphics[scale=\imscl,width=\imsize]{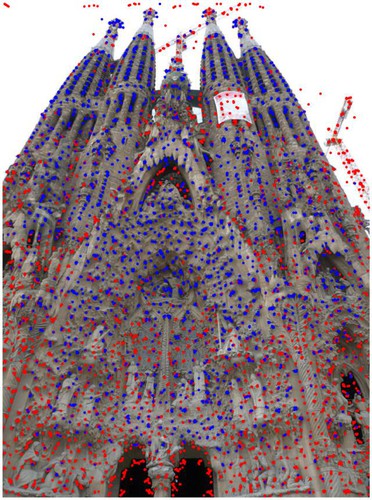} &
\includegraphics[scale=\imscl,width=\imsize]{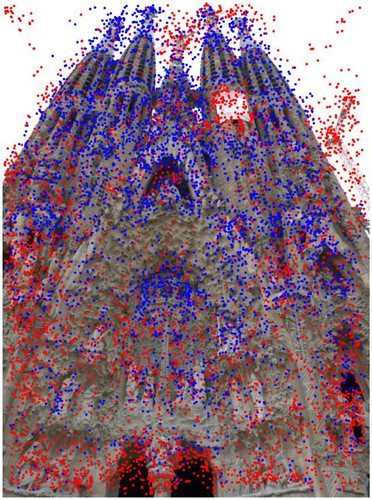} &
\includegraphics[scale=\imscl,width=\imsize]{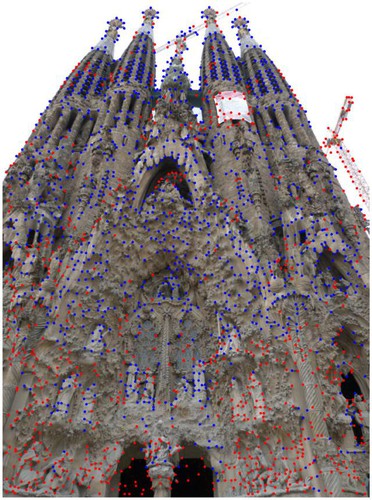} \\

\includegraphics[scale=\imscl,width=\imsize]{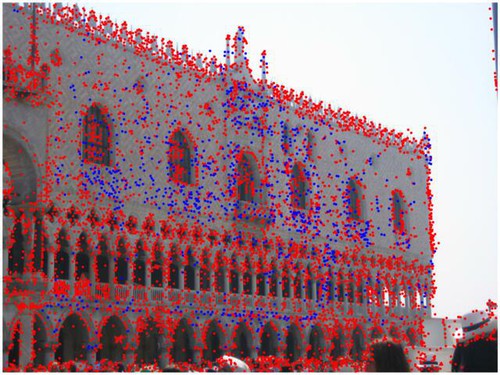} &
\includegraphics[scale=\imscl,width=\imsize]{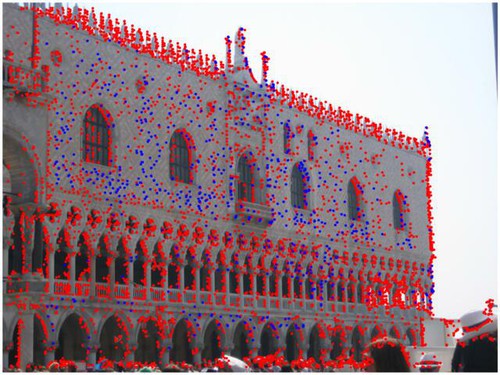} &
\includegraphics[scale=\imscl,width=\imsize]{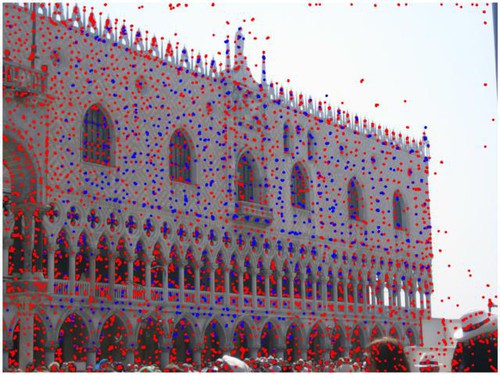} &
\includegraphics[scale=\imscl,width=\imsize]{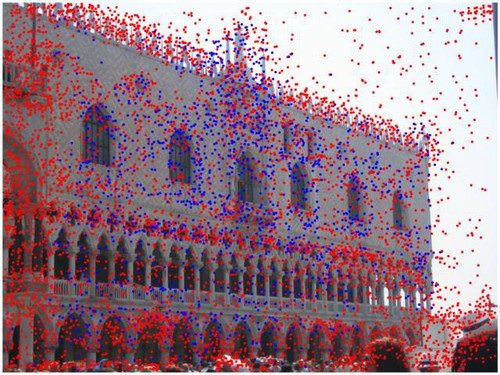} &
\includegraphics[scale=\imscl,width=\imsize]{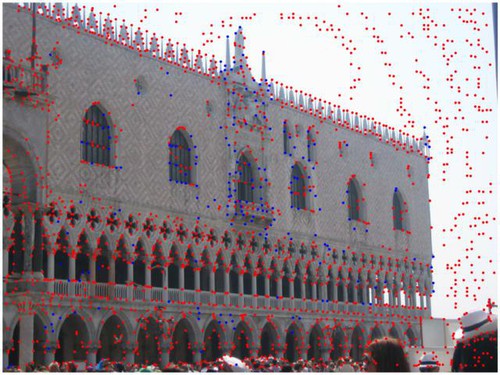} \\

\includegraphics[scale=\imscl,width=\imsize]{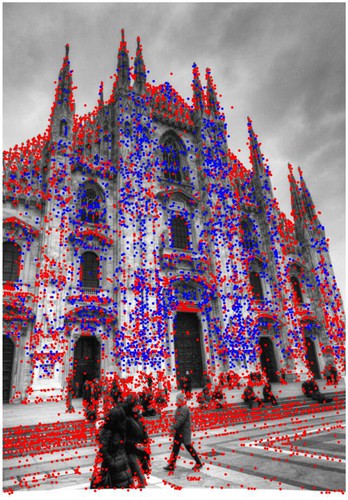} &
\includegraphics[scale=\imscl,width=\imsize]{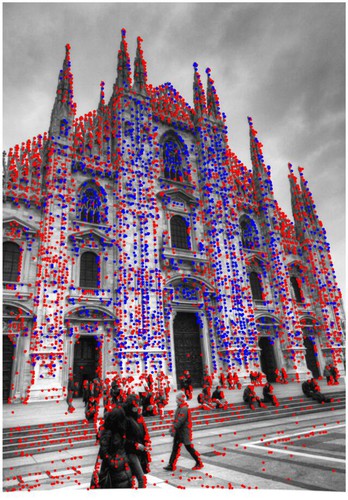} &
\includegraphics[scale=\imscl,width=\imsize]{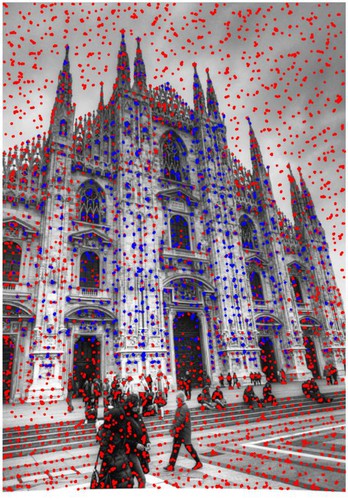} &
\includegraphics[scale=\imscl,width=\imsize]{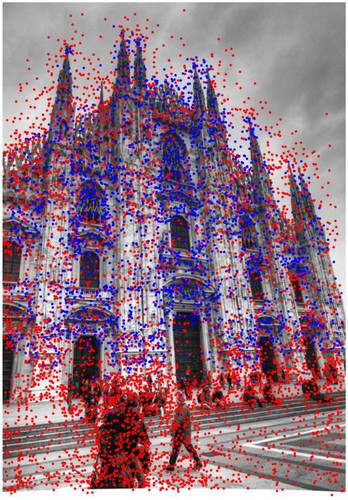} &
\includegraphics[scale=\imscl,width=\imsize]{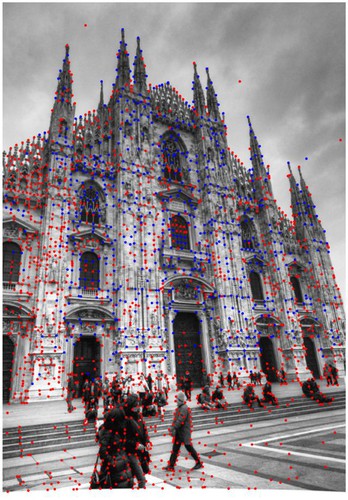} \\

(a) DoG+HardNet & (b) KeyNet+HardNet & (c) R2D2 & (d) D2-Net (MS) & (e) SuperPoint (2k) \\
\end{tabular}
\end{center}
\caption{{\bf Qualitative results for the multi-view task -- Learned features.}
\edit{Color-coded as in \fig{supp_multiview}.}
}
\label{fig:supp_multiview2}
\end{figure*}

\section{Conclusions}
\label{sec:conclusions}

We introduce a comprehensive benchmark for local features and robust estimation algorithms.
The modular structure of its pipeline allows to easily integrate, configure, and combine methods and heuristics.
We demonstrate this by evaluating dozens of popular algorithms, from seminal works to the cutting edge of machine learning research, and show that classical solutions may still outperform the perceived state of the art with proper settings.

The experiments carried out through the benchmark and reported in this paper have already revealed unexpected, non-intuitive properties of various components of the SfM pipeline, which will benefit SfM development, \eg, the need to tune RANSAC to the particular feature detector \emph{and} descriptor and to select specific settings for a particular RANSAC variant.
Other 
\edit{interesting}
facts have been uncovered by our tests, such as that the optimal set-ups across different tasks (stereo and multiview) may differ, or \edit{that methods that perform better on proxy tasks, like patch \edit{retrieval} or repeatability, may be inferior on the downstream task}.
Our work is open-sourced and makes the basis of an open challenge for image matching with sparse methods.

\bibliographystyle{ieee_fullname}
\bibliography{string,vision,learning,others}

\end{document}